\documentclass{article}

\usepackage[nonatbib,final]{neurips_2021}

\usepackage[utf8]{inputenc} 
\usepackage[T1]{fontenc}    
\usepackage[colorlinks = true,
            linkcolor = magenta,
            urlcolor  = blue,
            citecolor = green,
            anchorcolor = blue]{hyperref}

\usepackage{url}            
\usepackage{booktabs}       
\usepackage{amsfonts}       
\usepackage{nicefrac}       
\usepackage{microtype}      
\usepackage{xcolor}         

\usepackage{amsmath,amsfonts,amssymb,amsthm,bm}
\usepackage{enumitem}
\usepackage{microtype}
\usepackage{graphbox}
\usepackage{graphicx}
\usepackage{caption}
\usepackage{subcaption}
\usepackage{booktabs} 
\usepackage{amsfonts}       
\usepackage{nicefrac}       
\usepackage{float}
\usepackage{wrapfig}

\usepackage{amsmath,amsfonts,bm,bbm}

\def\1{\bm{1}}

\DeclareMathAlphabet{\mathsfit}{\encodingdefault}{\sfdefault}{m}{sl}
\SetMathAlphabet{\mathsfit}{bold}{\encodingdefault}{\sfdefault}{bx}{n}

\def\gP{{\mathcal{P}}}

\def\gZ{{\mathcal{Z}}}

\def\sR{{\mathbb{R}}}

\def\sV{{\mathbb{V}}}

\DeclareMathOperator{\E}{\mathbb{E}}

\newcommand{\R}{\mathbb{R}}

\DeclareMathOperator*{\argmin}{arg\,min}

\DeclareMathOperator{\sign}{sign}
\DeclareMathOperator{\Tr}{Tr}

\newcommand{\bs}[1]{\boldsymbol{#1}}

\newcommand{\sigx}[1]{\sigma_{x,#1}}
\newcommand{\sigb}[1]{\sigma_{\beta,#1}}

\newcommand{\dd}{\mathrm{d}} 

\newcommand{\citet}[1]{\cite{#1}}
\newcommand{\citep}[1]{\cite{#1}}

\title{On the interplay between data structure and loss function in classification problems}

\author{%
  St\'ephane d'Ascoli\\
  Facebook AI Research, Paris\\
  Department of Physics, École Normale Supérieure, Paris\\
  \texttt{stephane.dascoli@ens.fr}

  \AND
  Marylou Gabri\'e\\
 New York University, New York \\
  Flatiron Institute, New York \\
   \texttt{mgabrie@nyu.edu} \\
   \AND
  Levent Sagun\\
  Facebook AI Research, Paris\\
   \And
  Giulio Biroli\\
  Department of Physics, École Normale Supérieure, Paris\\
}

\begin{document}

\maketitle

\begin{abstract}
One of the central puzzles in modern machine learning is the ability of heavily overparametrized models to generalize well. Although the low-dimensional structure of typical datasets is key to this behavior, most theoretical studies of overparametrization focus on isotropic inputs. In this work, we instead consider an analytically tractable model of structured data, where the input covariance is built from independent blocks allowing us to tune the saliency of low-dimensional structures and their alignment with respect to the target function.

Using methods from statistical physics, we derive a precise asymptotic expression for the train and test error achieved by random feature models trained to classify such data, which is valid for any convex loss function. We study in detail how the data structure affects the double descent curve, and show that in the over-parametrized regime, its impact is greater for logistic loss than for mean-squared loss: the easier the task, the wider the gap in performance at the advantage of the logistic loss. 
Our insights are confirmed by numerical experiments on MNIST and CIFAR10.


\end{abstract}

\section{Introduction}

Classical wisdom teaches us that a learning model should have just the right number of parameters to learn from a dataset without overfitting it. However, recent years have seen the emergence of massively over-parametrized models which manage to generalize well on high-dimensional tasks~\citep{zagoruyko2016wide,brown2020language}, somehow avoiding both the curse of dimensionality and the pitfall of overfitting. This generalization capacity in the over-parametrized regime continues to puzzle rigorous understanding, in particular for 
deep neural networks~\citep{zhang2016understanding,neyshabur2018towards,advani2020high}, despite their remarkable achievements over the past decade~\citep{krizhevsky2012imagenet, lecun2015deep,hinton2012deep,sutskever2014sequence}. Various works have given evidence of a double descent curve ~\citep{advani2020high, belkin2018reconciling, hastie2019surprises, spigler2019jamming,geiger2020scaling}, whereby the test error first decreases as the number of parameters increases, then peaks, then decreases again monotonically. The overfitting peak occurs at the \emph{interpolation threshold} where training error vanishes, a well-studied phenomenon in the statistical physics literature~\citep{opper1996statistical,engel2001statistical,krzakala2007landscape,franz2016simplest,geiger2019jamming}. 

Although the underlying structure of data plays a major role in the generalization ability of over-parametrized models~\cite{pope2021intrinsic}, it has been little studied from a theoretical point of view. The first aspect of data structure is the distribution of the inputs: MNIST and CIFAR10 have the same number of classes and images, yet generalization is harder for CIFAR10 because the images are more complex than handwritten numbers. The second aspect is the rule between inputs and outputs: a random labelling of CIFAR10 can be learned by a neural network but offers no possibility of generalization \citep{zhang2016understanding}. Characterizing the structure of real-world data involves studying the interplay between these two aspects. In this direction, simple and interpretable models of structured data can prove useful to understand what underlies the behavior of generalization. 

In this work, we present a simple model of learning and its analytical solution which simultaneously enables us to:
\begin{itemize}
    \item[(i)] study the effect of overparametrization for a given task; 
    \item[(ii)] analyze both regression and classification tasks; 
    \item[(iii)] control the structure of the inputs and their relationship with the labels. 
\end{itemize}

\begin{figure} 
    \centering
    \begin{subfigure}[b]{.25\linewidth}
    \includegraphics[width=\linewidth]{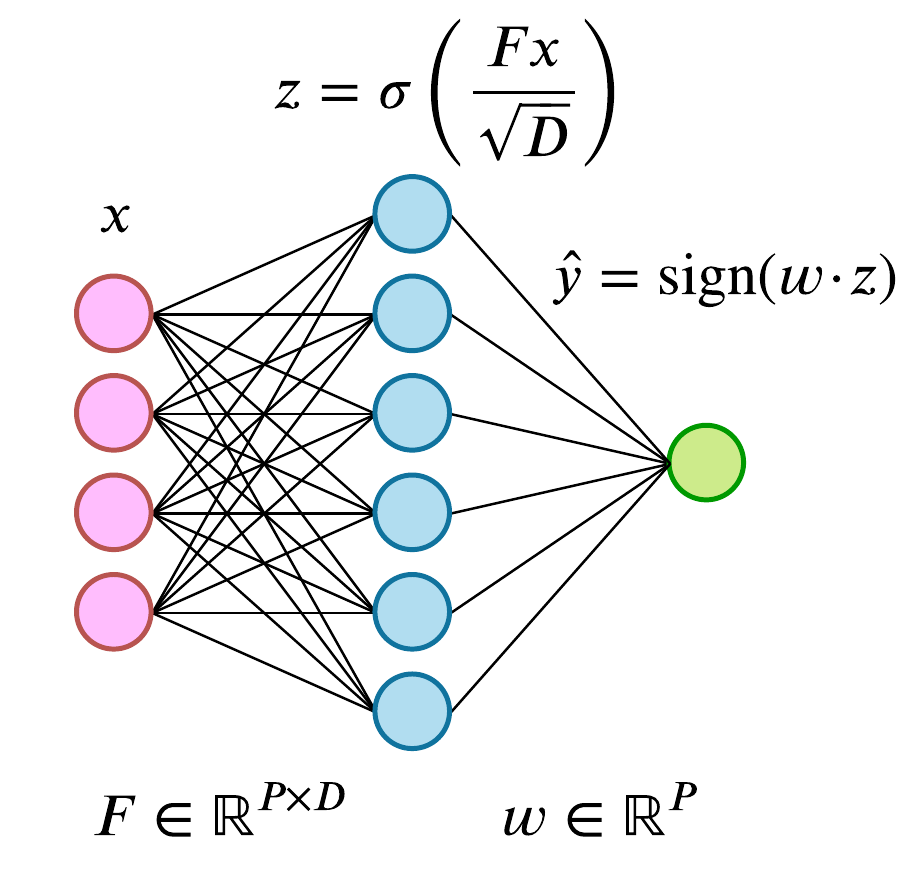}
    \caption{Random features model}
    \end{subfigure}
    \begin{subfigure}[b]{.74\linewidth}
    \includegraphics[width=\linewidth]{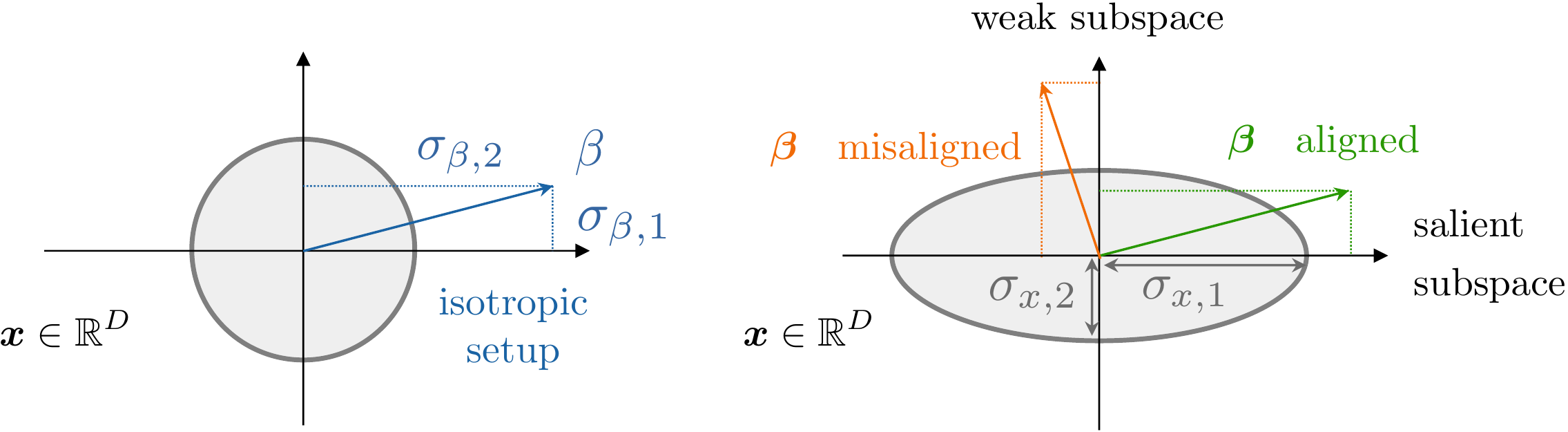}
    \caption{Strong and weak features model of data}
    \end{subfigure}    
    \caption{\textit{Left:} Random feature model considered here, which can be viewed as a two-layer networks where only the second layer is trained.
    \textit{Middle and Right:} Strong and weak features model considered here. Input space is decomposed into two subspaces with different variance in the anisotropic setting: a \emph{salient} one with strong variance $\sigma_{x,1}$ and a weak one with smaller variance $\sigma_{x,2}$. The labels are given by a linear teacher $y = {\rm sign}(\bs \beta \cdot \bs x / \sqrt{D})$ and flipped with a certain probability $\Delta$. We can adjust the \emph{alignment} of data subspaces with the teacher. 
    The task is easy when $\sigma_{\beta,1}>\sigma_{\beta,2}$, and hard in the opposite case.
    }
    \label{fig:RF}
\end{figure}

To satisfy \textit{(i)}, we need to disentangle the input dimension from the number of parameters in the learning model, which is impossible for the linear models often studied in litterature. We instead consider a random feature model~\citep{rahimi2008random}, which was shown to exhibit double descent by~\citet{mei2019generalization}. To satisfy \textit{(ii)}, we follow the lines of~\citep{gerace2020generalisation} and use an approach from statistical physics, enabling to generalize the result of~\citep{mei2019generalization} to any convex loss function. Finally, to satisfy \textit{(iii)}, we introduce a block model in which the input space is subdivided into various subspaces with different variances ({\it saliencies}) and different correlation ({\it alignments}) with the labels (see Fig. \ref{fig:RF}). This model was recently studied for linear regression under the name of \textit{strong and weak features model}~\citep{richards2020asymptotics}. 

Our analytical solution uses the replica method from statistical physics \cite{mezard1987spin}. Albeit non-rigorous in a strict mathematical sense, this method is considered exact in theoretical physics and
has been influential to statistical learning theory since the 80s, see \citet{gabrie2019mean} and \citet{bahri2020review} for recent reviews. 
Several of their conjectures have been proven exact in the last decade \citep{barbier2019optimal, aubin2018committe, gabrie2018entropy,  goldt2020gaussian}, including in settings very close to the considered model as discussed further at the end of Section \ref{sec:model-analytics} \citep{hu2020universality,Loureiro2021}. 

\paragraph{Contributions}
Our main analytical contribution, presented in Section \ref{sec:model-analytics}, is  the expression of the train and test errors of a random feature model trained on data generated by the strong and weak features model.
Our analysis is valid in the high-dimensional limit both for regression and classification tasks, although we focus here on the latter. We confirm that the asymptotic prediction matches simulations at finite size (see Fig. \ref{fig:agreement} and \ref{fig:agreement-train}).

We leverage this result to present a thorough analytical study of how the data structure and the loss function interplay to shape the generalization error, and in particular how they affect the double descent curve (Section \ref{sec:results}). We highlight behavioral differences between square loss and logistic loss, in particular the fact that logistic loss generalizes better for easy tasks (see Fig.~\ref{fig:anisotropy}). We validate our insights via controlled experiments on the MNIST and CIFAR10 datasets described in the Section \ref{sec:mnist}.

\paragraph{Reproducibility}
The code to reproduce our experiments is available at \url{https://github.com/sdascoli/data-structure}.

\section{Related Work}
Although most theoretical studies of generalization focus on structureless data, a few exceptions exist. The strong and weak features model has recently been studied in the context of least-squares regression, both empirically~\citep{nakkiran2020optimal} and theoretically~\citep{dobriban2015high,wu2020optimal,richards2020asymptotics}. Several intriguing observations emerge in strongly anisotropic setup: (i) several overfitting peaks can be seen ~\citep{nakkiran2020optimal,chen2020multiple}; (ii) the optimal ridge regularization parameter $\gamma$ can become negative~\citep{wu2020optimal,kobak2019optimal}, as it becomes helpful to encourage the weights to have very different magnitudes; (iii) extra features acting as pure noise can play a beneficial role by inducing some implicit regularization~\citep{richards2020asymptotics}.

The strong and weak features model also includes the setup studied in~\citet{ghorbani2020neural}, called the spiked covariates model. The latter involves a small block of size $D^{\eta}$ ($\eta < 1$) and a large ``junk'' block with no correlation with the labels. The question is then: can kernel methods learn to discard the junk features, hence ``beat the curse of dimensionality'' in the way neural networks do? The answer was shown to depend on the strength of the junk features: when the variance of these features is small, they are not problematic, and the kernel method ignores them, effectively learning a task of effective dimensionality $D^{\eta} \ll D$. 

More general models for the structure of data were considered in other works. In the special case of random Fourier Feature regression, \citet{liao2020random} derived the train and test error for a general input distribution. \citet{canatar2020statistical} achieved a similar result in the non-parametric setting of kernel regression, which can be viewed as the limiting case of random feature regression when the number of random features $P$ goes to infinity. 

The few works studying classification analytically have mostly focused on linear models, trained on linearly separable data~\citep{Soudry2018,montanari2019generalization,chatterji2020finite,liang2020precise} or gaussian mixtures~\citep{deng2019model,mignacco2020role}. One of the challenges in classification (in comparison to regression) is the large set of available loss functions~\citep{kini2020analytic,muthukumar2020classification}. In the context of random feature models, \citet{gerace2020generalisation} uses tools from statistical mechanics to derive the generalization loss of random features model for \textit{any} loss function with i.i.d. Gaussian input. More recently, \citet{Loureiro2021} shows that this framework could extend to more complex data distributions and learned feature maps provided that key population covariances are estimated by Monte Carlo methods. Our paper crucially builds on these contributions, by deriving a fully analytical analysis for a simple interpretable model of data structure, while also analyzing the effect of label flipping.

We also note that a few recent works investigate the roles of the loss and the structure in data in
 more realistic setups where theoretically robust results are harder to obtain. Several works have studied the low intrinsic dimensionality of real-world data distributions and how it impacts sample complexity for supervised tasks~\cite{spigler2019asymptotic,ansuini2019intrinsic,pope2021intrinsic,baratin2020implicit}. The impact of the loss function is also an active research area: \cite{hui2020evaluation,demirkaya2020exploring} show that square loss can perform equally or better than the ubiquitous cross-entropy loss in realistic multi-class classification problems, if one rescales the weight of the correct class to emphasize its importance. 


\section{A solvable model of data structure}
\label{sec:model-analytics}

In this work, we focus on the random features model\footnote{Note that this model is akin to the so-called lazy learning regime of neural networks where the weights stay close to their initial value~\citep{jacot2018neural}: assuming $f_{\theta_0}\!=\!0$, we have $f_{\theta}({\bf x}) \approx \nabla_{\theta}f_{\theta}({\bf x})|_{\theta\!=\!\theta_0} \cdot (\theta-\theta_0)$~\citep{chizat2018note}. In other words, lazy learning amounts to a linear fitting problem with a random feature vector $\left.\nabla_{\theta}f_{\theta}({\bf x})\right|_{\theta\!=\!\theta_0}$.}  introduced in \citet{rahimi2008random}.
Using tools from statistical physics, we derive the generalization error of a teacher-student task on the strong and weak features model of structured data. 

\paragraph{Random feature model}
The random features model can be viewed as a two-layer neural network (see Fig.~\ref{fig:RF}) whose first layer is a fixed random matrix containing $P$ random feature vectors $\{\bs F_i\in\mathbb R^D\}_{i=1\ldots P}$ acting on inputs $\bs x_\mu \in \mathbb R^D$:
\begin{equation}
    \hat y_\mu = \sum_{i=1}^P w_i \sigma\left(\frac{\bs F_i \cdot \bs x_\mu}{\sqrt{D}}\right),
\end{equation} 
where $\sigma(\cdot)$ is a pointwise activation function and $w_i\in \mathbb{R}$ are the second layer weights.
Elements of $\bs F$ are drawn i.i.d from $\mathcal{N}(0,1)$.  

The second layer weights, i.e. the elements of $\bs w \in \sR^P$, are trained by minimizing an $\ell_2$-regularized loss on $N$ training examples $\{\bs x_\mu\in\R^D\}_{\mu=1\ldots N}$ :
\begin{align}
    \label{eq:ridge-estimator}
    \hat{\boldsymbol{w}}=\underset{\boldsymbol{w}}{\operatorname{argmin}} \left[\epsilon_t(\bs w) +\frac{\lambda}{2}\|\boldsymbol{w}\|_{2}^{2} \right],\quad\quad\quad
    \epsilon_t(\bs w) = \sum_{\mu=1}^{N} \ell\left(y_{\mu}, \hat y_\mu \right),
\end{align}
where $\ell$ denotes the logistic loss $\ell(y,\hat y)=\log(1+e^{-y\hat y})$. The target labels are given by a probabilistic teacher $y\sim \mathcal{P}_t(y|\bs \beta \cdot \bs x)$ corresponding to the sign of a linear function possibly corrupted by label flipping:
\begin{align}
    y_\mu = \eta_\mu\ \text{sign}\left(\frac{\bs \beta \cdot \bs x_\mu}{\sqrt D}\right), \quad\quad\quad \eta_\mu = \begin{cases}
    1  &\text{ with probability }1-\Delta\\
    -1 &\text{ with probability }\Delta.
    \end{cases}
\end{align}

The generalization error is computed as the 0-1 loss,
\begin{align}
    \epsilon_{g}= \E_{\boldsymbol{x}, y}\left[\mathbbm{1}_{\operatorname{sign}\left(\hat{y}(\boldsymbol{x})\right),y(\boldsymbol x)}\right].
\end{align}

\paragraph{Strong and weak features model}
To impose structure on the input space,
we introduce a block-structured covariance matrix from which the elements of the inputs $\bs x\in \mathbb R^D$ and the teacher $\bs \beta \in \sR^D$ are sampled as:
\begin{align*}
&\bs x\sim\mathcal{N}(0, \Sigma_x),  &&\Sigma_x = \begin{bmatrix}
\sigx{1} \mathbb I_{\phi_1 D} & 0 & 0\\
0 & \sigx{2} \mathbb I_{\phi_2 D} & 0\\
0 & 0 & \ddots
\end{bmatrix},\\
&\bs \beta \sim\mathcal{N}(0, \Sigma_\beta),  &&\Sigma_\beta = \begin{bmatrix}
\sigb{1} \mathbb I_{\phi_1 D} & 0 & 0 \\
0 & \sigb{2} \mathbb I_{\phi_2 D} & 0\\
0 & 0 & \ddots
\end{bmatrix}.
\end{align*}

Our result presented in the rest of Section \ref{sec:model-analytics} is valid for an arbitrary number of blocks. In Section \ref{sec:results} we will focus for interpretability on the special case where we only have two blocks of sizes $\phi_1 D$ and $\phi_2 D$, with $\phi_1 + \phi_2 = 1$. We will typically be interested in the strongly anisotropic setup where the first subspace is much smaller ($\phi_1 \ll 1$), but potentially has higher \emph{saliency} $r_x = \sigx{1}/ \sigx{2} \gg 1$ (see Fig.~\ref{fig:RF}). 

\paragraph{Main analytical result}
Using the replica method from statistical physics~\citep{mezard1987spin} and the Gaussian Equivalence Theorem (GET)~\citep{el2010spectrum,mei2019generalization,goldt2020gaussian,hu2020universality}, we derive the generalization and training errors in the high-dimensional limit where $D, N$ and $P\to\infty$ with fixed ratios.
The asymptotic generalization and training errors are given by 
\begin{align}
    \lim _{N \rightarrow \infty} \epsilon_{g}& = \frac{1}{\pi} \cos ^{-1}\left(\frac{M}{\sqrt{\rho Q}}\right),
    \label{eq:test-error}\\
    \lim _{N \rightarrow \infty} \epsilon_{t} & = \E_\xi \int_{\mathbb{R}} dy\left[ \mathcal{Z}^0\left(y , \xi M / \sqrt Q, \rho - M^2/ Q\right) \ell(y, \eta(y, \sqrt Q \xi, V)\right]
\end{align}
with the proximal operator $\eta(y, a, b) =\underset{x}{\argmin} (x - a)^2/(2b) + \ell(x,y) $, the random variable $\xi \sim \mathcal{N}(0,1)$, the functional 
$\mathcal{Z}^0\left(y, a, b\right) = \int_{\mathbb{R}}dx\; \mathcal{N}(x; a, b) \mathcal{P}_t(y|x)$ and
the scalars
\begin{align}
\label{eq:replica-M-Q}
&\rho = \sum_i \phi_i \sigb{i} \sigx{i}, \quad M= \kappa_{1} \sum_i \sigx{i} m_{s,i},
\quad Q = \kappa_{1}^2 \sum_i \sigx{i} q_{s,i} +  \kappa_{\star}^{2} q_w, \quad V = \kappa_1^2 \sum_i \sigx{i} v_{s,i} + \kappa_\star^2 v_w 
. \notag
\end{align}
The parameters $\kappa_1, \kappa_\star$ are related to the activation function: denoting $r=\sum_i \phi_i \sigx{i}$ and $\xi \sim \mathcal{N}(0,r)$, one has
\begin{align}
\kappa_{1}=\frac{1}{r}\E_\xi[\xi \sigma(\xi)]], \quad &\kappa_{\star}=\sqrt{\E_\xi\left[\sigma(\xi)^{2}\right] - r \kappa_{1}^{2}}.
\end{align}
Besides these constants and the ones defining the data structure ($\phi_i ,\sigb{i} ,\sigx{i}$), the key ingredients to obtain asymptotic errors are the so-called \emph{order parameters} $m_s, q_s, q_w, v_s$ and $v_w$. They correspond to the high-dimensional limit of the following expectations and variances (denoted by $\sV$):
\begin{align*}
m_{s,i} &=\lim_{D\to\infty}\frac{1}{D} \E_\mathcal{P}\left[\bs{s}_i \cdot \bs\beta_i\right],\quad
q_{s,i} = \lim_{D\to\infty}\frac{1}{D} \E_\mathcal{P}\left[\bs{s}_i \cdot \bs{s}_i\right],\quad
q_{w} = \lim_{P\to\infty}\frac{1}{P} \E_\mathcal{P}\left[ \hat{\bs{w}} \cdot \hat{\bs{w}}\right], \\
v_{s,i} &=\lim_{D\to\infty} \frac{1}{D} \sV_\mathcal{P} \left[\bs{s}_i \cdot \bs{s}_i\right], \quad
v_{w}=\lim_{P\to\infty} \frac{1}{P} \sV_\mathcal{P} \left[ \hat{\bs{w}} \cdot \hat{\bs{w}}\right],
\end{align*}
where $\bs{s}=\frac{1}{\sqrt{P}} \bs F \hat{\bs{w}} \in \sR^{D}$ and $\bs s_i, \bs \beta_i \in \sR^{\phi_i D}$ denote the orthogonal projections of $\bs s$ and $\bs \beta$ onto subspace $i\in\{1,2\}$ and $\mathcal{P}$ denotes the joint distribution of all random quantities in the problem (the teacher weights, the random features and the training data). 

Intuitively, $\rho$ is the variance of the outputs of the teacher, $Q$ is the variance of the outputs of the student, and $M$ is their covariance. The generalization error is given by the ``angle'' between the teacher and the student, as expressed by Eq.~\ref{eq:test-error}. The order parameters allowing to obtain $Q$ and $M$ are one of the outputs of the replica computation deferred to SM~\ref{app:replica}. They are obtained by solving a set of non-linear saddle-point equations (see Section \ref{app:saddle-point} of the SM).
Our framework is valid for any convex loss function, although the replica equations need to be evaluated numerically in the general case. In the case of the square loss however, some simplifications arise, e.g. an explicit expression for the training error can be obtained from the order parameters (see Section \ref{app:train-derivation} of the SM). 

\paragraph{Steps and validity of the replica analysis}
The necessary steps of the derivation are detailed in SM~\ref{app:replica}. In particular, (i) we obtain an anistropic extension of the GET, (ii) conduct random matrix analysis for block matrices and finally (iii) derive the analytical saddle-point equations which yield the values of the order parameters. Our result generalizes the strategy of~\citet{gerace2020generalisation} from isotropic to anisotropic data and additionally covers the effect of label flipping. Our result is also related to \citet{Loureiro2021} which establishes rigorously the replica prediction in related learning problems. A rigorous proof of our replica results is within reach: it requires a small extension of \cite{hu2020universality} (to prove the anisotropic GET derived in Section \ref{app:get} of the SM) combined with the recent results of \citet{Loureiro2021}. 
Moreover, results in the following section show perfect agreement with numerical experiments.
%

\section{Effect of data structure and loss function on double descent}
\label{sec:results}

\begin{figure}[tb]
    \centering
    \begin{subfigure}[b]{.49\textwidth}	   
    \includegraphics[width=\linewidth]{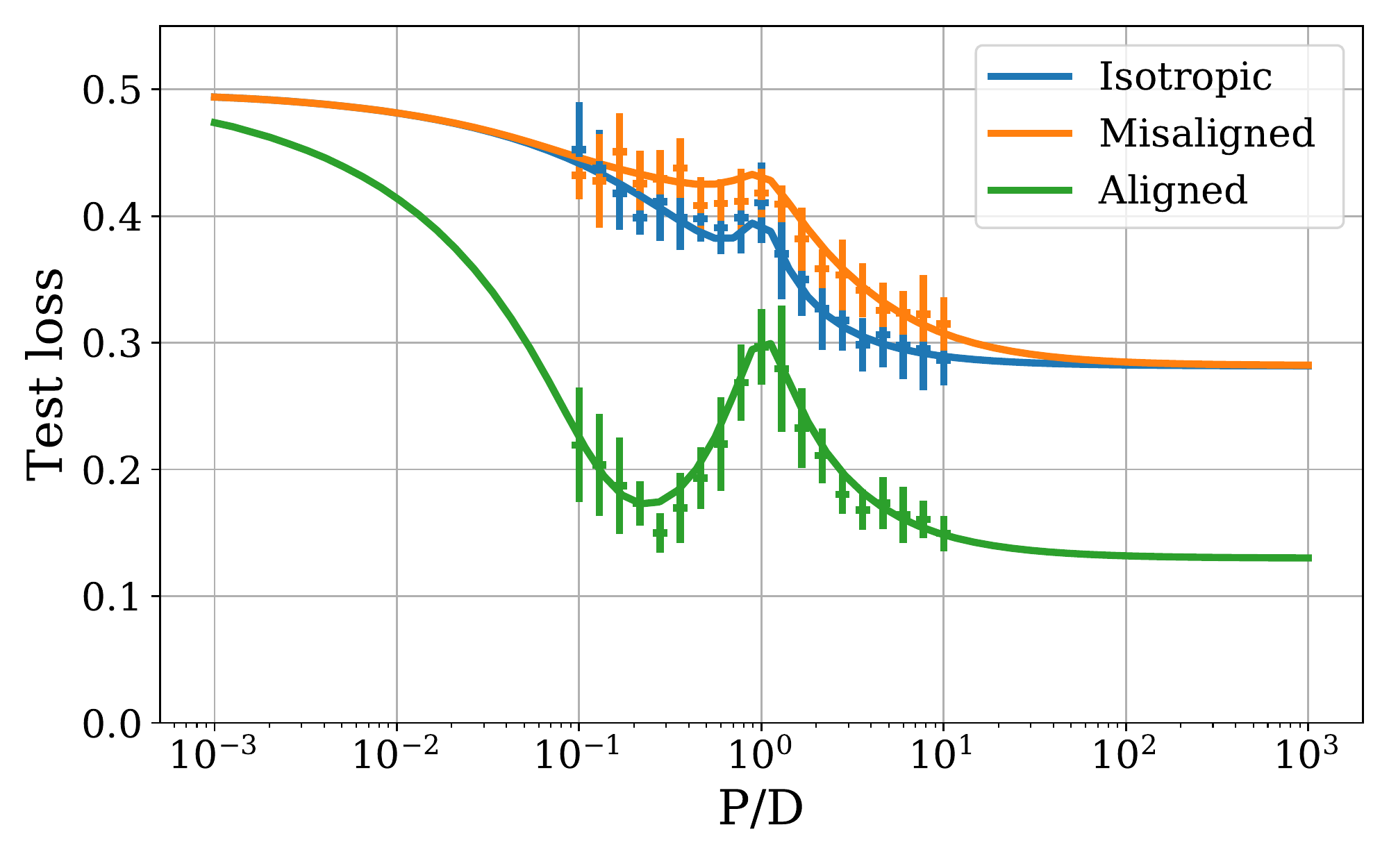}
    \caption{Square loss, true labels}
    \end{subfigure}
    \begin{subfigure}[b]{.49\textwidth}	   
    \includegraphics[width=\linewidth]{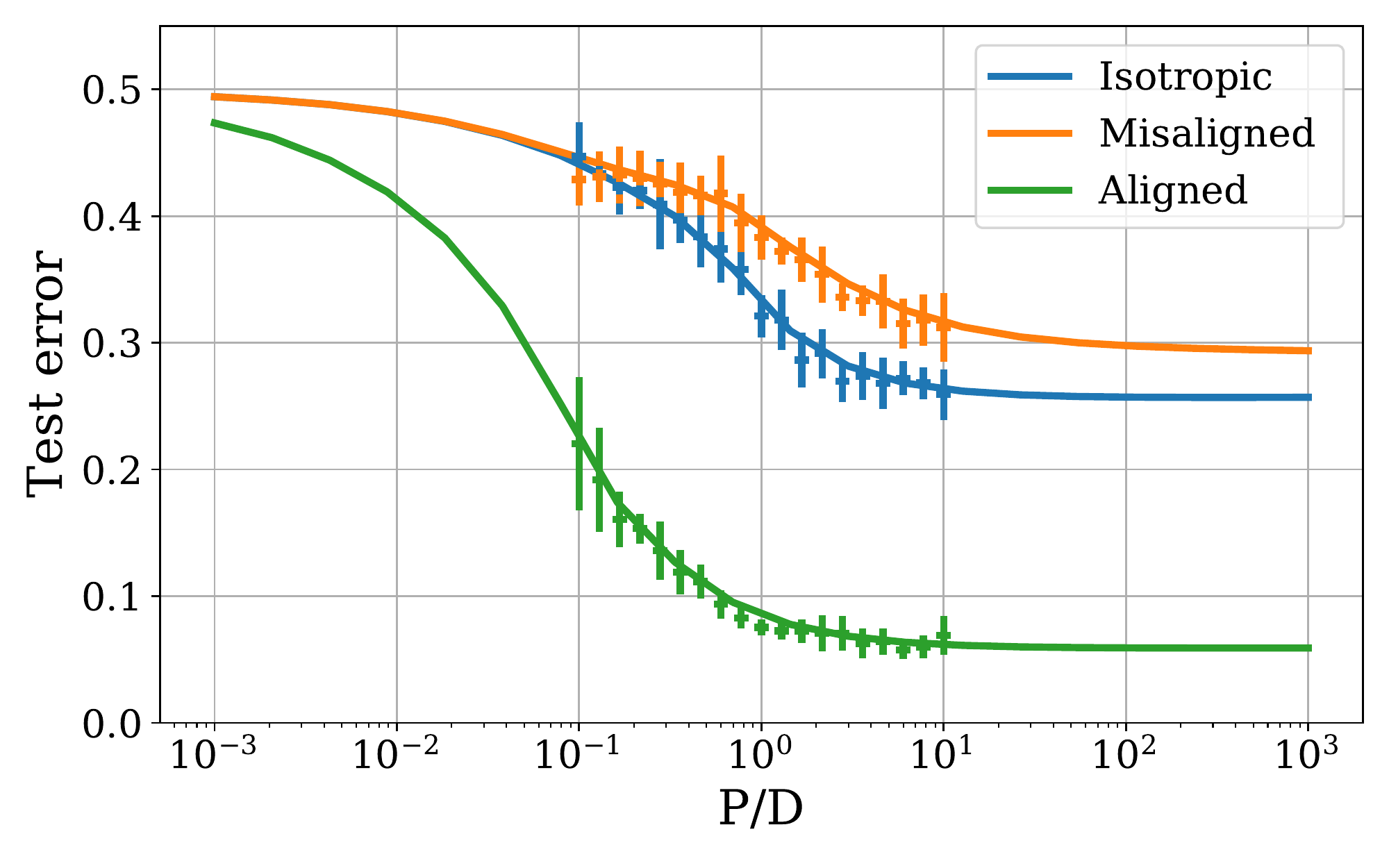}
    \caption{Logistic loss, true labels}
    \end{subfigure}
    \begin{subfigure}[b]{.49\textwidth}	   
    \includegraphics[width=\linewidth]{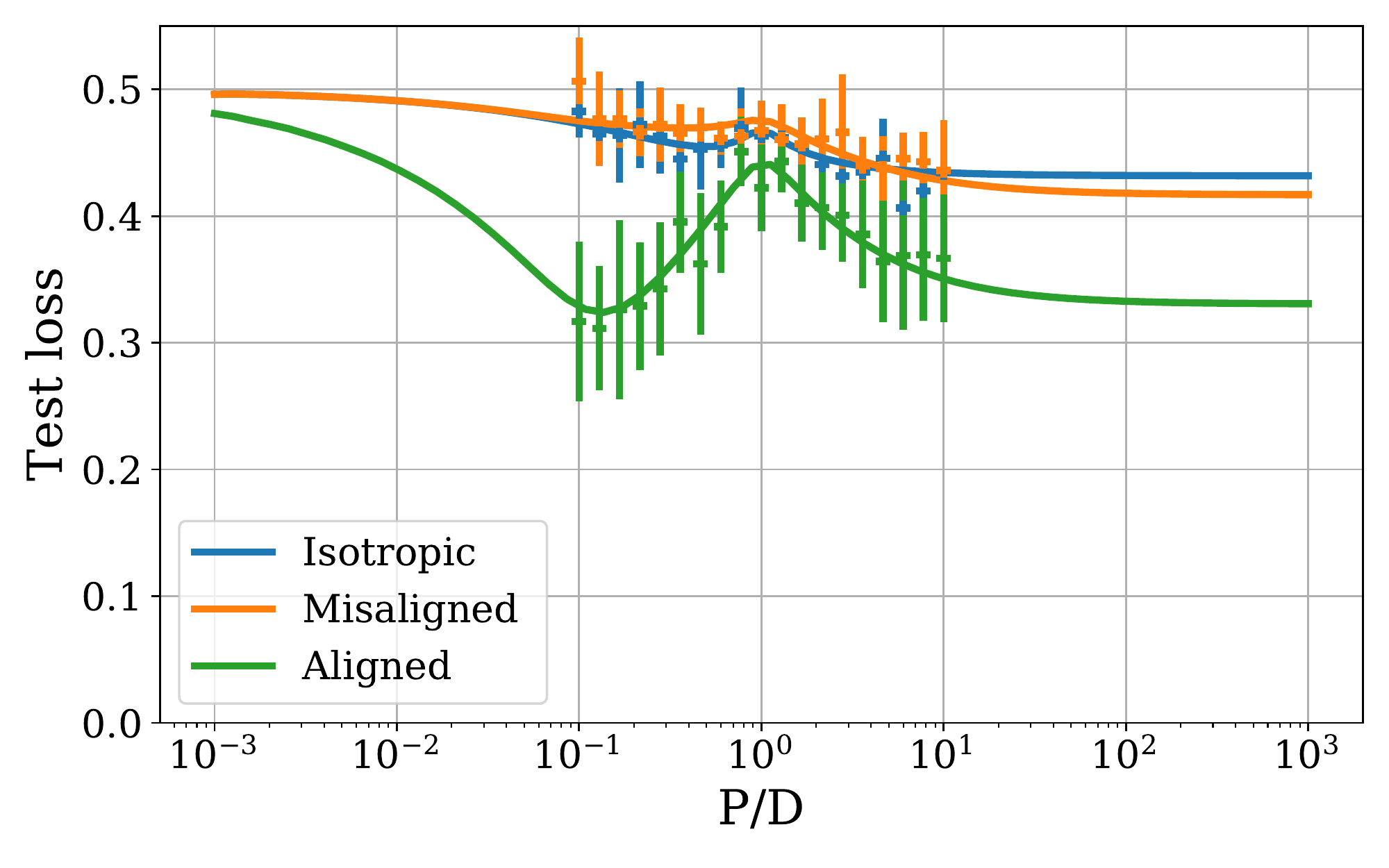}
    \caption{Square loss, label flip probability $\Delta=0.3$}
    \end{subfigure}
    \begin{subfigure}[b]{.49\textwidth}	   
    \includegraphics[width=\linewidth]{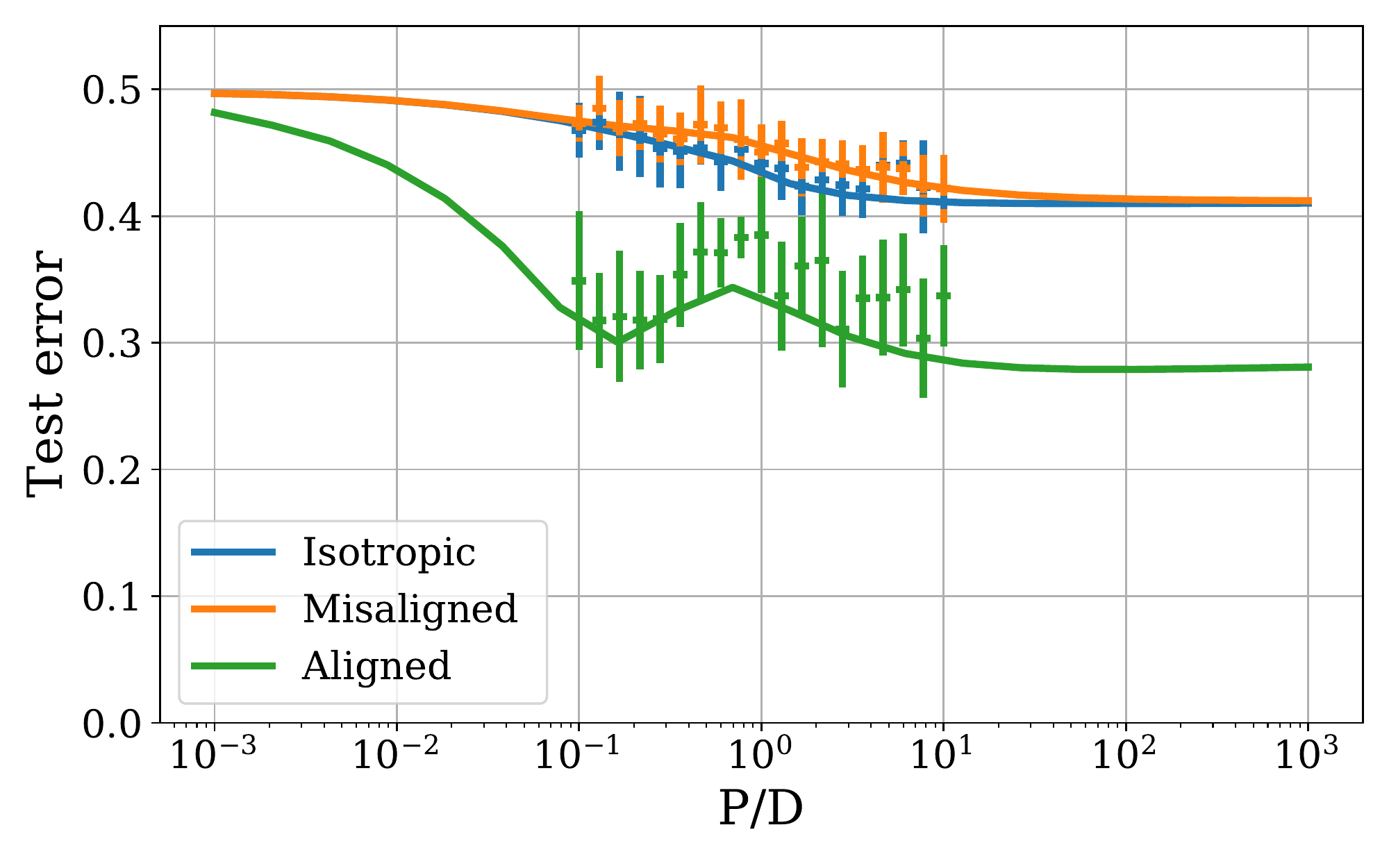}
    \caption{Logistic loss, label flip probability $\Delta=0.3$}
    \end{subfigure}
    \caption{\textbf{Anisotropic data strongly affects the double descent curves.} Theoretical results (solid curves) and numerical results (dots with vertical bars denoting standard deviation over 10 runs) agree even at moderate size $D=100$. We set $\sigma = \mathrm{Tanh}$, $\lambda=10^{-3}$ and $N/D=1$.}
    \label{fig:agreement}
\end{figure}

\begin{figure}[tb]
    \centering
    \begin{subfigure}[b]{.49\linewidth}
    \includegraphics[width=\linewidth]{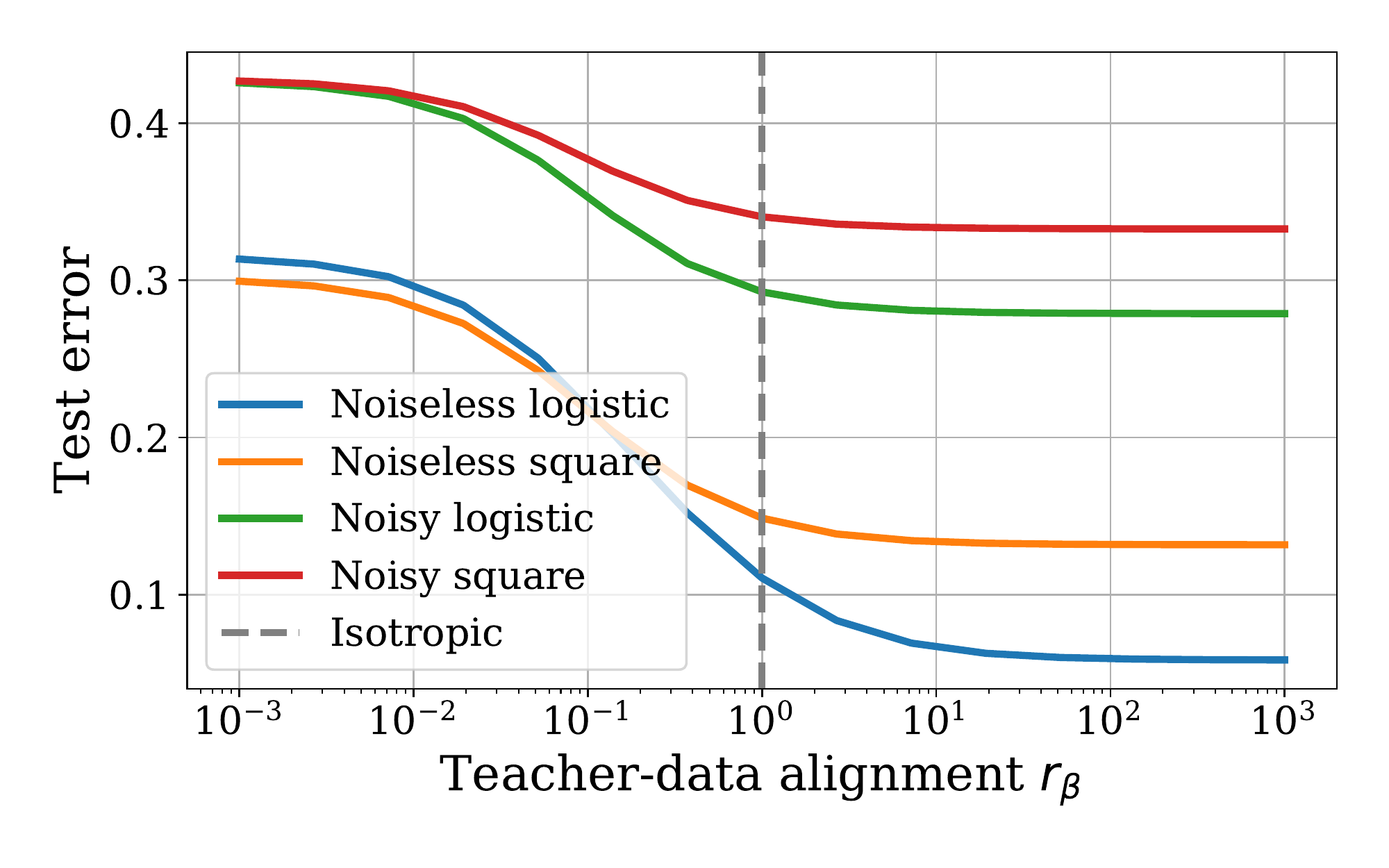}    
    \caption{Strong and weak features model}
    \end{subfigure}
    \begin{subfigure}[b]{.49\linewidth}
    \includegraphics[width=\linewidth]{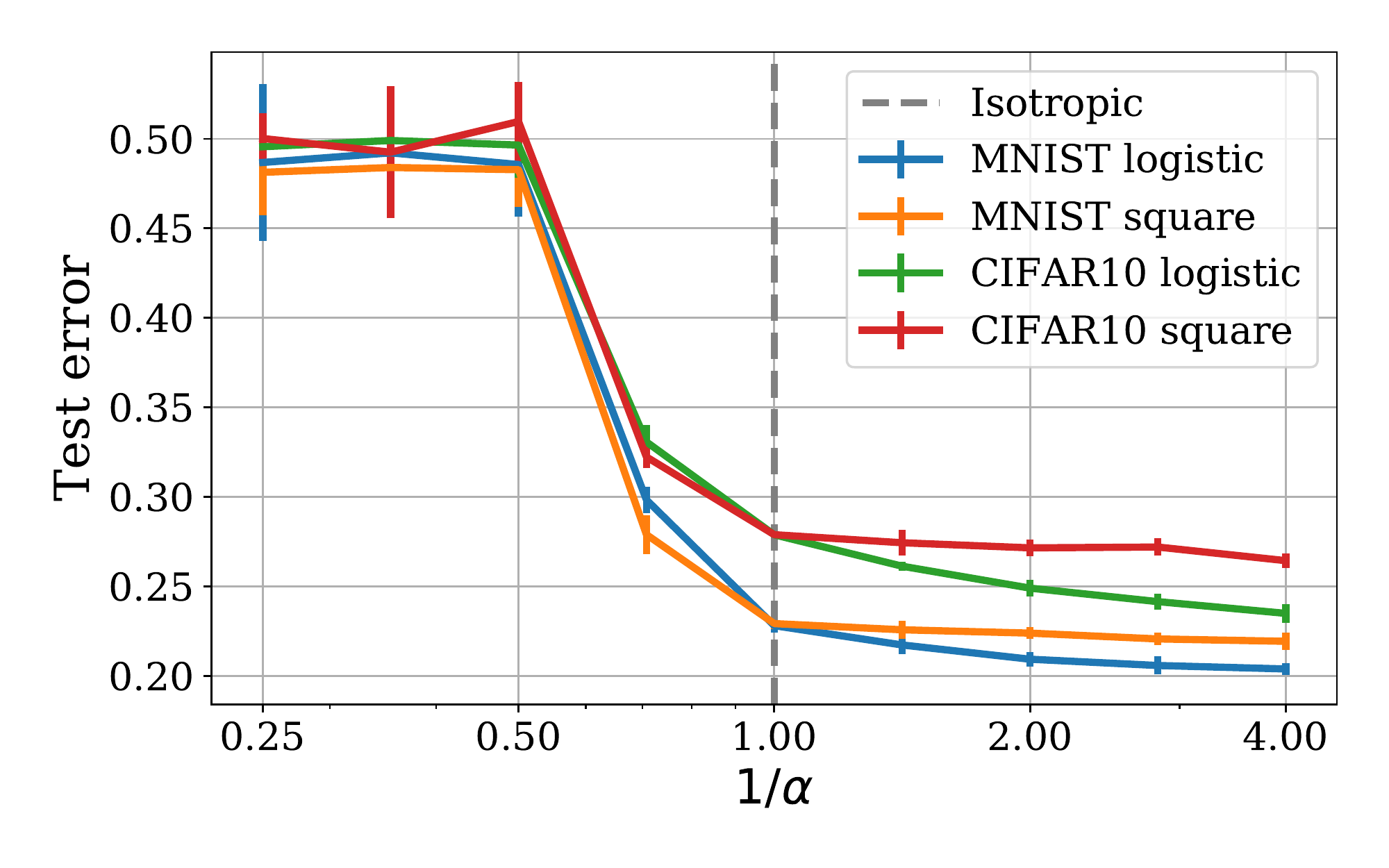}
    \caption{Real data}
    \end{subfigure}
    \caption{\textbf{The easier the task, the wider the gap between logistic and square loss.} (a) Strong and weak feature model in the noiseless ($\Delta=0$) and noisy ($\Delta=0.3$) setups, where we make the task easier from left to right by increasing the alignment between the data and the teacher. (b) Real data (MNIST parity and CIFAR10 airplanes vs cars), where we make the task easier by decreasing the exponent $\alpha$ controlling the saliency of the top PCA components (see Sec.~\ref{sec:mnist}). In both cases, we considered an over-parametrized RF model ($P/D=100$) learning from a moderate amount of data ($N/D=1$), with $\sigma=\mathrm{Tanh}$ and $\lambda=10^{-4}$. }
    \label{fig:anisotropy}
\end{figure}

In this section, we investigate how the interplay between the data structure and the loss function shapes the train and test error curves. In the main text, we only examine parameter-wise curves, where we increase the number of parameters $P$ at fixed number of data $N$. In SM.~\ref{app:phase-space}, we also examine the sample-wise dependency by plotting the train and test error in the entire $(N,P)$ phase space.

\paragraph{Modulating the teacher-data alignment}

We compare three cases illustrated on Fig. \ref{fig:RF}. The first is the \textbf{isotropic} setup where $r_x=1$ (blue curves in Fig.~\ref{fig:agreement}). In the two next setups, the data is anisotropic with a small subspace ($\phi_1=0.1$) of large variance and a large subspace ($\phi_2=0.9$) of small variance. The ratio of the variances $r_x= \sigma_{x,1} / \sigma_{x,2}$ is set to 10, and their values are chosen to keep the total variance of the inputs unchanged: $\frac{1}{D}\E[ \Vert \bs x \Vert^2]= \phi_1 \sigma_{x,1} + \phi_2 \sigma_{x,2}=1$.

We study two cases for the outputs: (i) the \textbf{aligned} scenario where the strong features are highly correlated with the labels ($r_\beta = \sigma_{\beta,1} / \sigma_{\beta,2} =100$, green curves in Fig.~\ref{fig:agreement}); (ii) the \textbf{misaligned} scenario, where the strong features have low correlation with the labels ($r_\beta=0.01$, orange curves in Fig.~\ref{fig:agreement}). In both cases, we choose the $\sigma_{\beta,i}$ such that the total variance of the teacher scores is unchanged: $\frac{1}{D}\E [(\bs \beta \cdot \bs x)^2] = \phi_1 \sigma_{x,1} \sigma_{\beta,1} + \phi_2 \sigma_{x,2}\sigma_{\beta,2}=1$.

\begin{figure}[tb]
    \centering
    \begin{subfigure}[b]{.49\textwidth}	   
    \includegraphics[width=\linewidth]{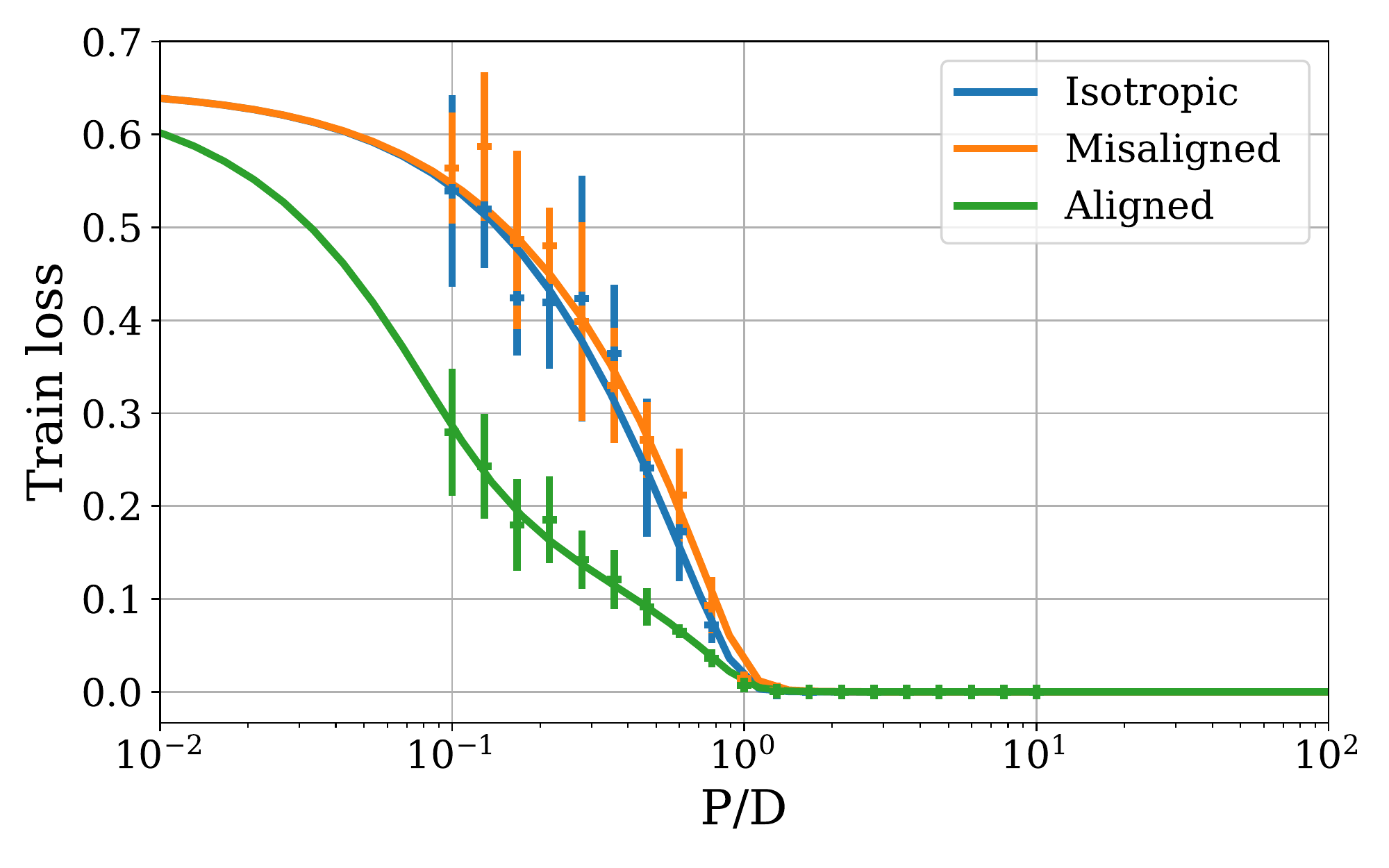}
    \caption{Square loss}
    \end{subfigure}
    \begin{subfigure}[b]{.49\textwidth}	   
    \includegraphics[width=\linewidth]{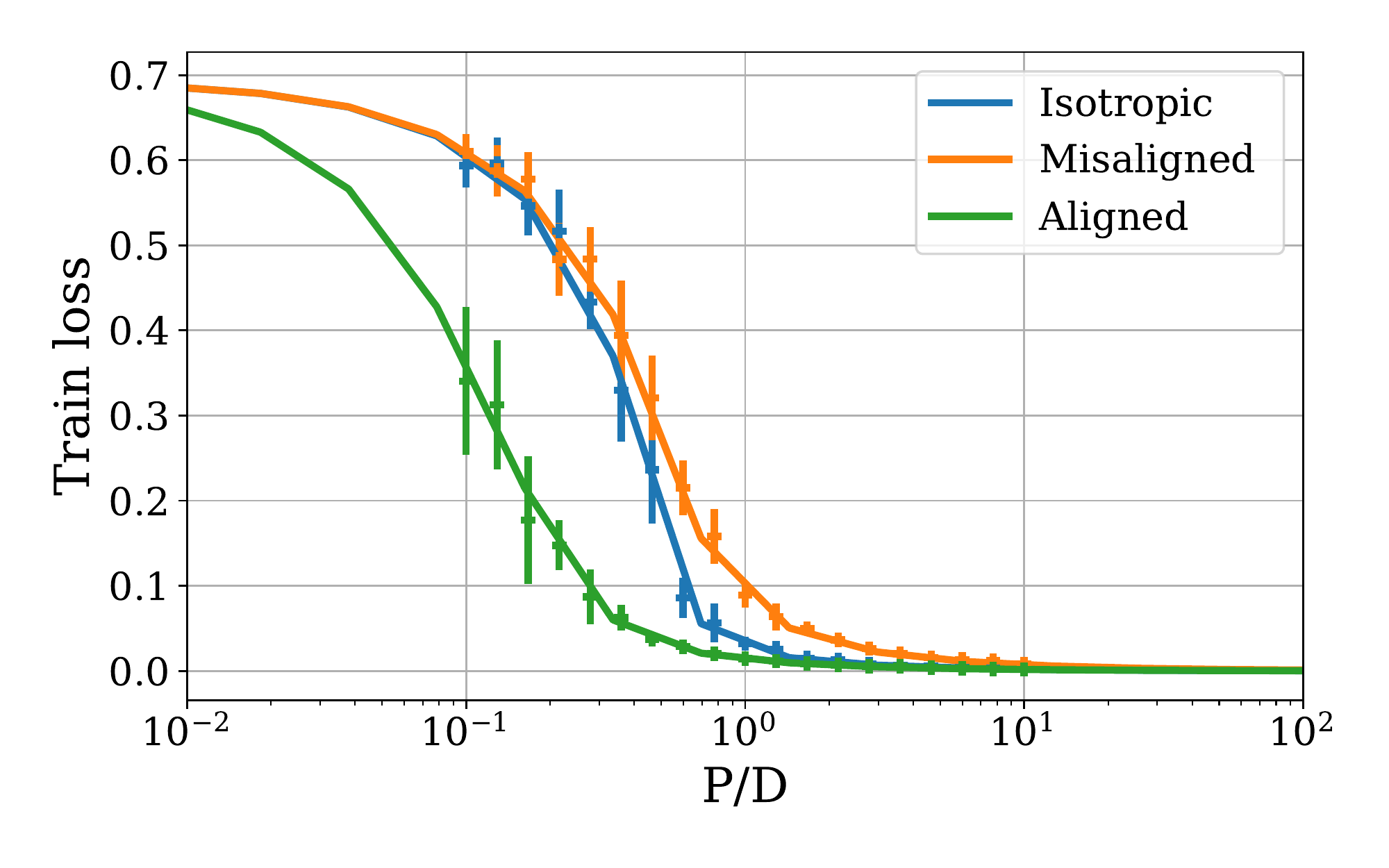}
    \caption{Logistic loss}
    \end{subfigure}
    \caption{\textbf{Structure of data affects the position of the interpolation threshold for logistic loss.} We depicted the train loss curves in the noiseless setup of Fig.~\ref{fig:agreement}, where $\sigma = \mathrm{Tanh}$, $\lambda=10^{-3}$, $N/D=1$.
    }
    \label{fig:agreement-train}
\end{figure}

\paragraph{Validity at finite size}
We begin by comparing our analytical predictions with the outcomes of numerical experiments, both for test loss (Fig.~\ref{fig:agreement}) and  train loss (Fig.~\ref{fig:agreement-train}). The agreement is excellent even for moderately large dimensions $D=100$. Note that the replica method, which relies on solving a set of scalar fixed point equations, is also computationally efficient. It allows here to probe ratios of $P/D$ and $N/D$ far beyond what is tractable by the numerics ($P$, $D$ and $N$ only appear in the replica equations through the values of the ratios $N/P$ and $P/D$). 

\paragraph{Effect of data structure on generalization}
Looking at Fig.~\ref{fig:agreement}, a first immediate observation is that strong teacher-data alignment makes the task easier: as number of parameters $P$ increases the test loss drops earlier and eventually reaches a lower asymptotic value, both for square loss and logistic loss. In SM.~\ref{app:phase-space}, we show that the same phenomenon occurs when varying the number of samples $N$ instead of the number of parameters $P$. These observations are in line with the results of~\citet{ghorbani2020neural} and show that relevant salient features make the problem low-dimensional with an effective dimension close to $\phi_1D$. This setup is the most akin to real-world tasks in which the most salient features of an image are often the most relevant to its recognition. In this sense, the impressive performance of kernel methods such as the Convolutional NTK on real-world datasets~\citep{arora2019harnessing} can be associated with the anisotropy of the data: feature learning is not indispensable to beat the curse of dimensionality if the irrelevant features are weakly salient to begin with~\cite{baratin2020implicit}.

Conversely, misalignment generally makes the task harder and increases the value of the test loss. Note however that for square loss, an interesting crossover occurs in presence of noise (panel c): the irrelevant features are detrimental from small $P$, but become helpful at large $P$, as can be seen from the orange curve reaching a lower asymptotic value than the blue curve. We associate this to the phenomenon discovered for linear regression in~\citet{richards2020asymptotics}, whereby adding noisy features acts as a form of implicit regularization. 

\paragraph{Logistic is better than square loss}

Comparing the two loss functions, we observe two beneficial effects of using logistic loss rather than square loss.

First, we observe that the overfitting peak characteristic of the double descent curve which appears at $P=D$ for square loss is absent for logistic loss in the noiseless setting~(Fig.~\ref{fig:agreement}), and vastly reduced in presence of noise (we use in both cases the same small amount of regularization for square and logistic loss). This suggests that the logistic loss exerts some form of implicit regularization, reducing the amount of overfitting. 

Second, in the aligned and isotropic setups, the asymptotic test loss reached in the ``kernel'' regime $P/D\to\infty$ is lower with logistic loss, especially in the aligned setup. To better highlight this phenomenon, we continuously vary the teacher-data alignment in Fig.~\ref{fig:anisotropy}(a) for an overparametrized model. Logistic loss performs similarly or worse than square loss at very small alignment, but outperforms square loss as soon as the alignment is sufficient. The gap between the two then grows as we increase alignment. In other words, logistic loss is particularly powerful on tasks made easy by the structure in the data. 

The better ability of logistic loss to detect structure in the data is also reflected in the train loss curves of Fig.~\ref{fig:agreement-train}. For square loss, the interpolation threshold, i.e. the point when the the train loss vanishes, occurs at $P=N$. For logistic loss, there is no interpolation threshold strictly speaking since the train loss cannot be zero. However, one can define an effective threshold as the point where the training loss reaches the near-zero plateau. Notably, this effective threshold depends on the data structure: it is lower for the aligned setup, where the data is easier to fit, and higher for the misaligned setup, where the data is harder to fit. 

\begin{figure}[tb]
    \centering
    \begin{subfigure}[b]{.49\textwidth}	   
    \includegraphics[width=\linewidth]{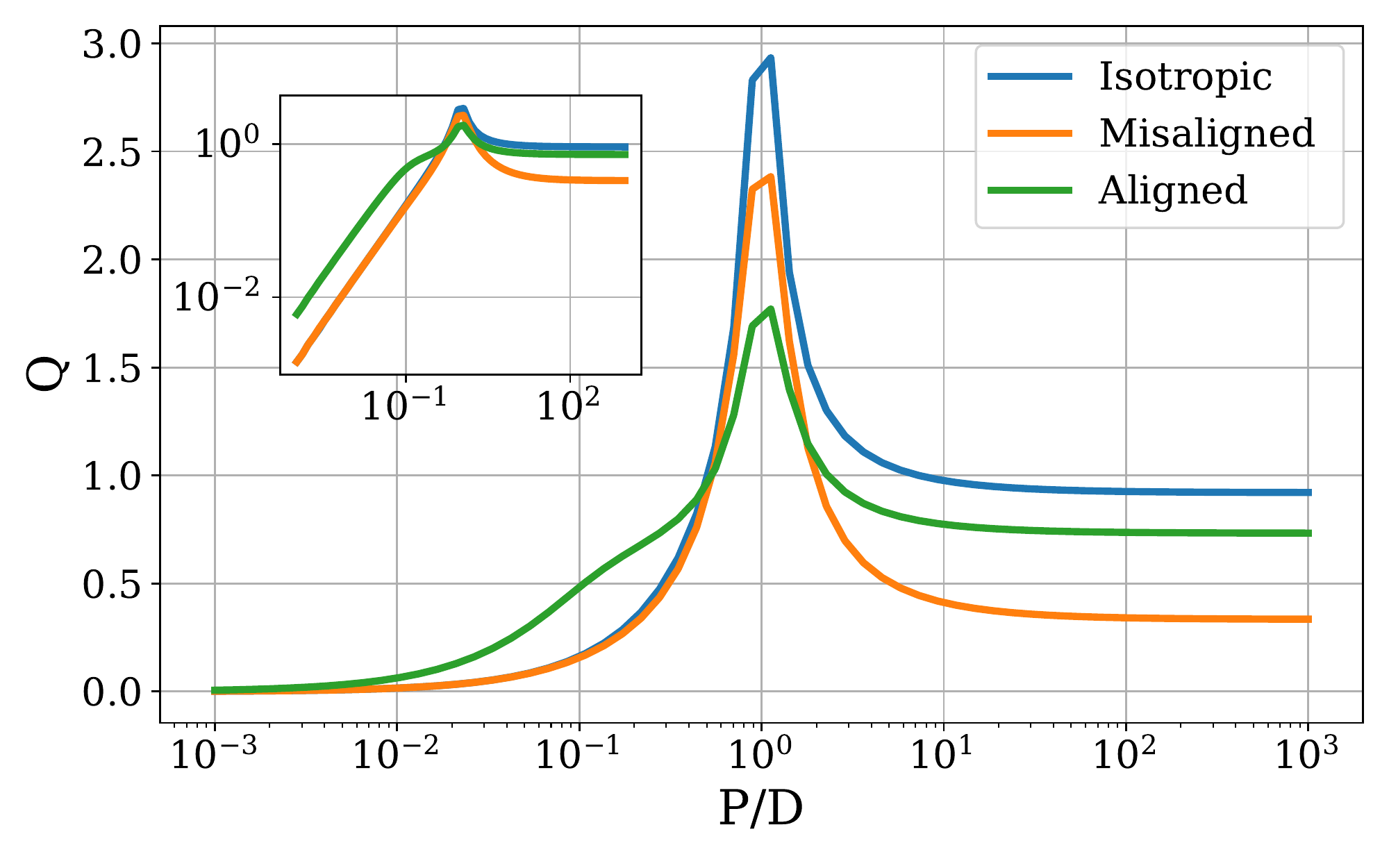}
    \includegraphics[width=\linewidth]{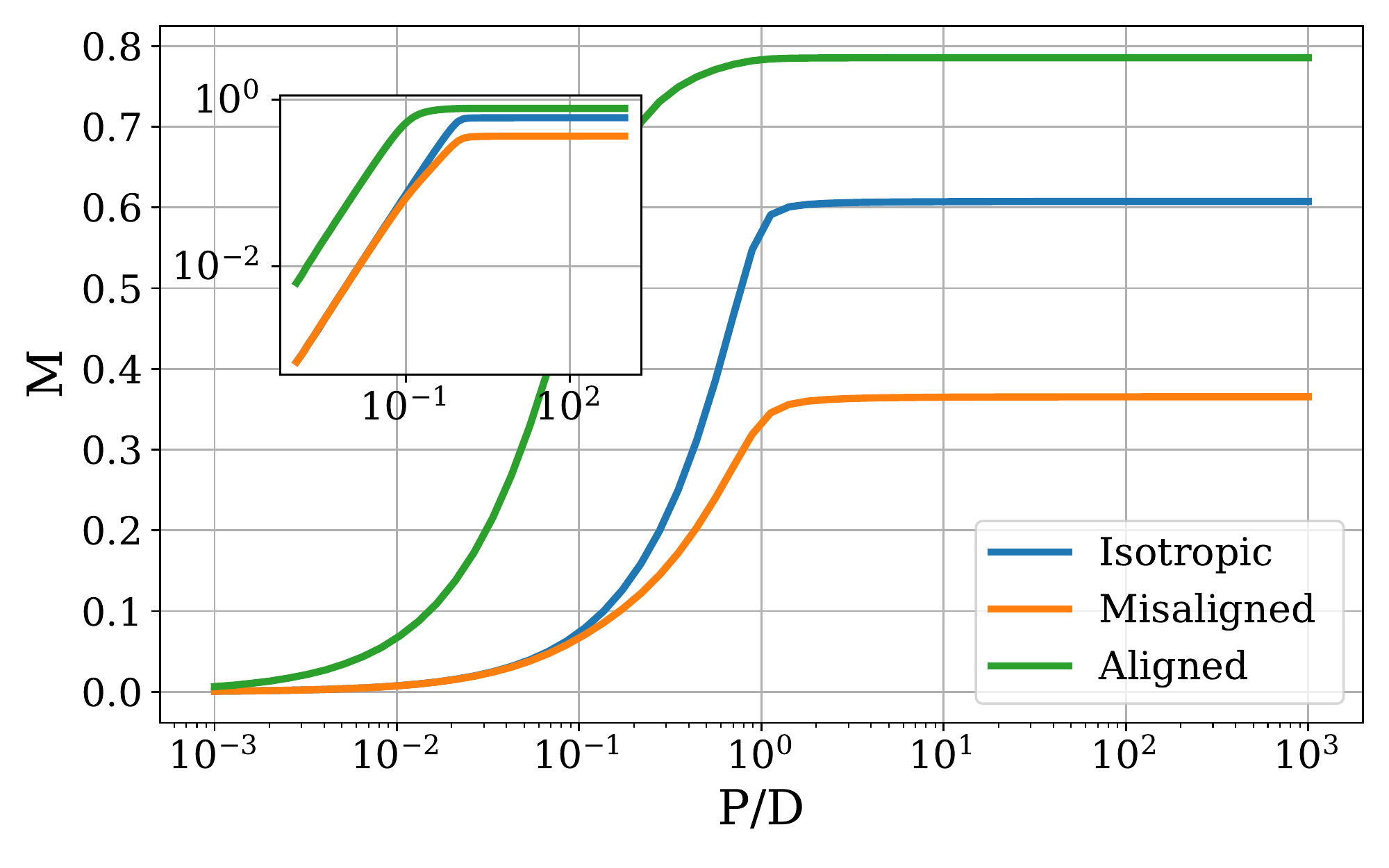}
    \caption{Square loss}
    \end{subfigure}
    \begin{subfigure}[b]{.49\textwidth}	   
    \includegraphics[width=\linewidth]{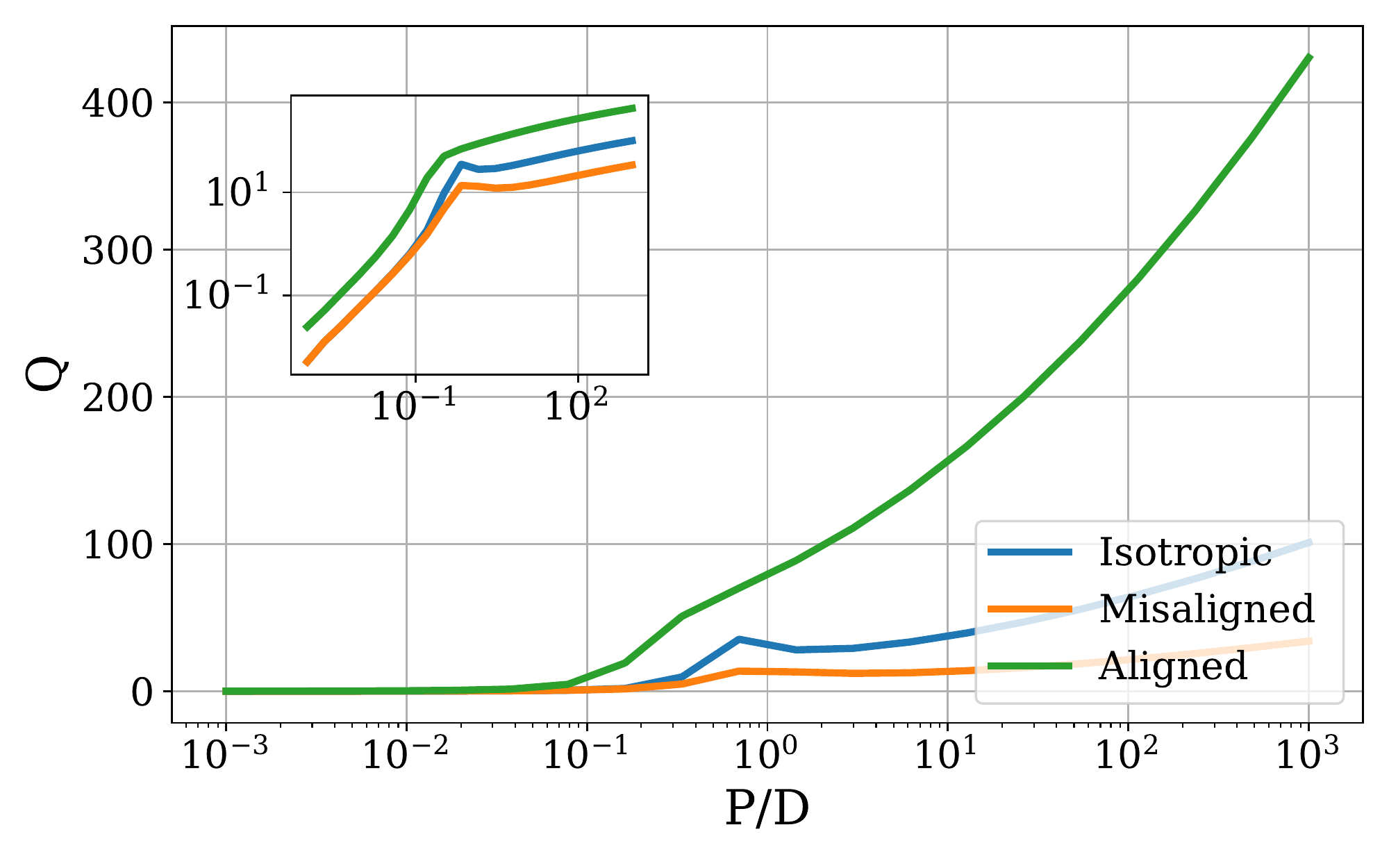}
    \includegraphics[width=\linewidth]{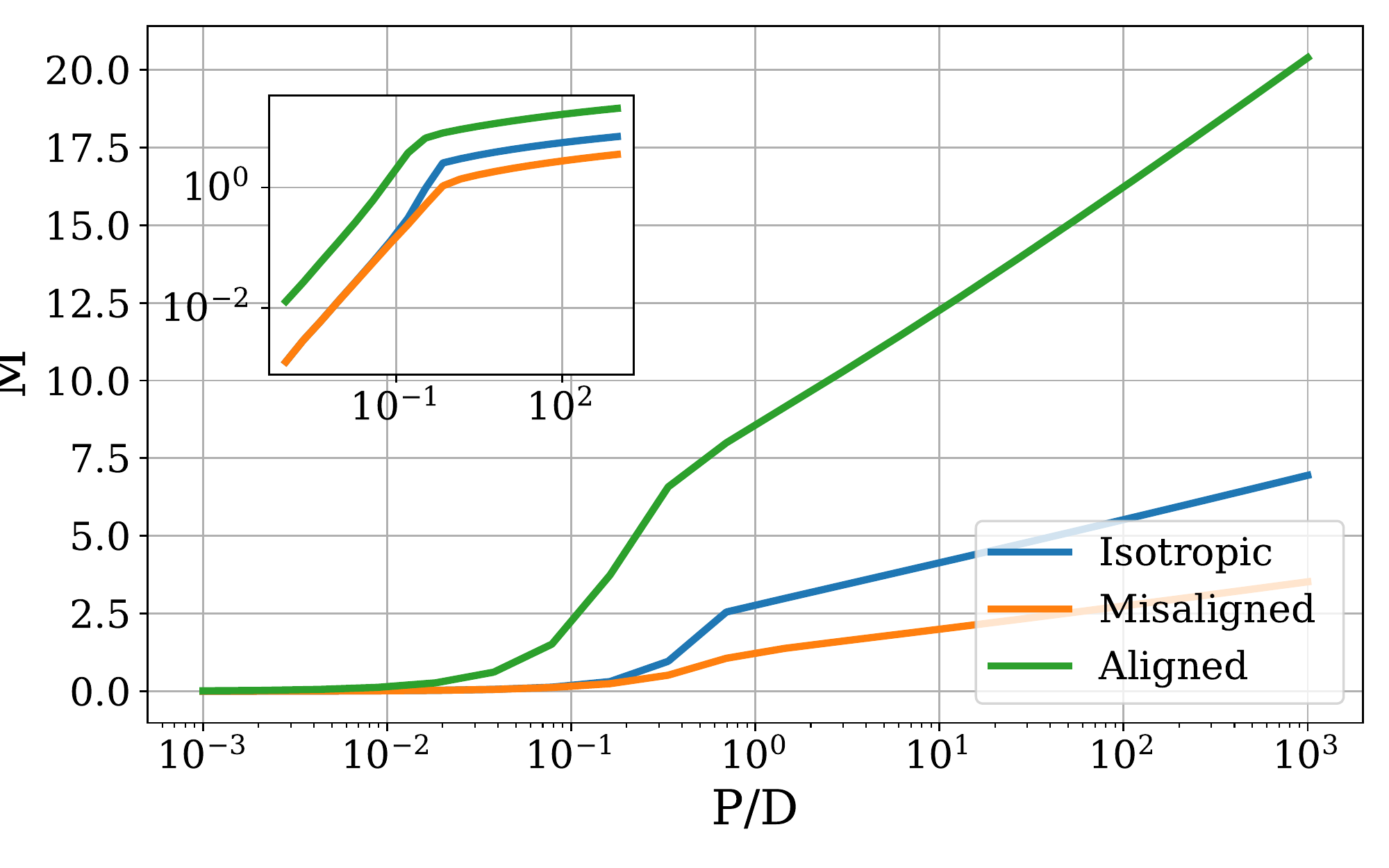}
    \caption{Logistic loss}
    \end{subfigure}
    \caption{\textbf{Overparametrization causes the weights to diverge for logistic loss.} We depicted the order parameters $Q$ and $M$, quantifying the variance of the outputs of the student and their covariance with the outputs of the teacher, in the noiseless setup of Fig.~\ref{fig:agreement}. \textit{Insets:} log-log plot, showing the power-law asymptotic behaviors. 
    }
    \label{fig:norm}
\end{figure}

\paragraph{Behavioral differences between losses}


Further understanding of the differences between logistic and square loss can be gained thanks to the replica approach which gives access to the order parameter $Q$ and $M$ defined in Eq.~\ref{eq:replica-M-Q}, see Fig.~\ref{fig:norm}. For recall, $Q$ corresponds to the variance of the outputs of the student $\hat y$ and $M$ to their covariance with the linear scores of the teacher. 

For square loss, $Q$ and $M$ increase and reach a finite value (with a peak in $Q$ at the interpolation threshold), reflecting the fact that the linearized estimator $\bs F\bs w$ converges towards a fixed norm vector more or less correlated with the teacher vector $\bs \beta$ depending on the depending on the data structure. For logistic loss, $M$ increases linearly with overparametrization and $Q$ increases as a power law (see logarithmic insets), reflecting the fact the the estimator endlessly grows in the direction of the teacher vector~\cite{Soudry2018} as the number of parameters increases.
This growth appears to shield the peak observed in $Q$ for the squared loss explaining the very mild double descent observed in Fig. \ref{fig:agreement}.
Interestingly, these quantities grow much faster in the aligned setup, where the estimator is more ``confident'' in its predictions, which also hints at the better performance of the logistic loss when the structure of data is favorable.

\begin{figure}[tb]
    \centering
    \begin{subfigure}[b]{.49\textwidth}	   
    \includegraphics[width=\linewidth]{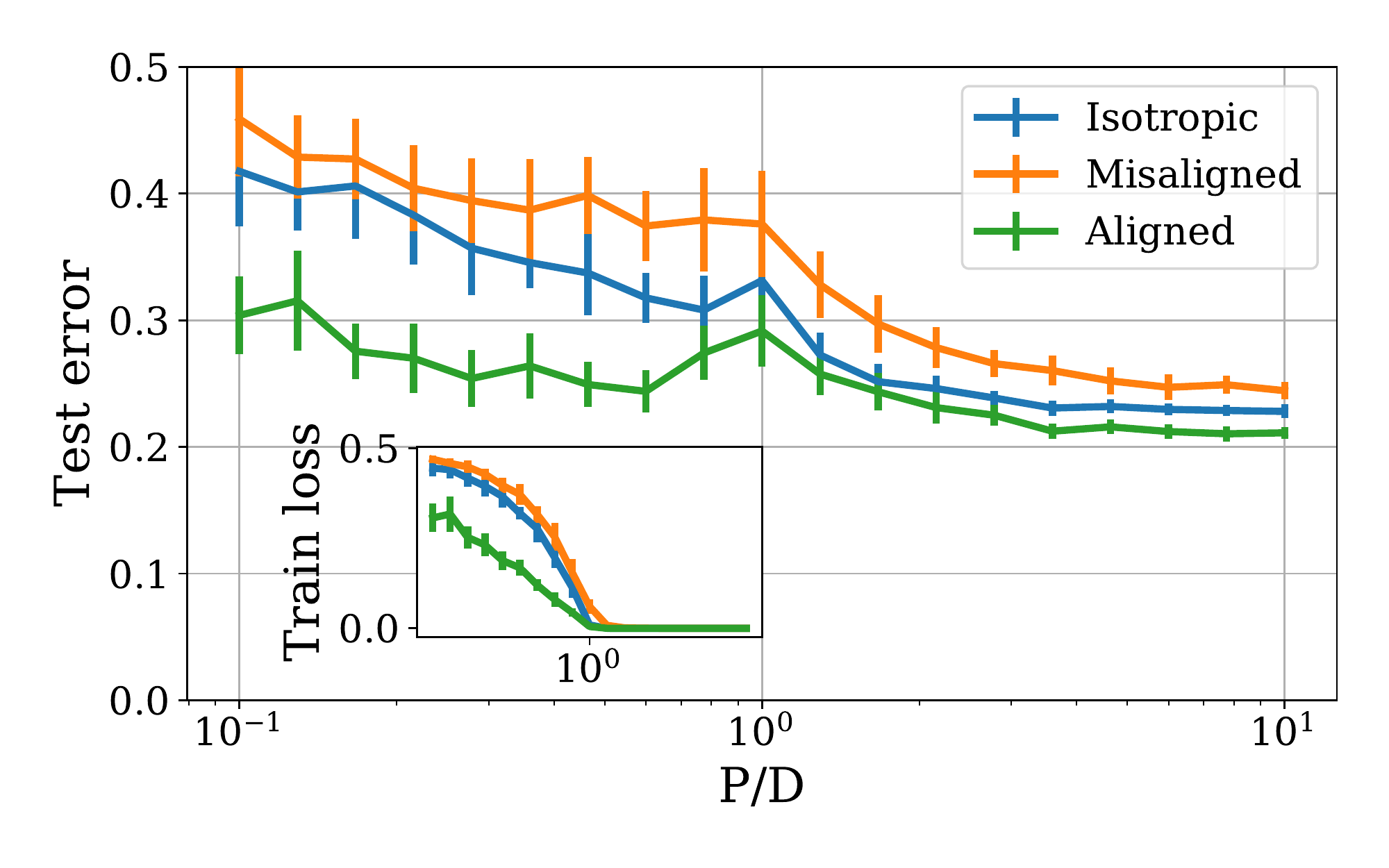}
    \caption{Square loss, MNIST}
    \end{subfigure}
    \begin{subfigure}[b]{.49\textwidth}	   
    \includegraphics[width=\linewidth]{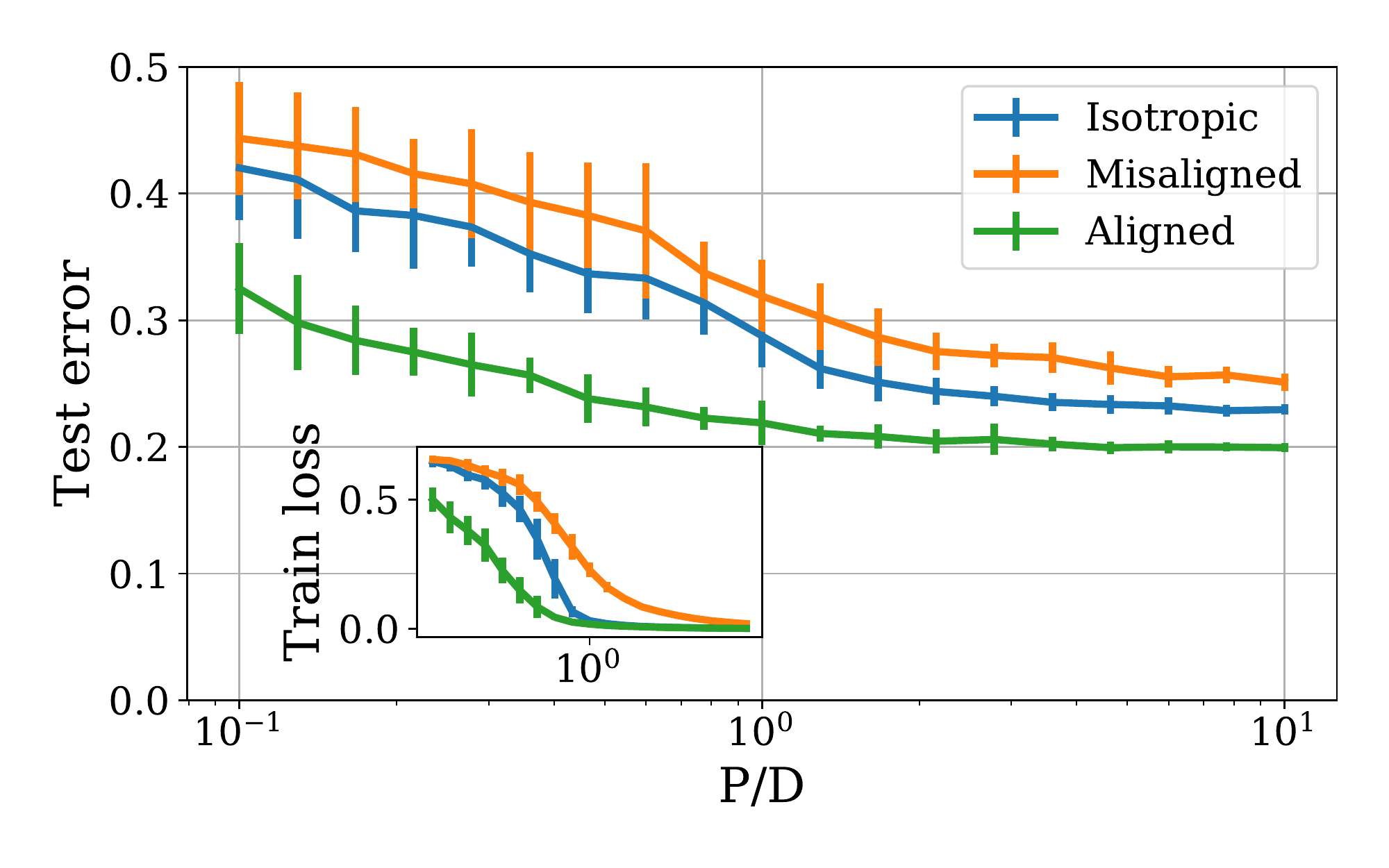}
    \caption{Logistic loss, MNIST}
    \end{subfigure}
    \begin{subfigure}[b]{.49\textwidth}	   
    \includegraphics[width=\linewidth]{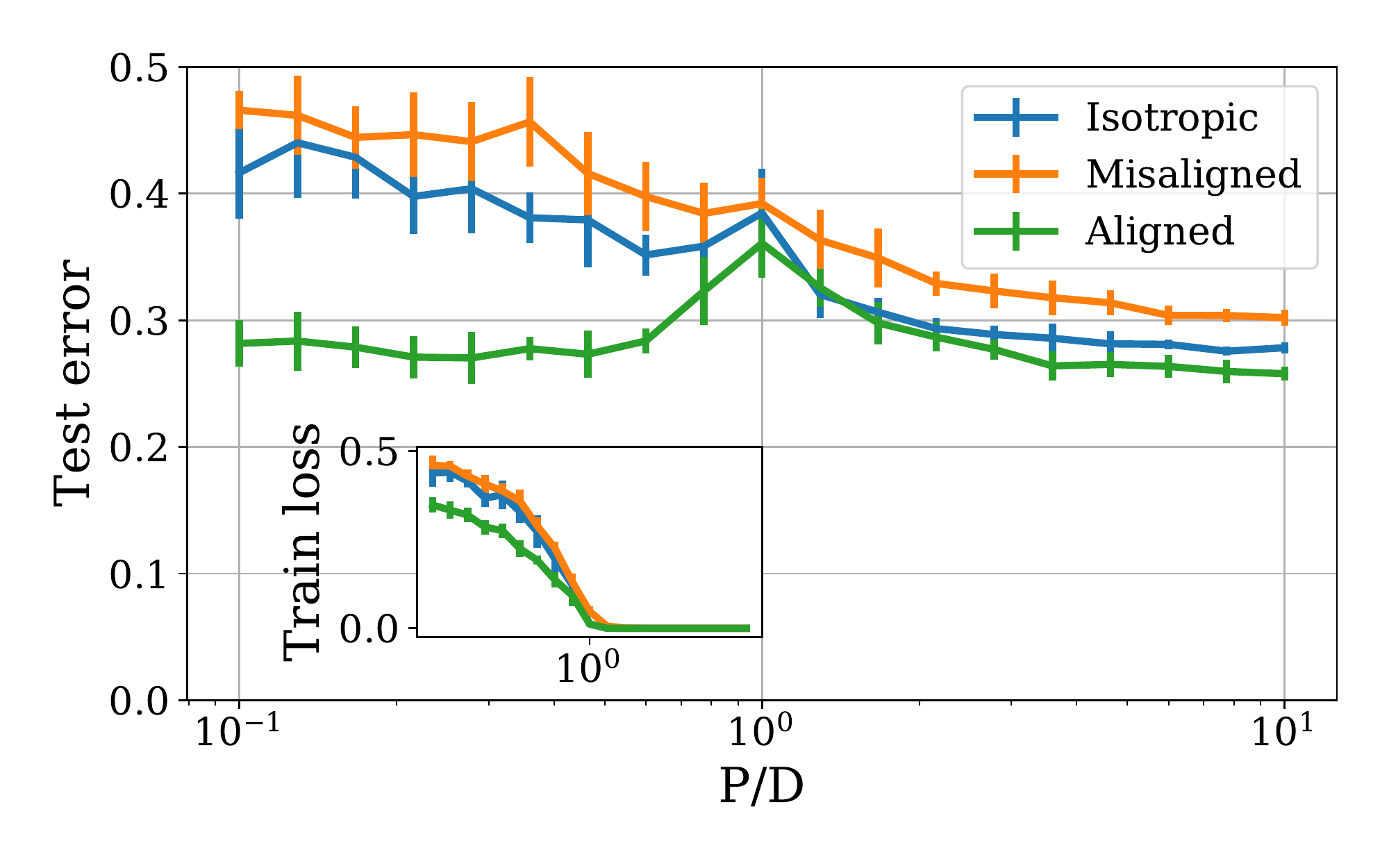}
    \caption{Square loss, CIFAR10}
    \end{subfigure}
    \begin{subfigure}[b]{.49\textwidth}	   
    \includegraphics[width=\linewidth]{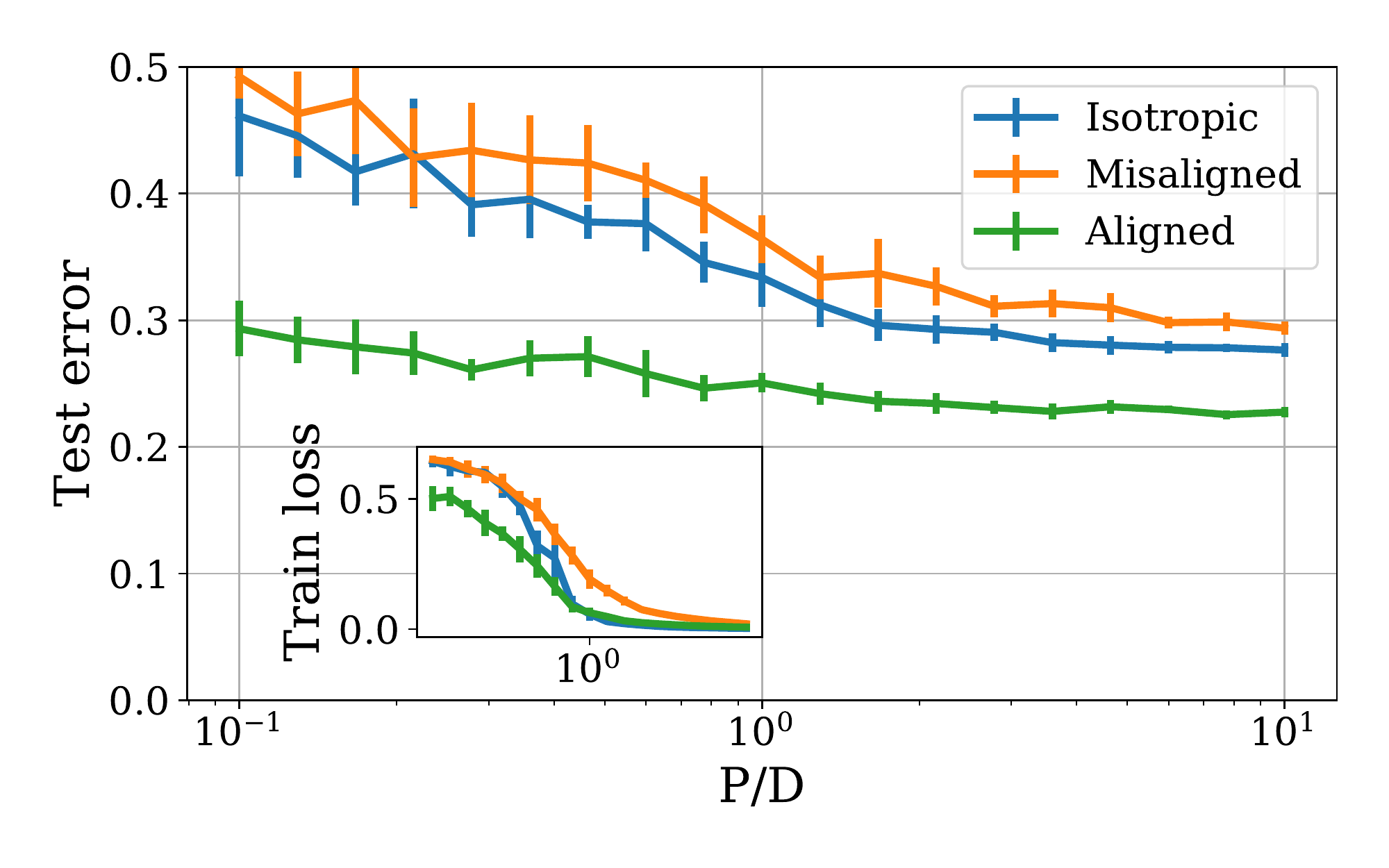}
    \caption{Logistic loss, CIFAR10}
    \end{subfigure}
    \caption{\textbf{Test error and train loss (inset) on realistic tasks.} \textit{Top:} MNIST dataset with labels given by the digit parity. \textit{Bottom:} CIFAR10 dataset with airplanes and cars. We synthetically reproduce the isotropic, aligned and misaligned scenarios by applying a PCA transformation to the inputs and tuning how salient the largest PCA features are compared to the smallest PCA features (see Sec.~\ref{sec:mnist}). We set $\sigma = \mathrm{Tanh}$, $\lambda=10^{-3}$ and $N/D=1$.}
    \label{fig:mnist}
\end{figure}


\section{Numerical results}
\label{sec:mnist}

To examine the applicability of our results, we consider two realistic binary classification tasks: parity of digits in the MNIST dataset and airplanes vs cars in the CIFAR10 dataset. In both cases, we learn with an RF model in the same setup as described above. To control the alignment, we apply a PCA transformation to the inputs (keeping the top $D=100$ components and discarding the rest), then apply the following component-wise rescaling:
\begin{align}
    \bs x_i\to \bs x_i/\operatorname{std}(\bs x_i)^{\alpha},
\end{align}
where $\operatorname{std}(\bs x_i)$ denotes the standard deviation of feature $\bs x_i$ over the whole training dataset, and the exponent $\alpha$ allows us to tune the saliency of the features:
\begin{itemize}
    \item $\alpha<1$ yields an \textbf{aligned} scenario, since the top PCA features are naturally the most relevant;
    \item $\alpha=1$ yields the \textbf{isotropic} scenario, where all features have same variance;
    \item $\alpha>1$ yields a \textbf{misaligned} scenario, since the strong features will become weak and the weak features become strong. 
\end{itemize} 

The corresponding train and test error curves are shown in Fig.~\ref{fig:mnist} (we set $\alpha=0$ for the aligned scenario and $\alpha=1.5$ for the misaligned scenario). Remarkably, we recover many of the phenomenological features described previously. The test error drops earlier and reaches a lower asymptotic value in the aligned setup, and conversely reaches a higher value in the misaligned setup. We observe a double descent curve for square loss, but the peak is suppressed for logistic loss. The location of the interpolation threshold depends on the teacher-data alignment for logistic loss, whereas it does not for square loss. Finally, logistic loss has a lower asymptotic error than square loss in the aligned setup, signalling that it is favorable for ``easy'' data distributions.

To strengthen the latter observation, we vary continuously the difficulty of the task by adjusting the exponent $\alpha$ and show the results in Fig.~\ref{fig:anisotropy}(b) (increasing $\alpha$ makes the task harder). As observed analytically in Fig.~\ref{fig:anisotropy}(a), the gap between square loss and logistic loss increases as we decrease $\alpha$.

\section{Conclusion}

In this work, we studied how the loss function interplays with the data structure to shape the generalization curve of random feature models. Our results show strong behavioral differences between quadratic and logistic loss, the latter performing particularly well for easy datasets where most of the information comes from low-dimension projections of the inputs. 

As a possible direction of future work, we conclude with the observation that our results, which apply to random feature (or lazy learning) tasks, appear in contrast with those of \citet{hui2020evaluation}, which suggest that cross-entropy losses can be traded at no cost for quadratic losses in modern deep learning tasks, which are known to have low intrinsic dimensionality~\cite{pope2021intrinsic}. This opens up an interesting direction for future work: does feature learning help quadratic losses by better capturing the low-dimensional structure of the inputs, as suggested by~\cite{baratin2020implicit}? Or does the key difference reside in the multi-class nature of practical classification problems~\cite{demirkaya2020exploring}? 

\paragraph{Acknowledgements}
We thank Berfin Simsek, Armand Joulin, Bruno Loureiro and Federica Gerace for helpful discussions. SD and GB acknowledge funding from the French government under management of Agence Nationale de la Recherche as part of the “Investissements d’avenir” program, reference ANR-19-P3IA-0001 (PRAIRIE 3IA Institute). MG acknowledges funding from the Flatiron Institute. We also thank Lenka Zdeborova and Florent Krzakala, and Les Houches Summer School for organizing a workshop on
Statistical Physics and Machine Learning where this work was partially done. 
\clearpage
\bibliographystyle{unsrt}
\bibliography{refs}

\appendix
\setlength{\parindent}{0pt}

\section{Phase spaces}
\label{app:phase-space}

\begin{figure}[h!]
    \centering
    \begin{subfigure}[b]{.24\textwidth}	   
    \includegraphics[width=\linewidth]{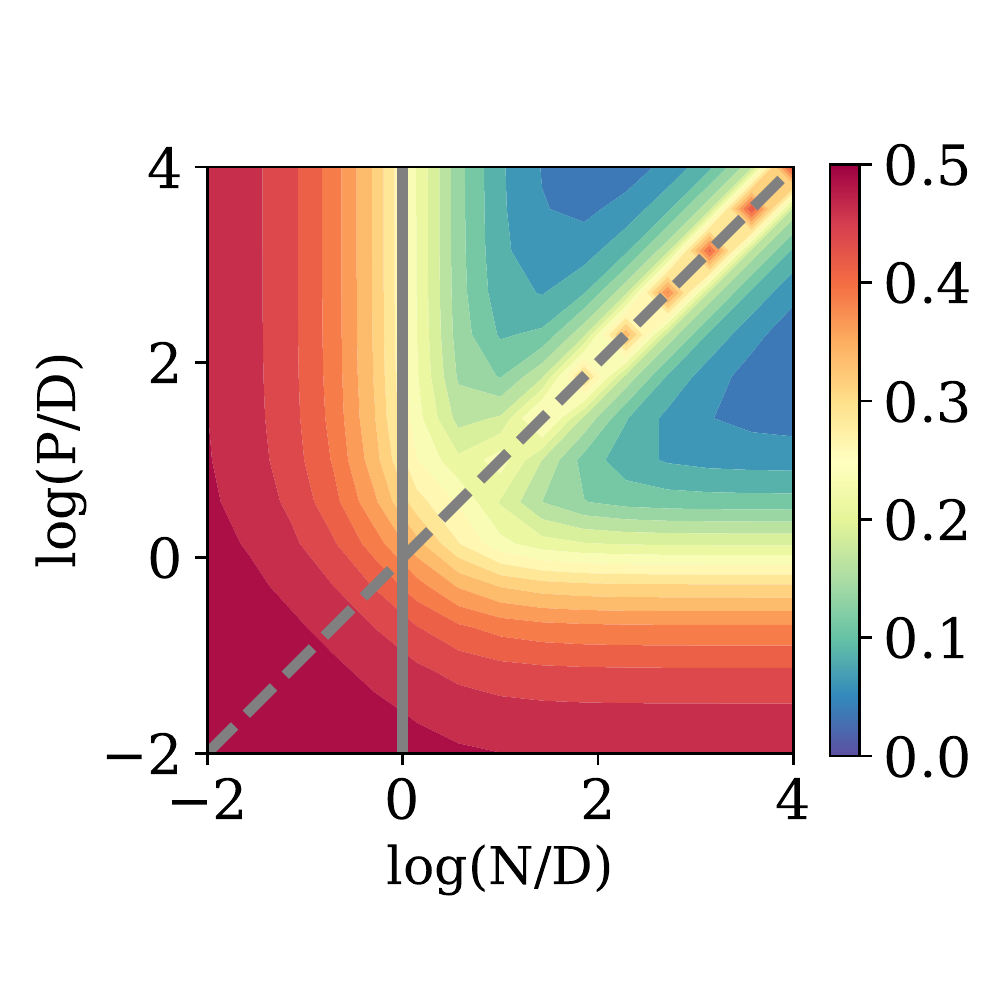}
    \caption{Isotropic, noiseless}
    \end{subfigure}
    \begin{subfigure}[b]{.24\textwidth}	   
    \includegraphics[width=\linewidth]{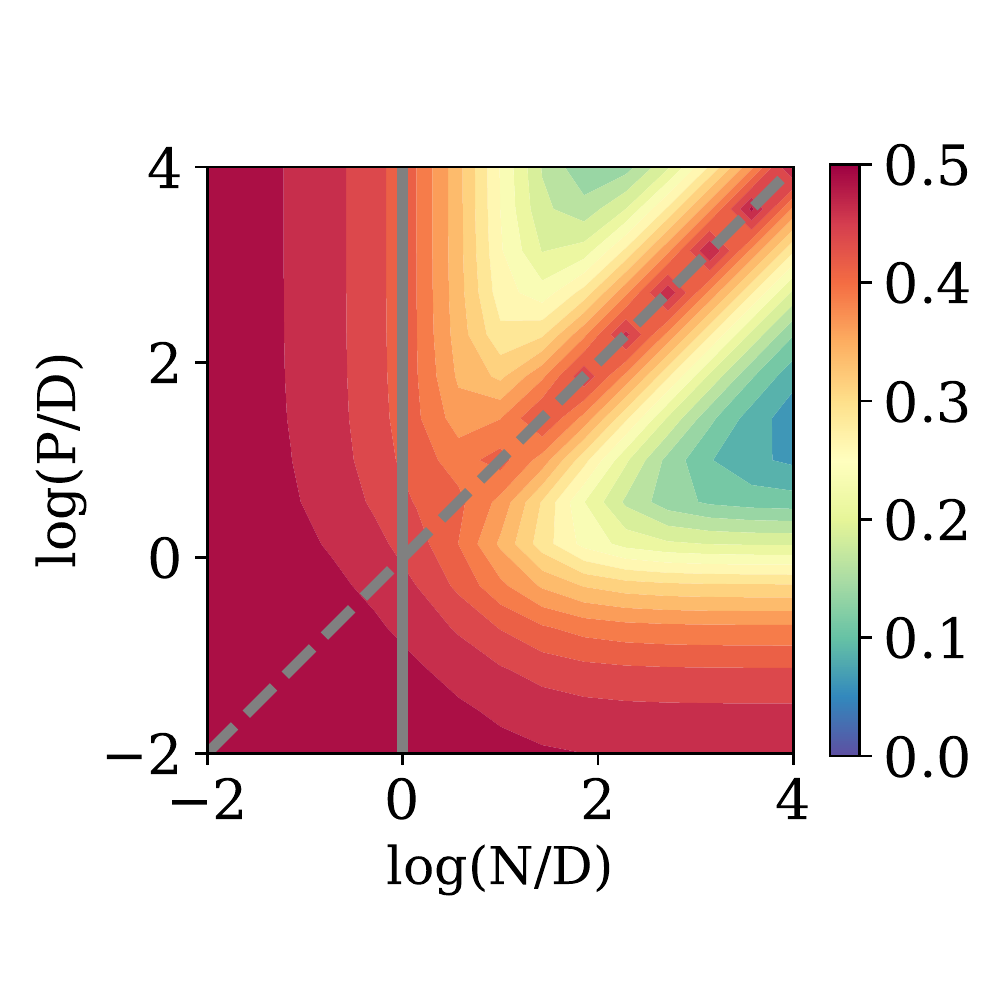}
    \caption{Isotropic, $\Delta=0.4$}
    \end{subfigure}
    \begin{subfigure}[b]{.24\textwidth}	   
    \includegraphics[width=\linewidth]{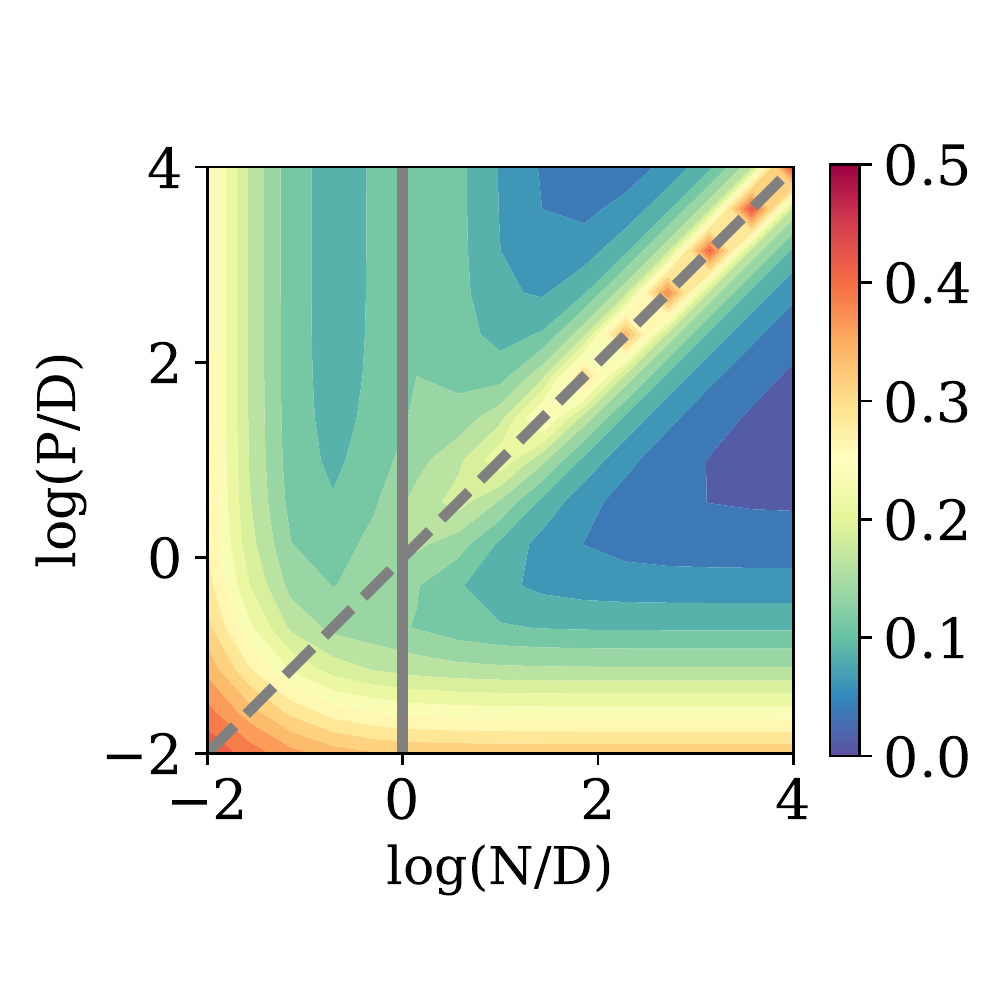}
    \caption{Aligned, $\Delta=0$}
    \end{subfigure}
    \begin{subfigure}[b]{.24\textwidth}	   
    \includegraphics[width=\linewidth]{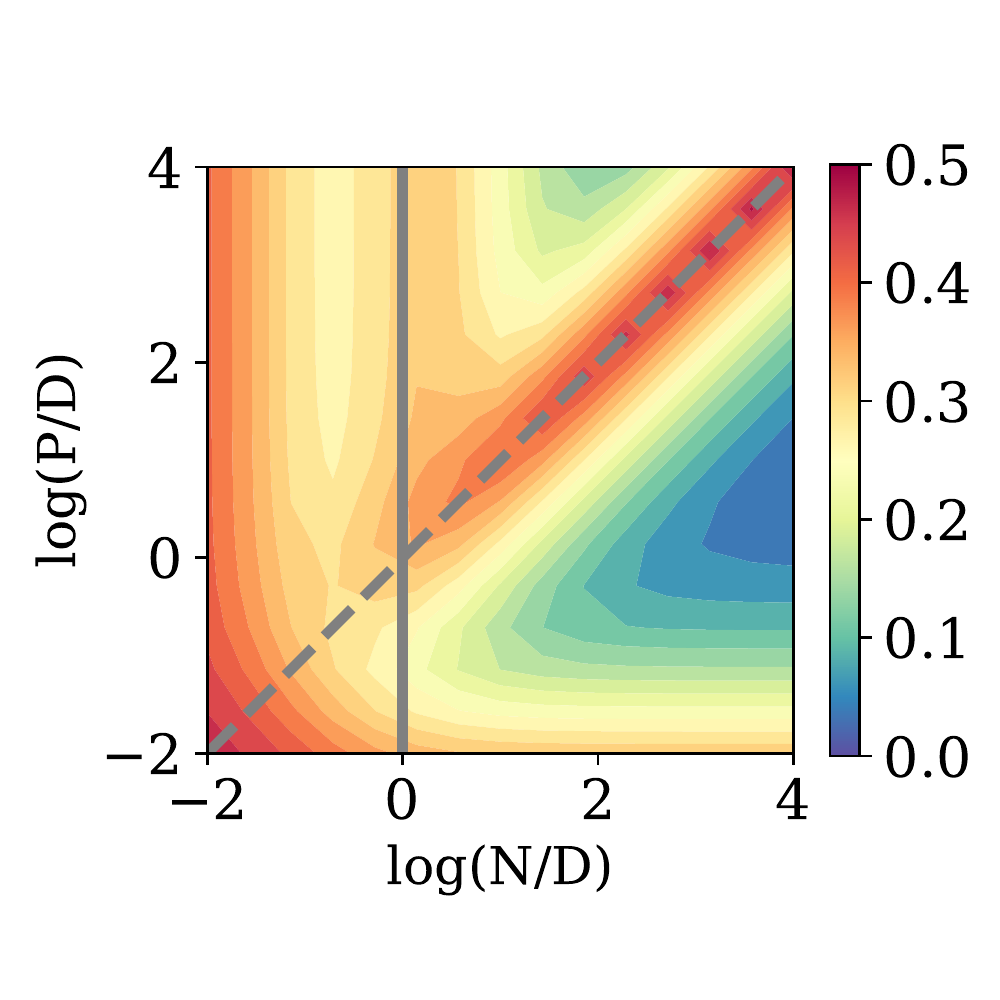}
    \caption{Aligned, $\Delta=0.4$}
    \end{subfigure}
    \begin{subfigure}[b]{.24\textwidth}	   
    \includegraphics[width=\linewidth]{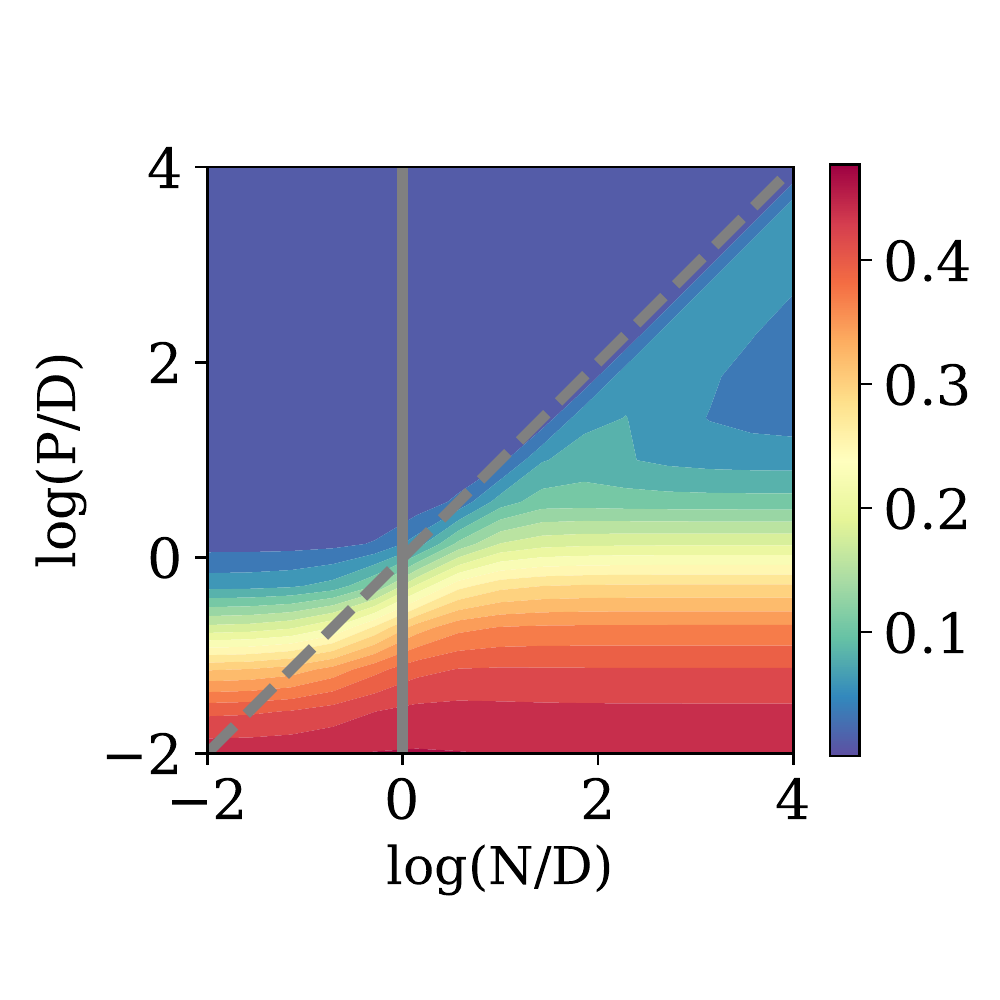}
    \caption{Isotropic, noiseless}
    \end{subfigure}
    \begin{subfigure}[b]{.24\textwidth}	   
    \includegraphics[width=\linewidth]{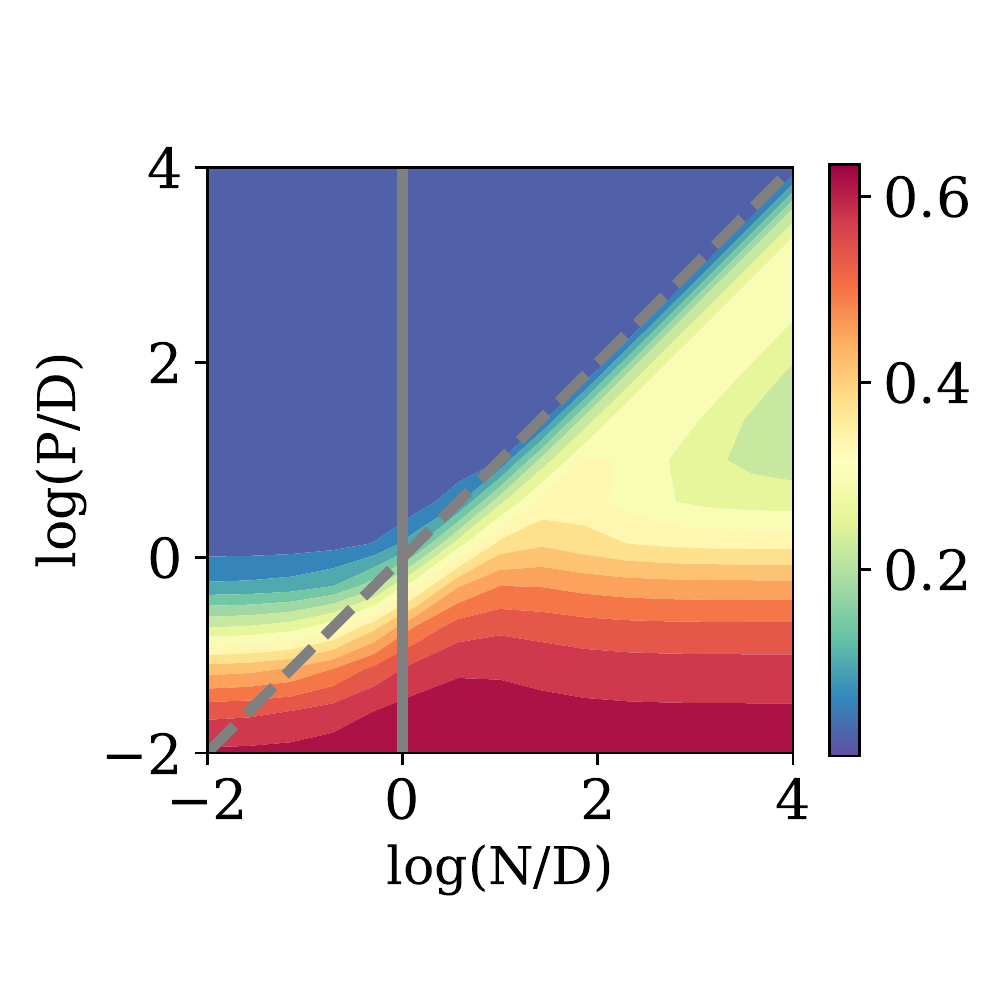}
    \caption{Isotropic, $\Delta=0.4$}
    \end{subfigure}
    \begin{subfigure}[b]{.24\textwidth}	   
    \includegraphics[width=\linewidth]{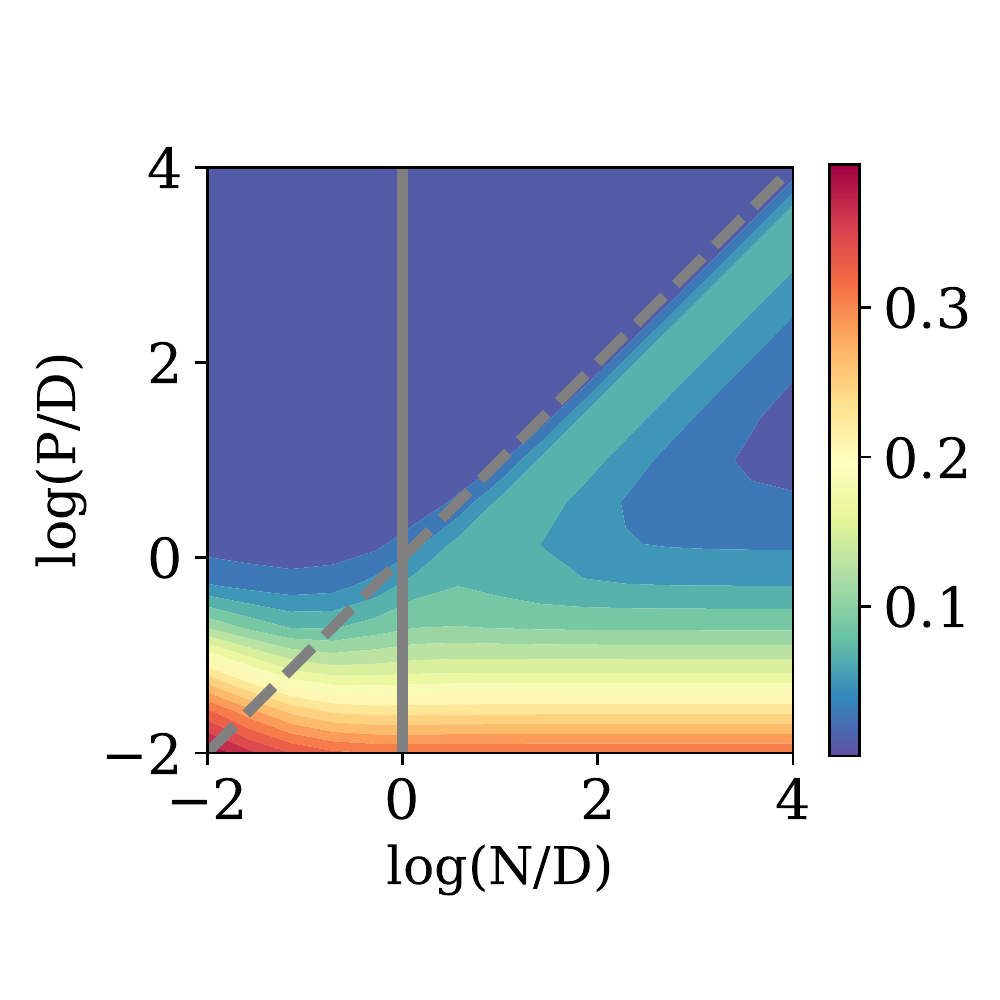}
    \caption{Aligned, $\Delta=0$}
    \end{subfigure}
    \begin{subfigure}[b]{.24\textwidth}	   
    \includegraphics[width=\linewidth]{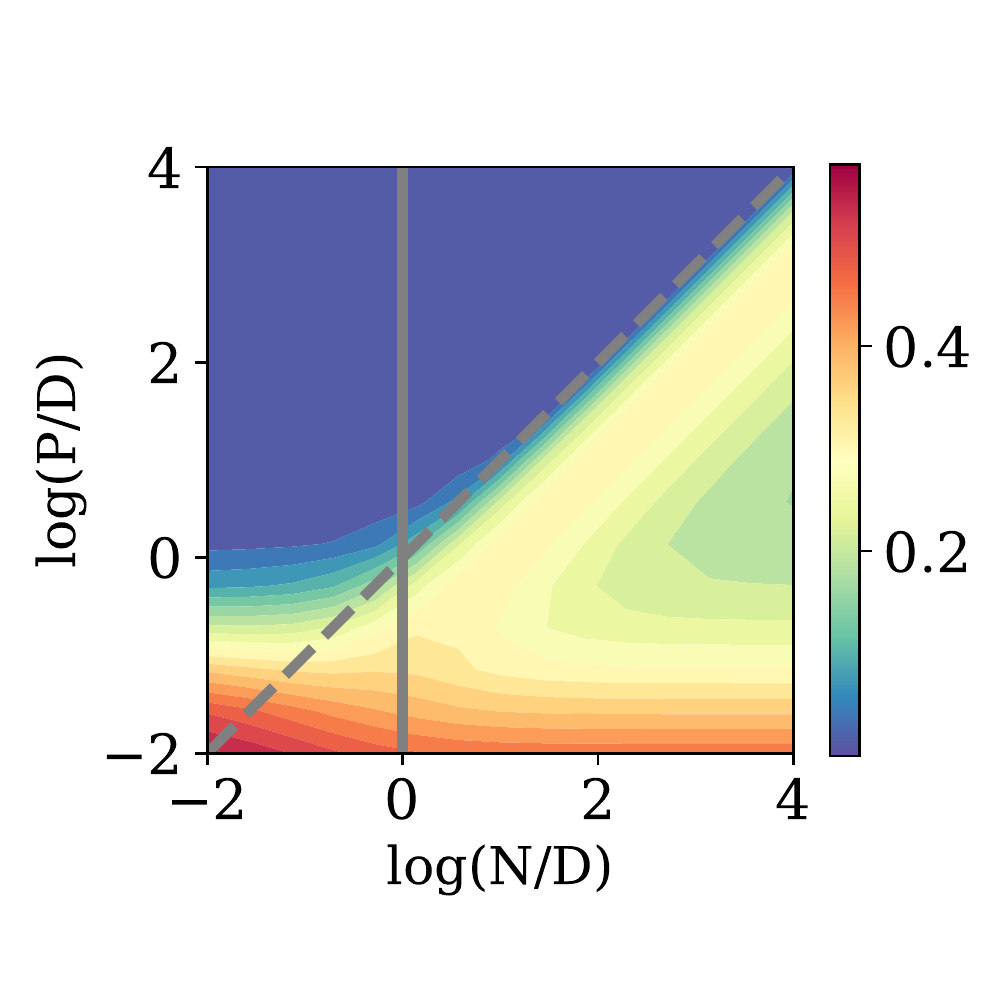}
    \caption{Aligned, $\Delta=0.4$}
    \end{subfigure}
    \begin{subfigure}[b]{.24\textwidth}	   
    \includegraphics[width=\linewidth]{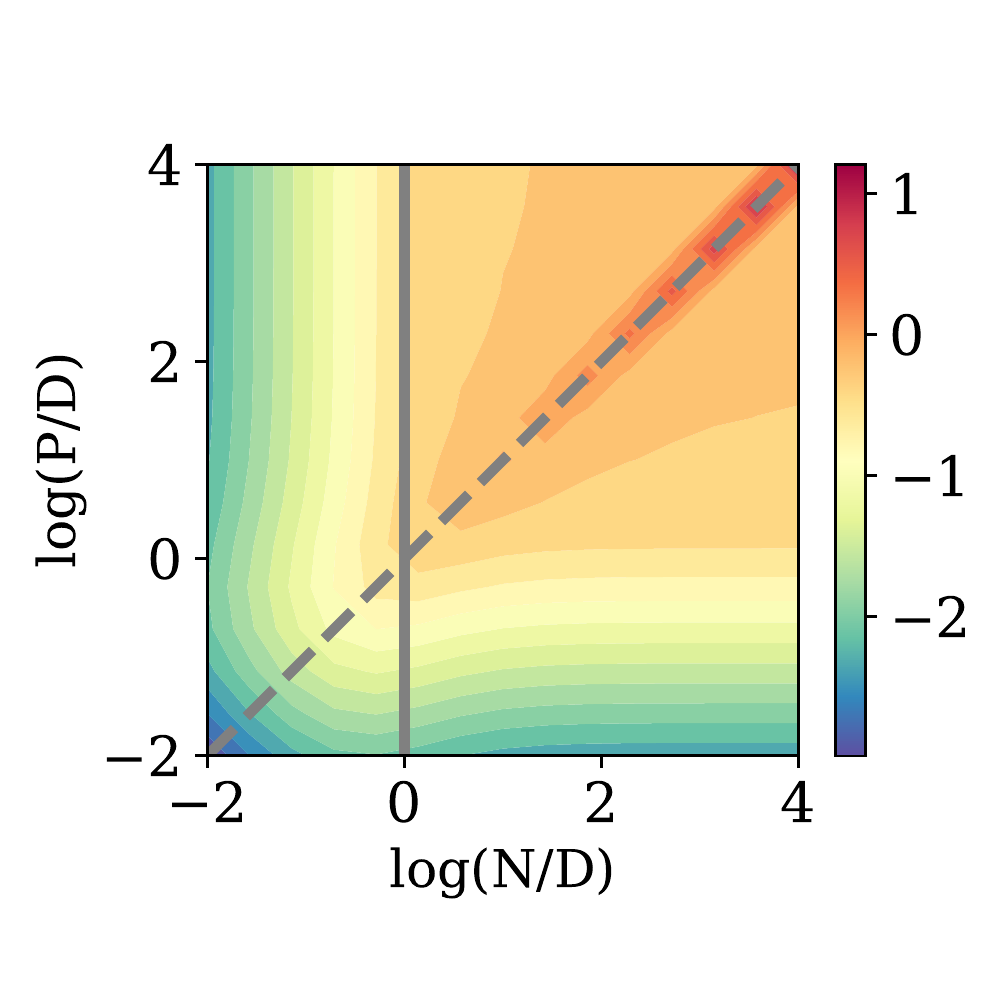}
    \caption{Isotropic, noiseless}
    \end{subfigure}
    \begin{subfigure}[b]{.24\textwidth}	   
    \includegraphics[width=\linewidth]{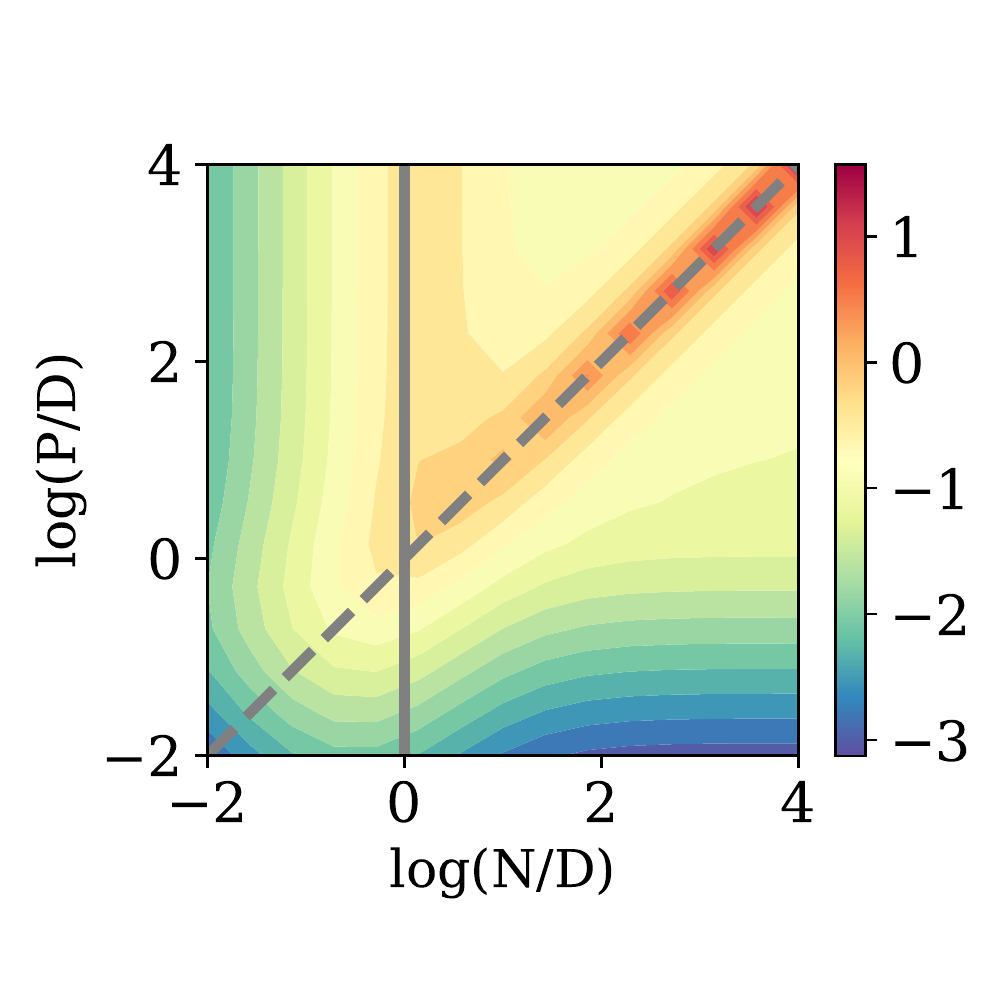}
    \caption{Isotropic, $\Delta=0.4$}
    \end{subfigure}
    \begin{subfigure}[b]{.24\textwidth}	   
    \includegraphics[width=\linewidth]{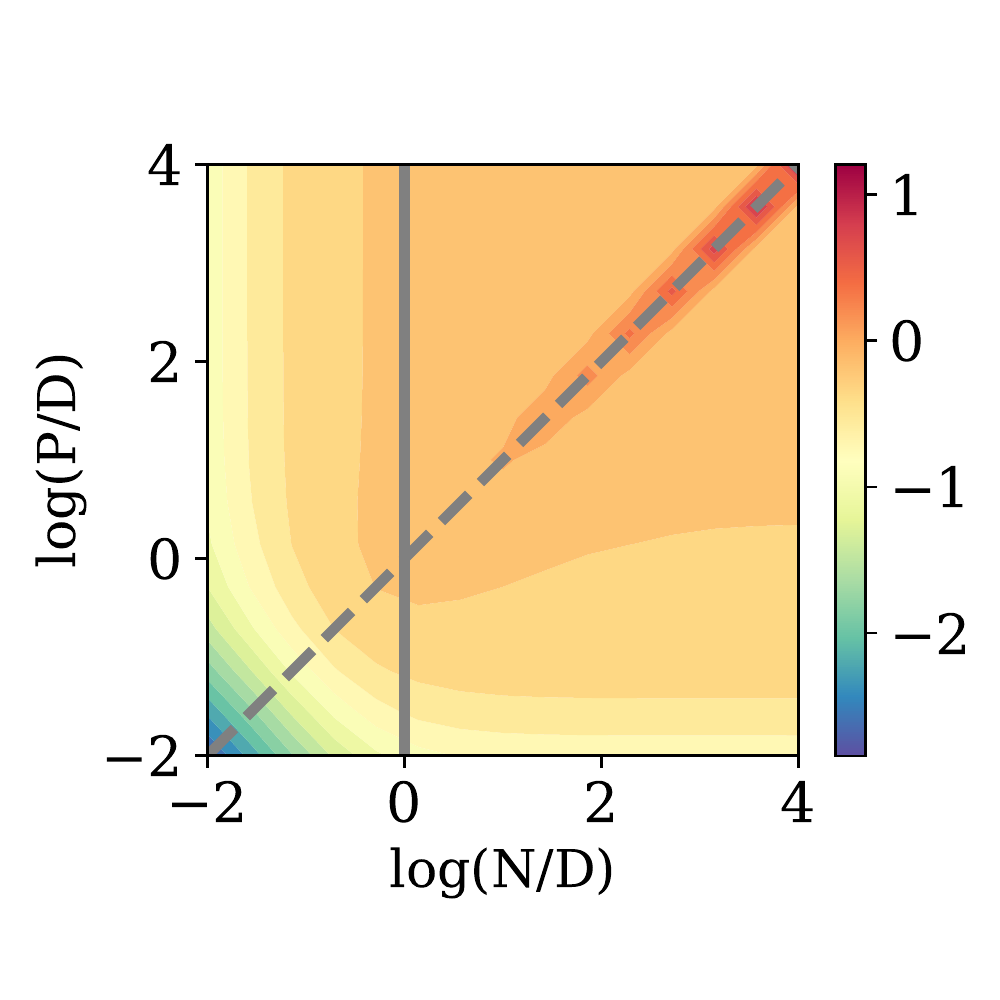}
    \caption{Aligned, $\Delta=0$}
    \end{subfigure}
    \begin{subfigure}[b]{.24\textwidth}	   
    \includegraphics[width=\linewidth]{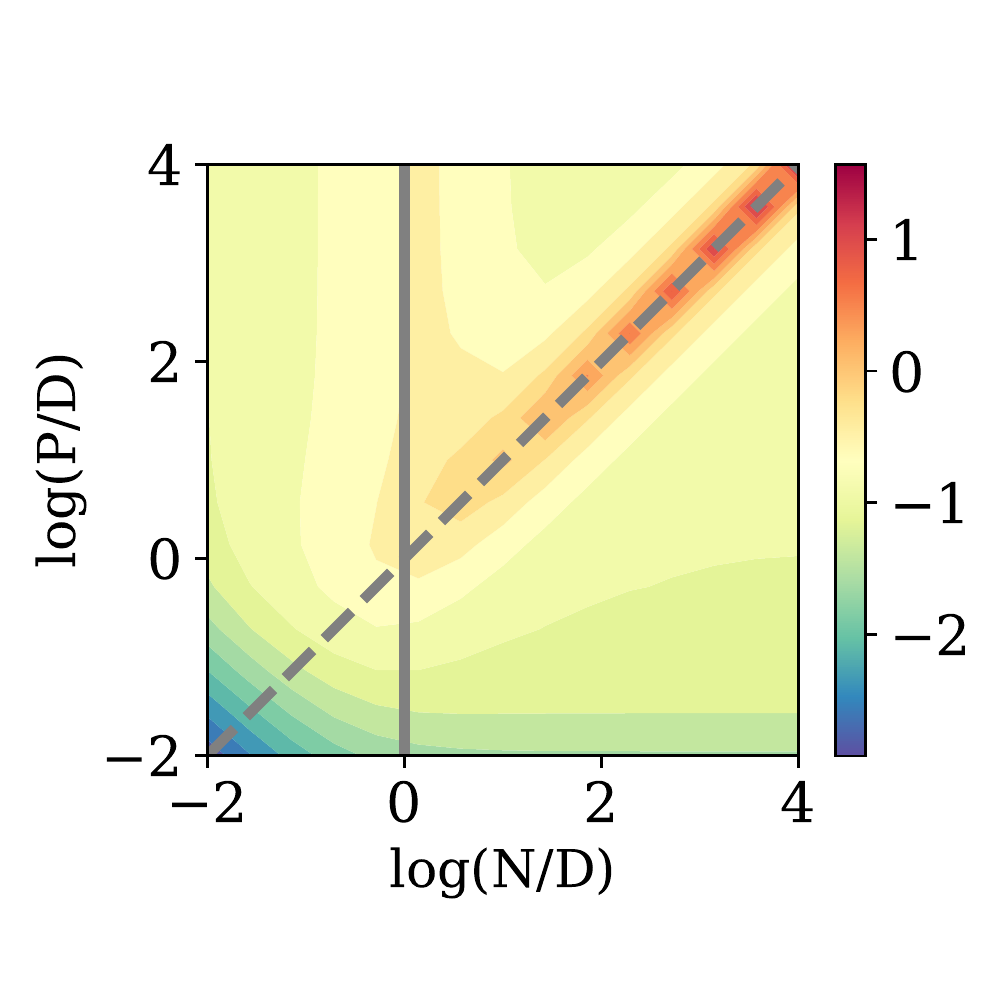}
    \caption{Aligned, $\Delta=0.4$}
    \end{subfigure}
    \begin{subfigure}[b]{.24\textwidth}	   
    \includegraphics[width=\linewidth]{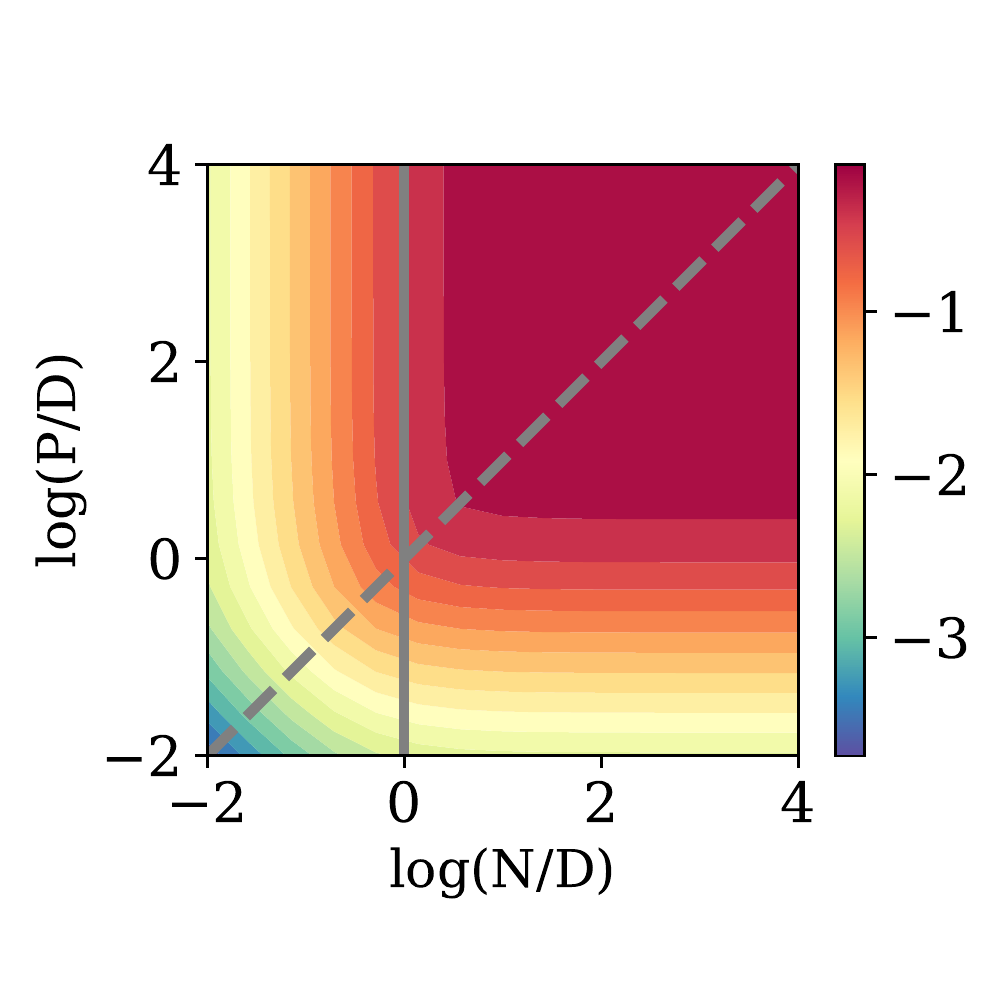}
    \caption{Isotropic, noiseless}
    \end{subfigure}
    \begin{subfigure}[b]{.24\textwidth}	   
    \includegraphics[width=\linewidth]{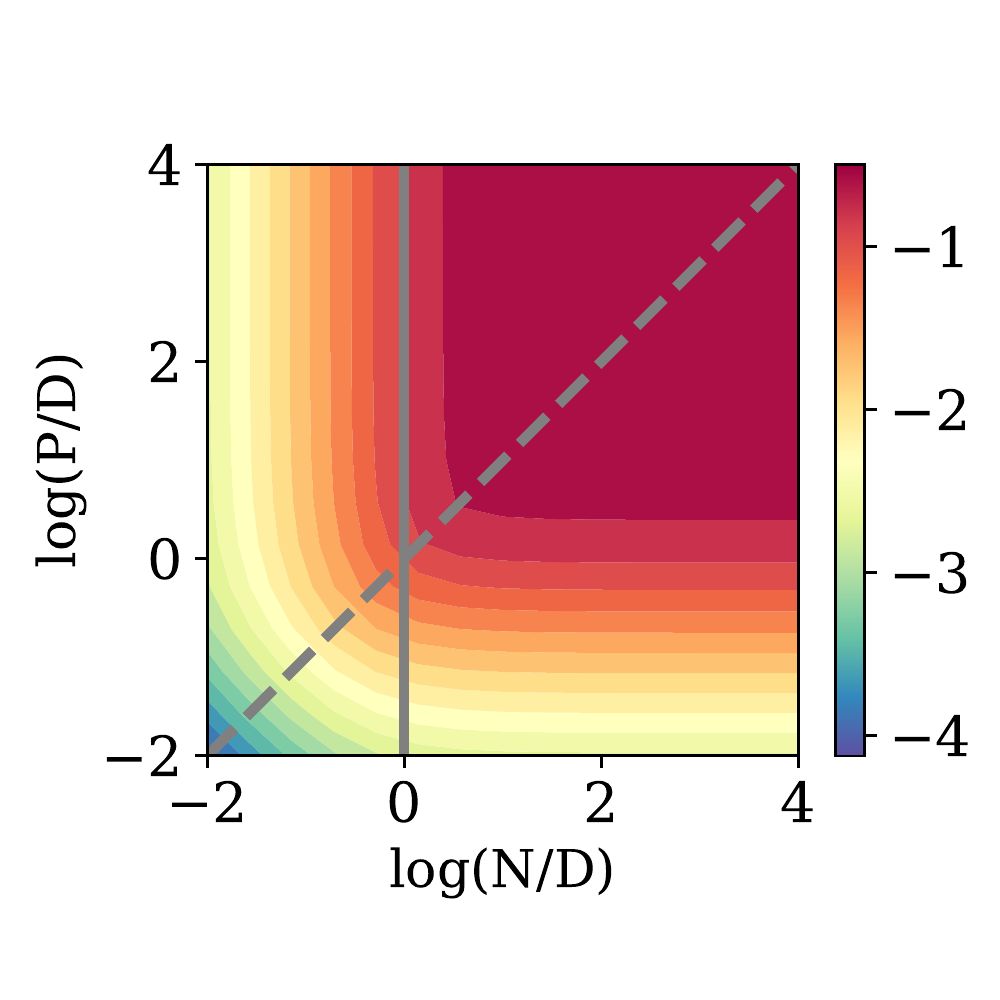}
    \caption{Isotropic, $\Delta=0.4$}
    \end{subfigure}
    \begin{subfigure}[b]{.24\textwidth}	   
    \includegraphics[width=\linewidth]{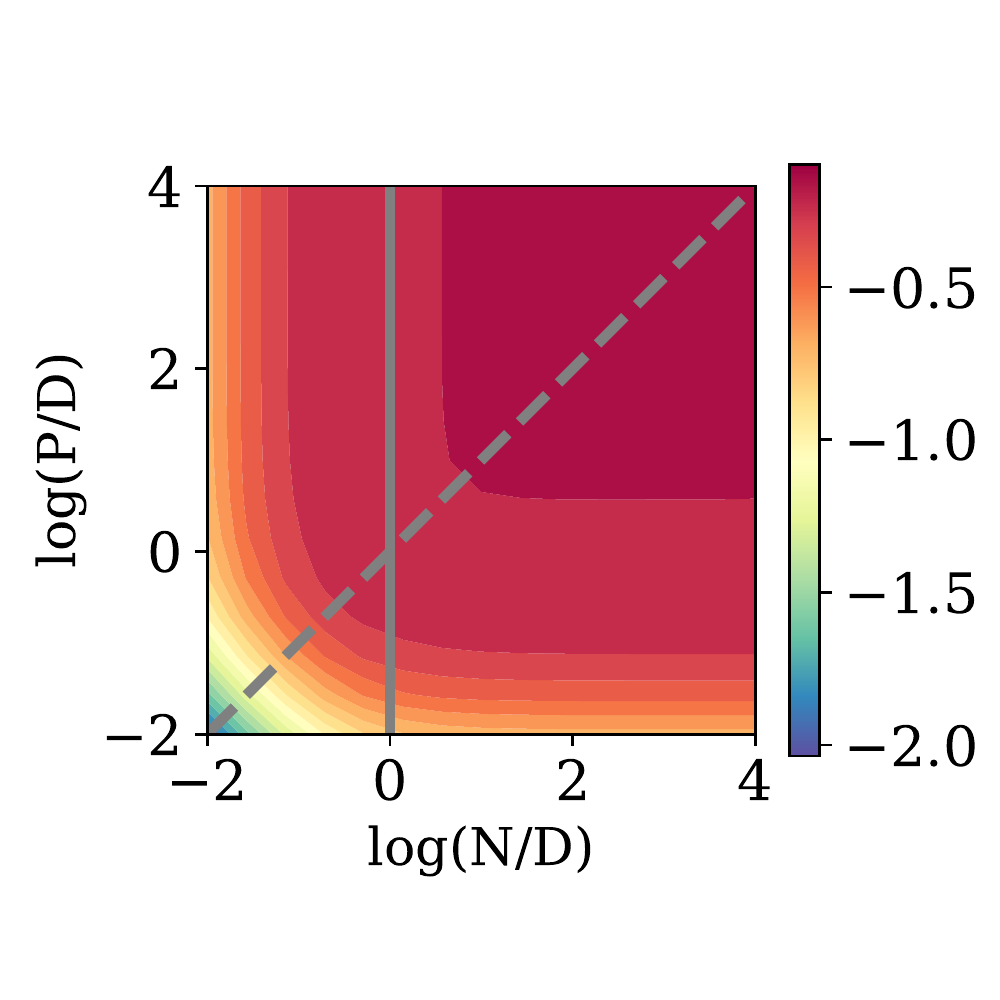}
    \caption{Aligned, $\Delta=0$}
    \end{subfigure}
    \begin{subfigure}[b]{.24\textwidth}	   
    \includegraphics[width=\linewidth]{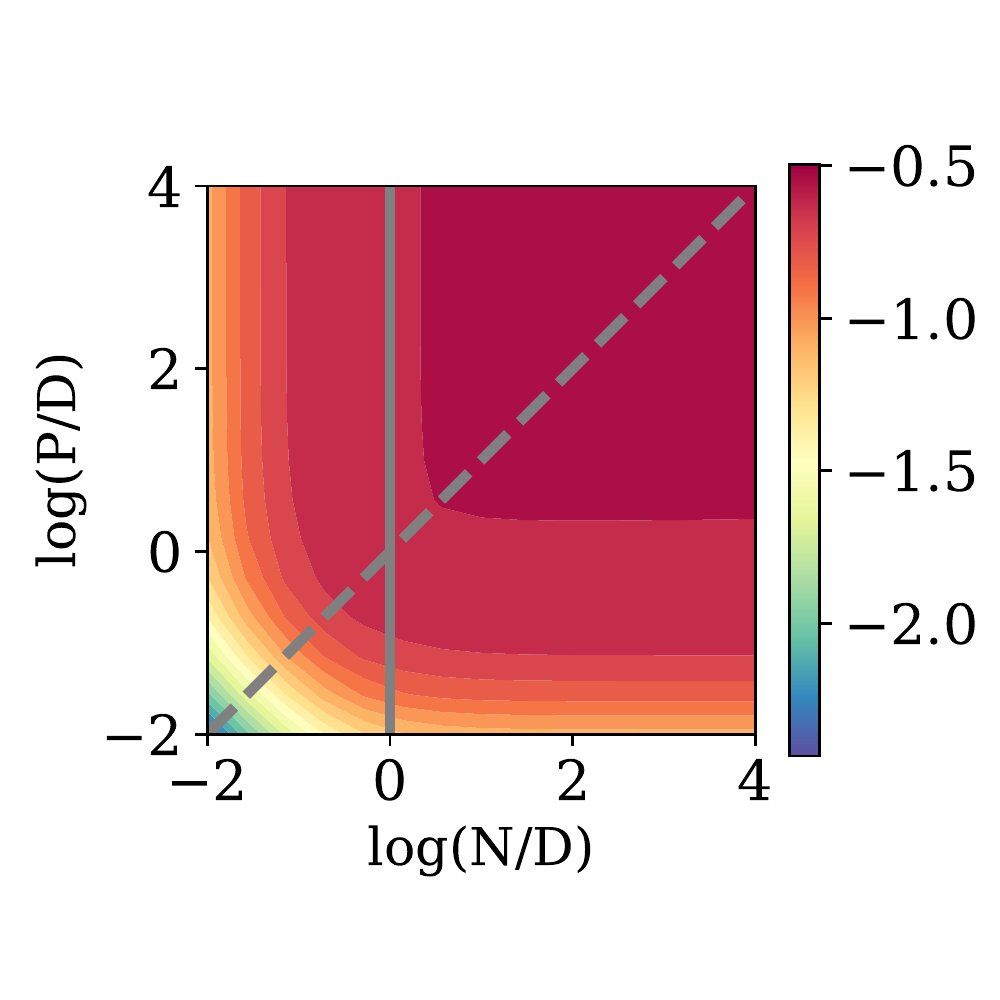}
    \caption{Aligned, $\Delta=0.4$}
    \end{subfigure}
    \caption{\textbf{Square loss phase spaces.} We studied the classification task for $\sigma = \mathrm{ReLU}$, $\lambda=0.1$. In the anisotropic phase spaces we set $\phi_1=0.01$, $r_x=100$, $r_\beta=100$. \textit{First row:} test error. \textit{Second row:} train error. \textit{Third row:} Q. \textit{Fourth row:} M. The solid and dashed grey lines represent the $N=D$ and $N=P$ lines, where one can find overfitting peaks~\cite{d2020triple}. For $Q$ and $M$, the colormaps are logarithmic.}
    \label{fig:phase-spaces-square}
\end{figure}

\begin{figure}[h!]
    \centering
    \begin{subfigure}[b]{.24\textwidth}	   
    \includegraphics[width=\linewidth]{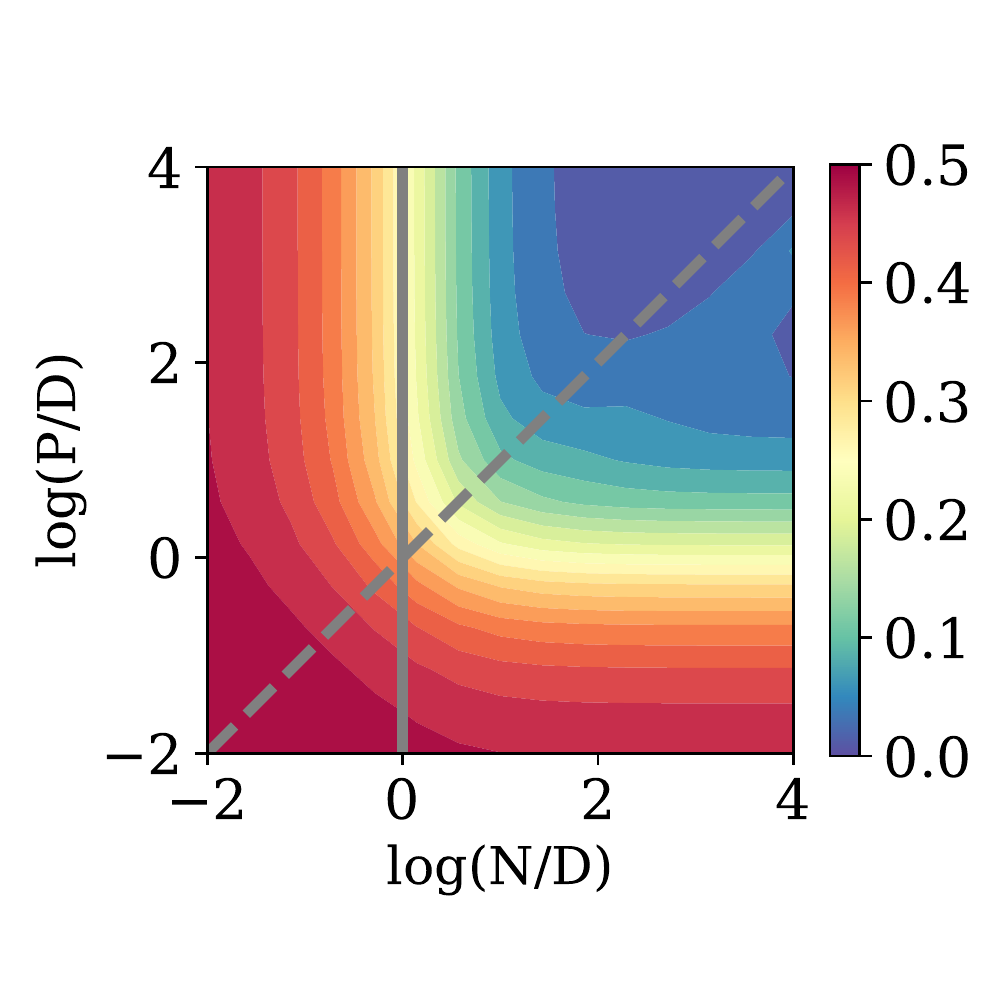}
    \caption{Isotropic, noiseless}
    \end{subfigure}
    \begin{subfigure}[b]{.24\textwidth}	   
    \includegraphics[width=\linewidth]{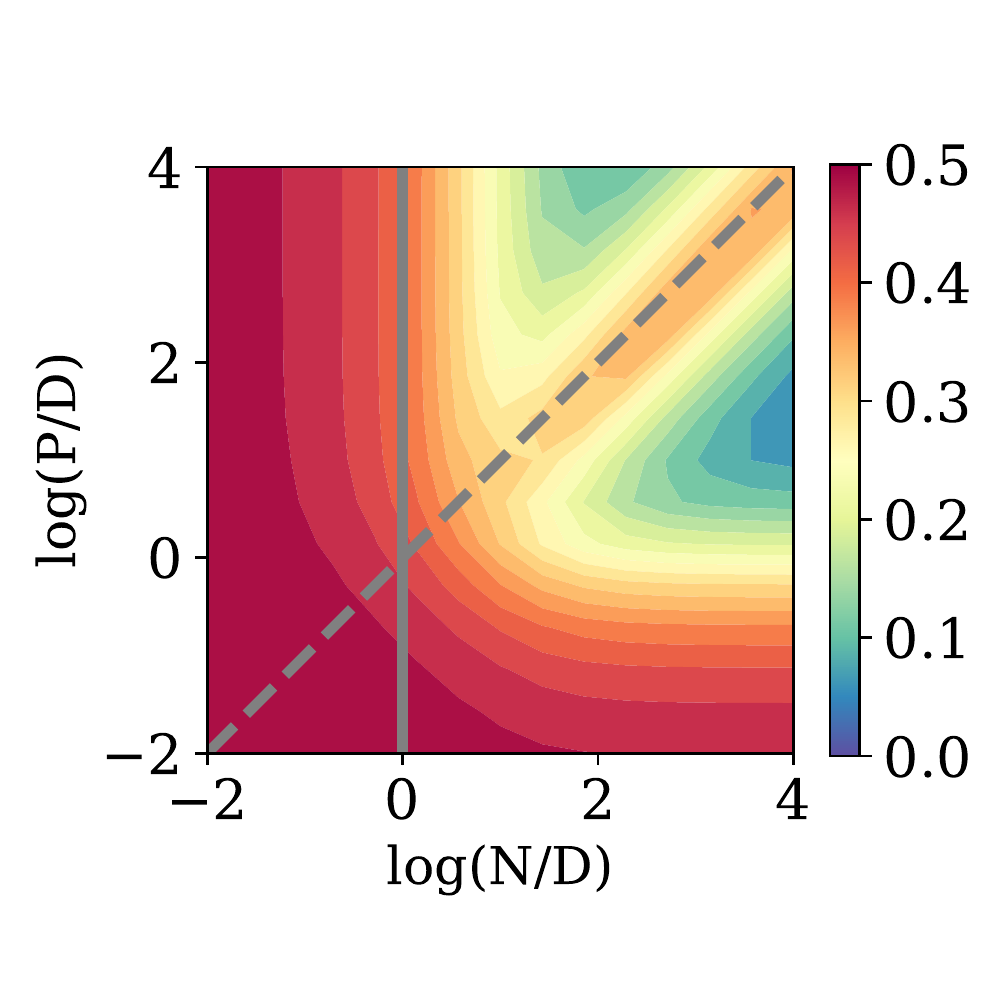}
    \caption{Isotropic, $\Delta=0.4$}
    \end{subfigure}
    \begin{subfigure}[b]{.24\textwidth}	   
    \includegraphics[width=\linewidth]{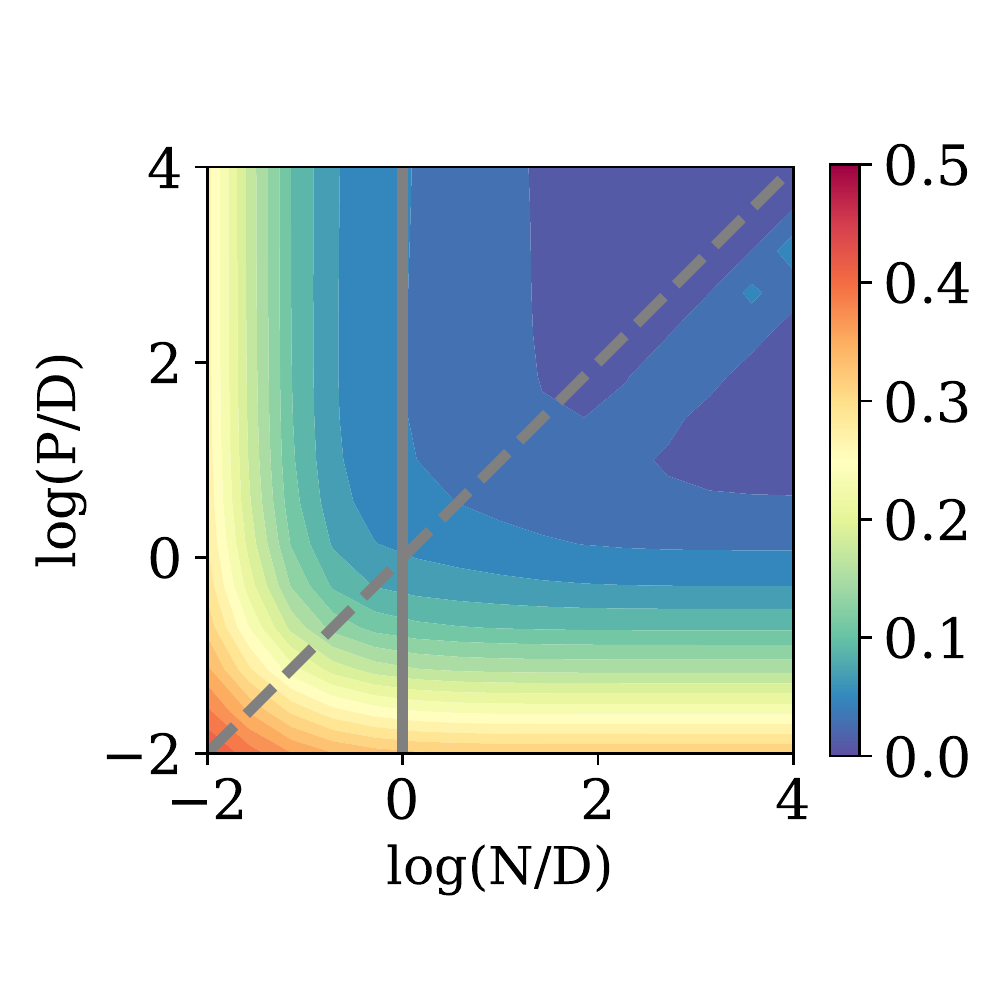}
    \caption{Aligned, $\Delta=0$}
    \end{subfigure}
    \begin{subfigure}[b]{.24\textwidth}	   
    \includegraphics[width=\linewidth]{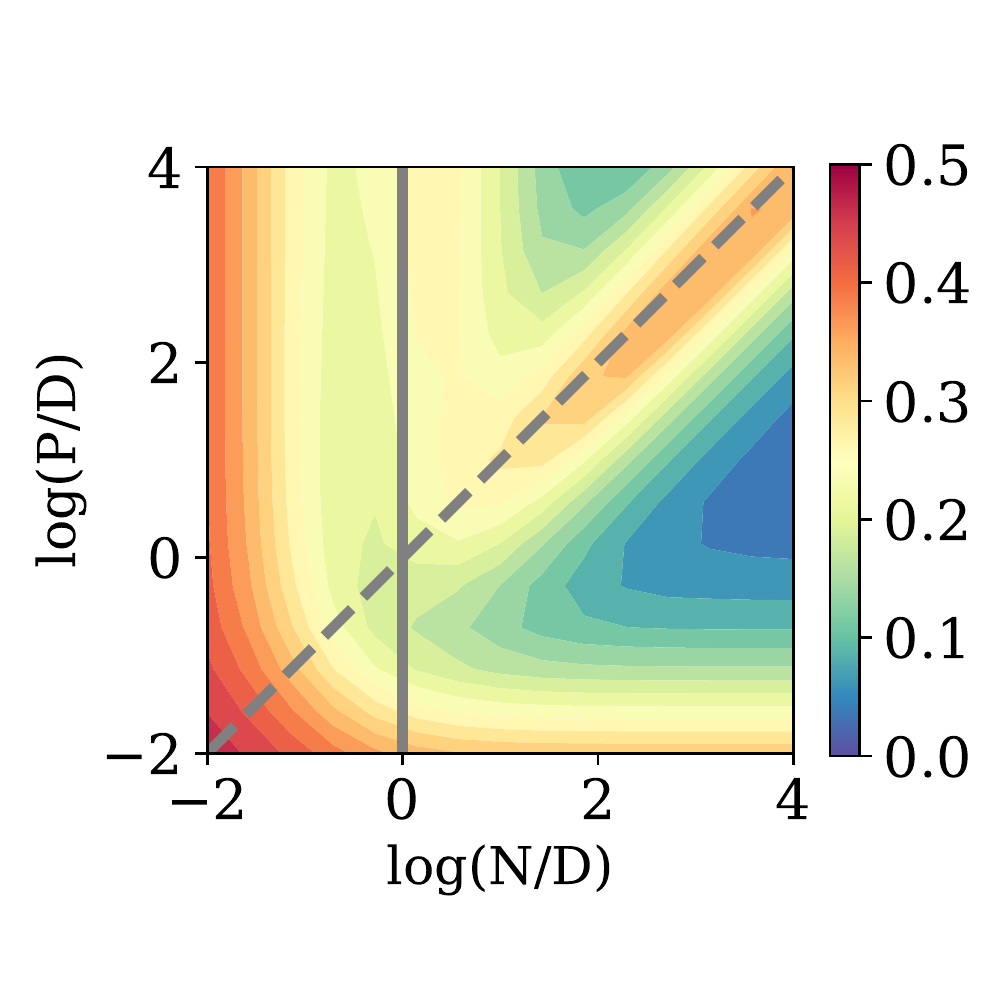}
    \caption{Aligned, $\Delta=0.4$}
    \end{subfigure}
    \begin{subfigure}[b]{.24\textwidth}	   
    \includegraphics[width=\linewidth]{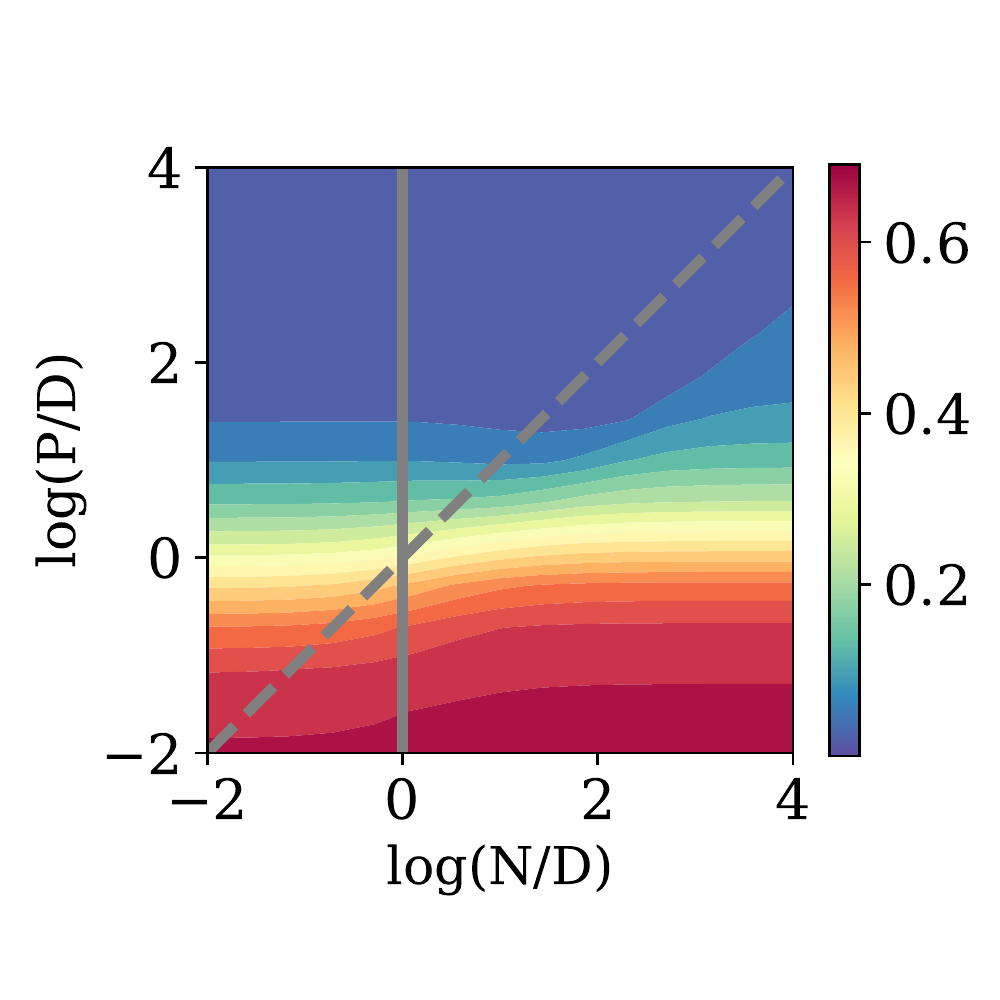}
    \caption{Isotropic, noiseless}
    \end{subfigure}
    \begin{subfigure}[b]{.24\textwidth}	   
    \includegraphics[width=\linewidth]{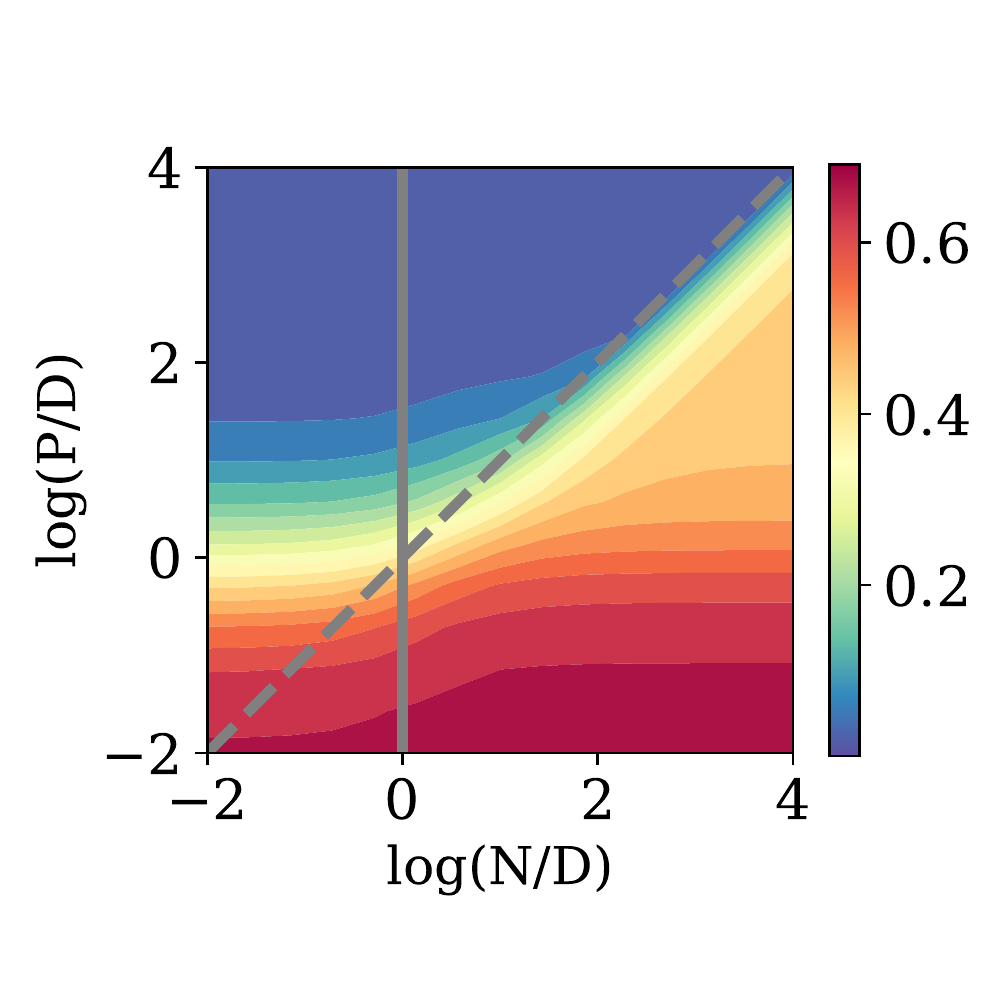}
    \caption{Isotropic, $\Delta=0.4$}
    \end{subfigure}
    \begin{subfigure}[b]{.24\textwidth}	   
    \includegraphics[width=\linewidth]{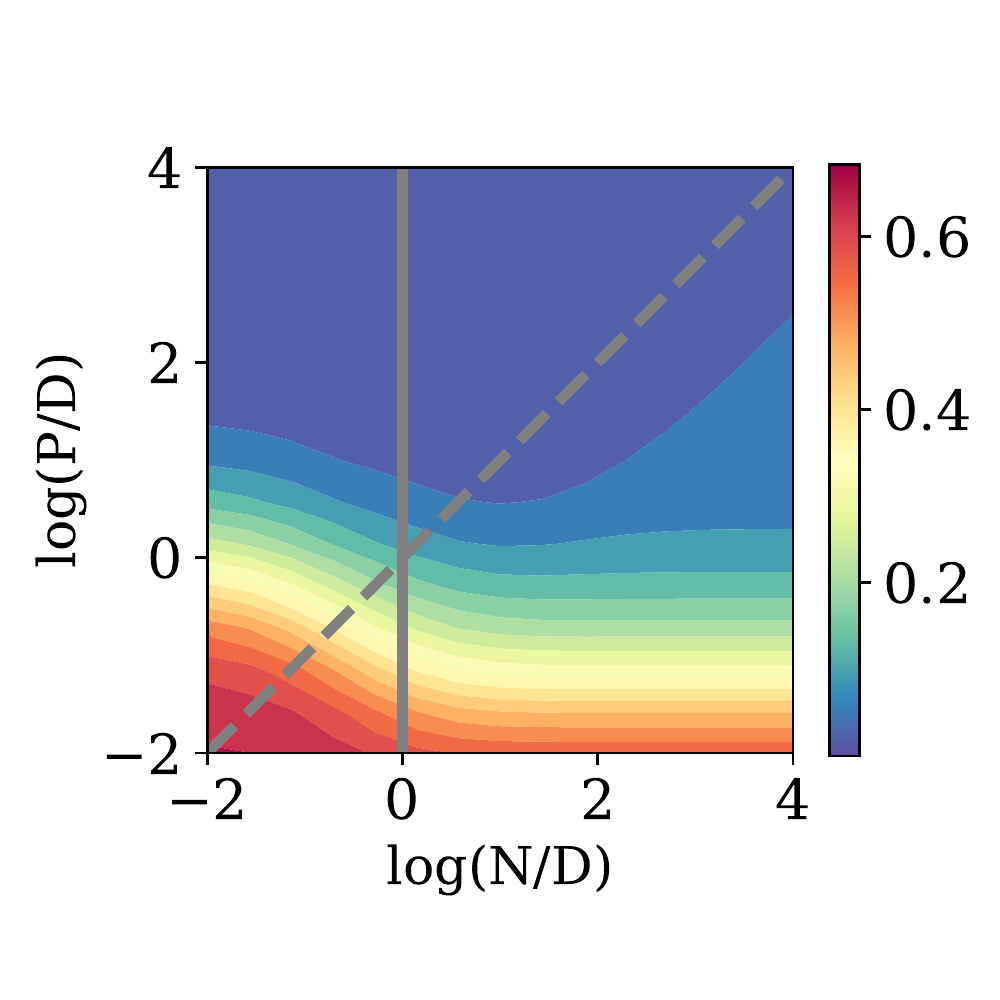}
    \caption{Aligned, $\Delta=0$}
    \end{subfigure}
    \begin{subfigure}[b]{.24\textwidth}	   
    \includegraphics[width=\linewidth]{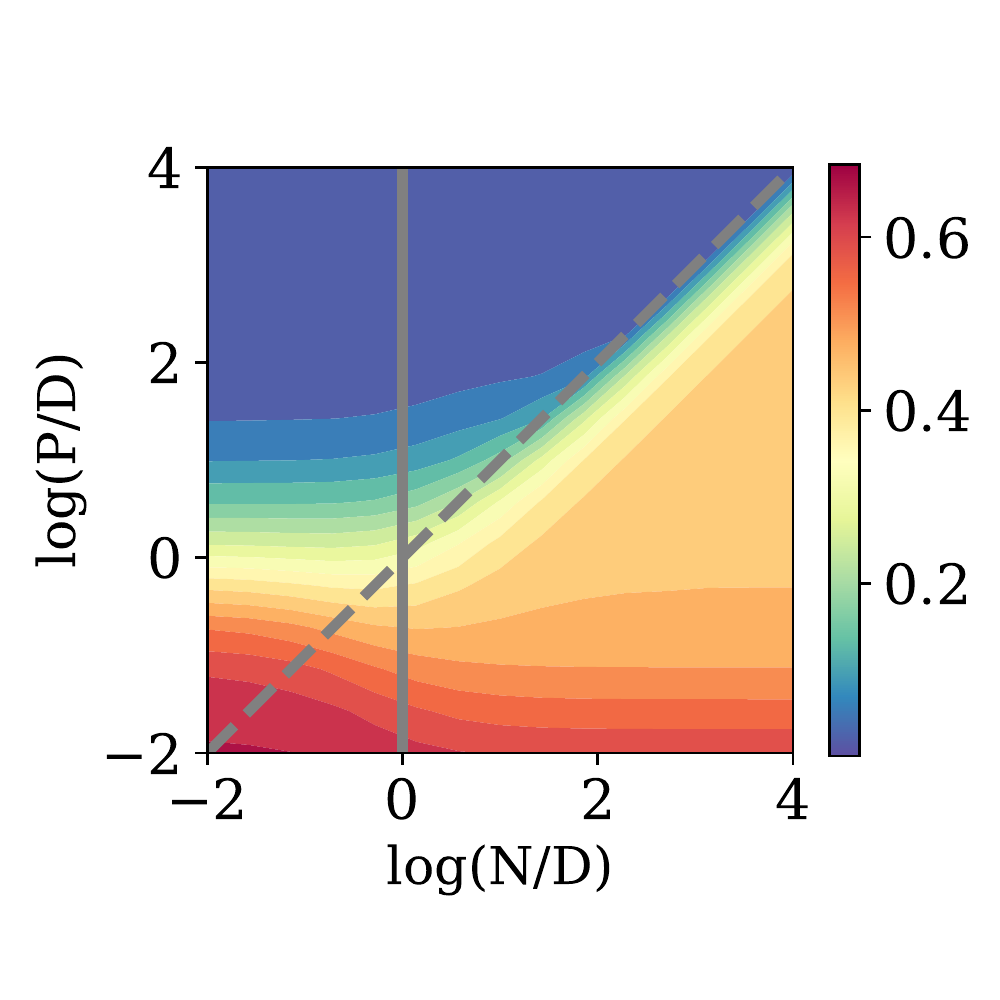}
    \caption{Aligned, $\Delta=0.4$}
    \end{subfigure}
    \begin{subfigure}[b]{.24\textwidth}	   
    \includegraphics[width=\linewidth]{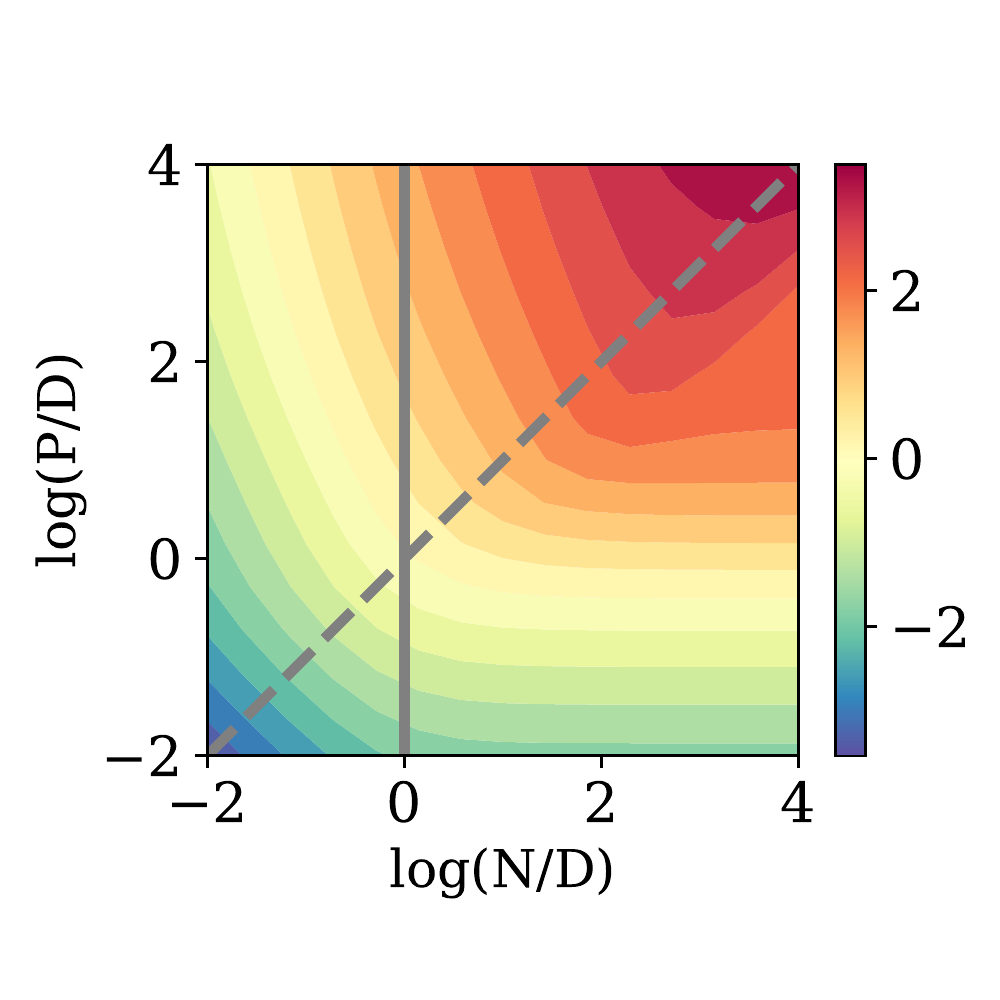}
    \caption{Isotropic, noiseless}
    \end{subfigure}
    \begin{subfigure}[b]{.24\textwidth}	   
    \includegraphics[width=\linewidth]{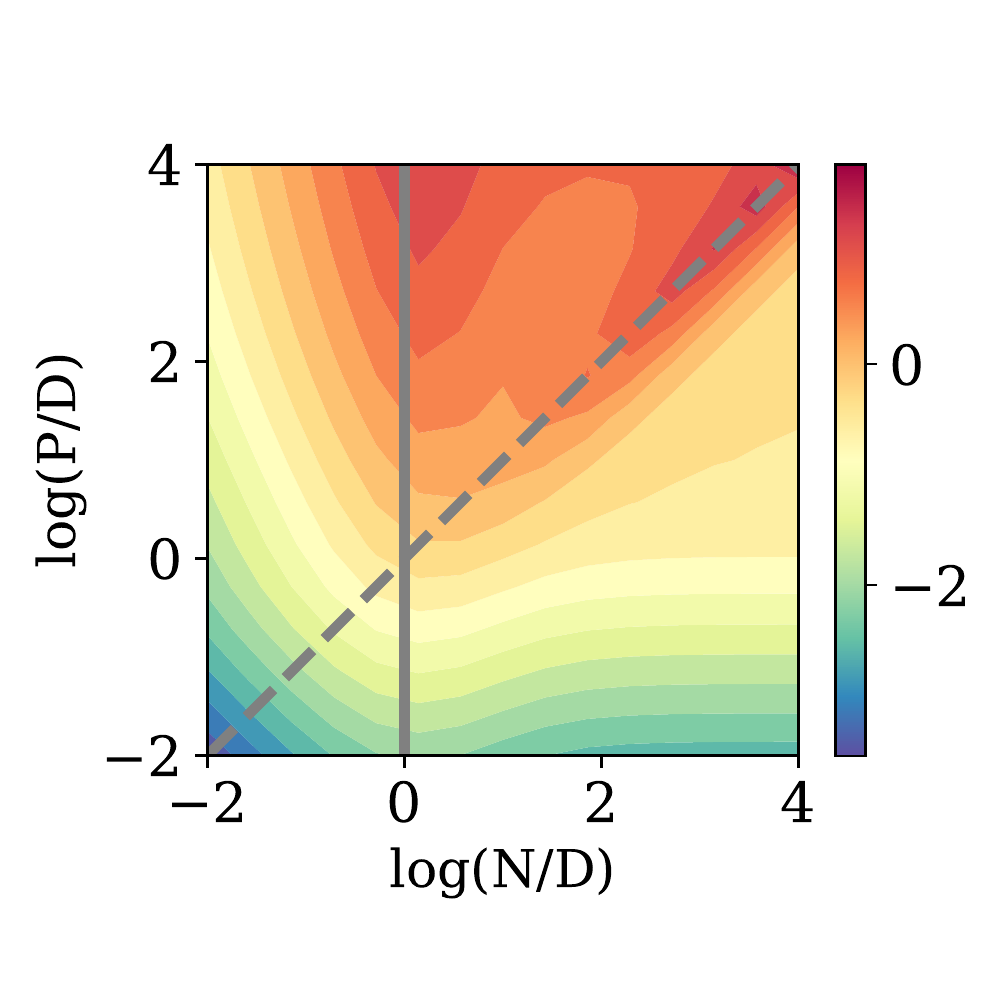}
    \caption{Isotropic, $\Delta=0.4$}
    \end{subfigure}
    \begin{subfigure}[b]{.24\textwidth}	   
    \includegraphics[width=\linewidth]{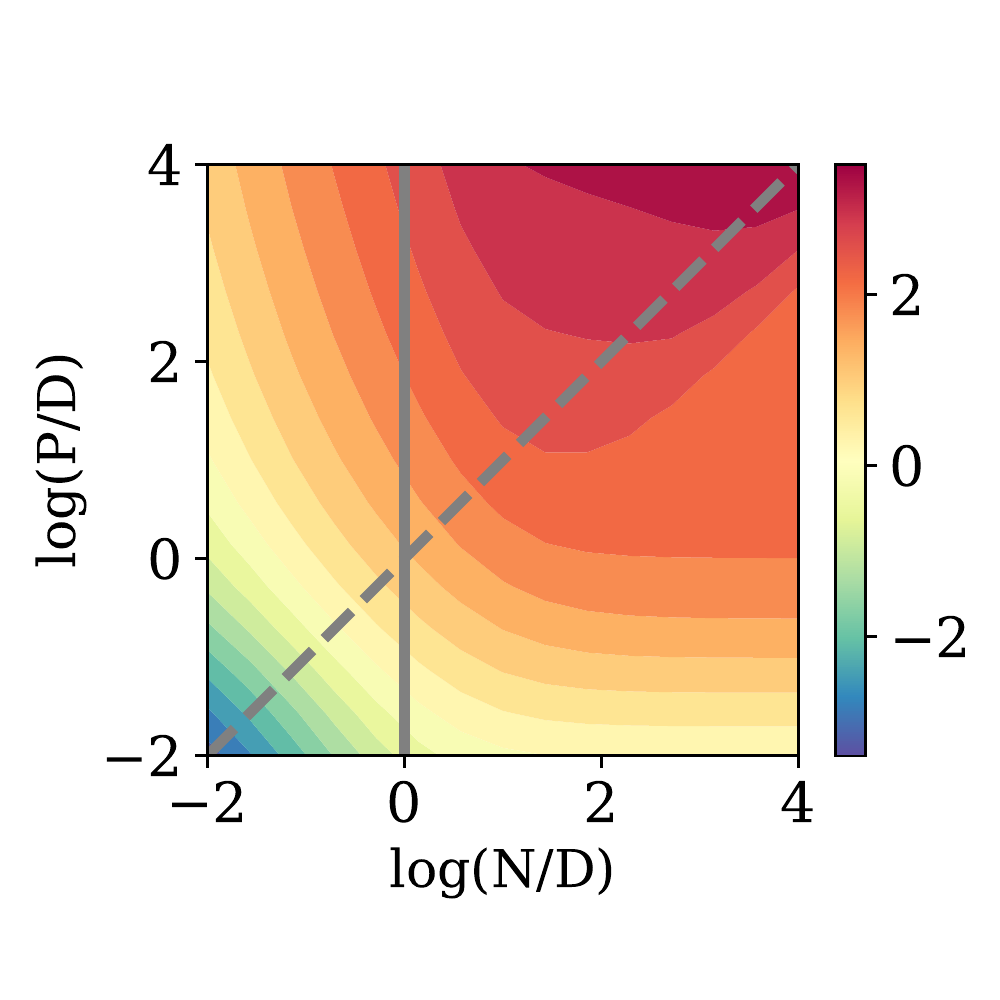}
    \caption{Aligned, $\Delta=0$}
    \end{subfigure}
    \begin{subfigure}[b]{.24\textwidth}	   
    \includegraphics[width=\linewidth]{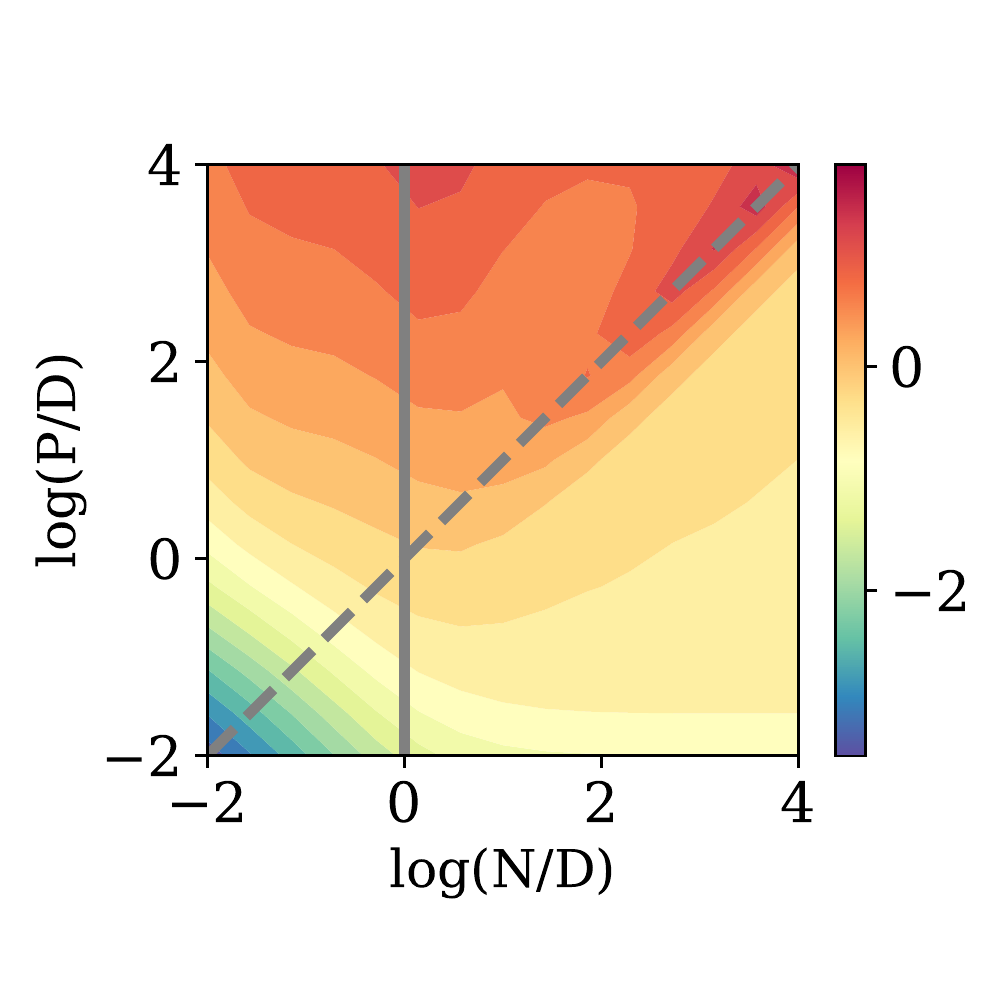}
    \caption{Aligned, $\Delta=0.4$}
    \end{subfigure}
    \begin{subfigure}[b]{.24\textwidth}	   
    \includegraphics[width=\linewidth]{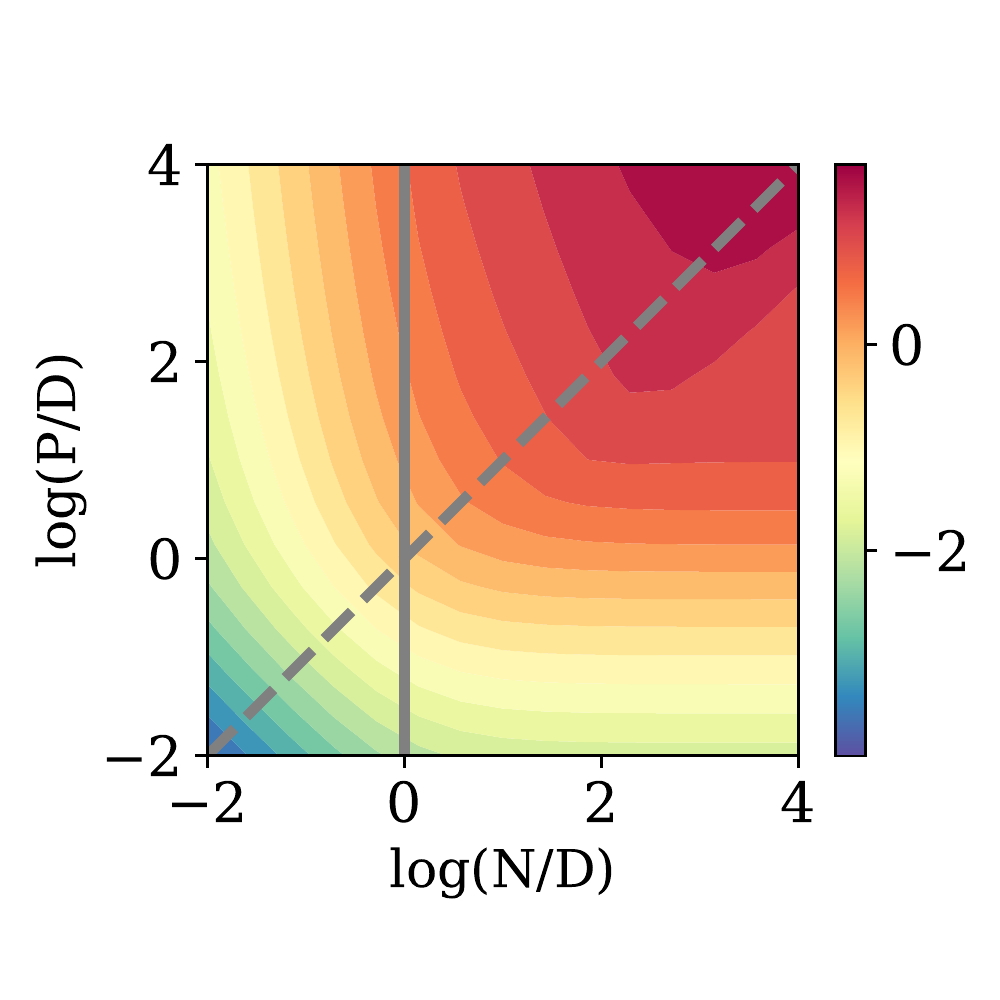}
    \caption{Isotropic, noiseless}
    \end{subfigure}
    \begin{subfigure}[b]{.24\textwidth}	   
    \includegraphics[width=\linewidth]{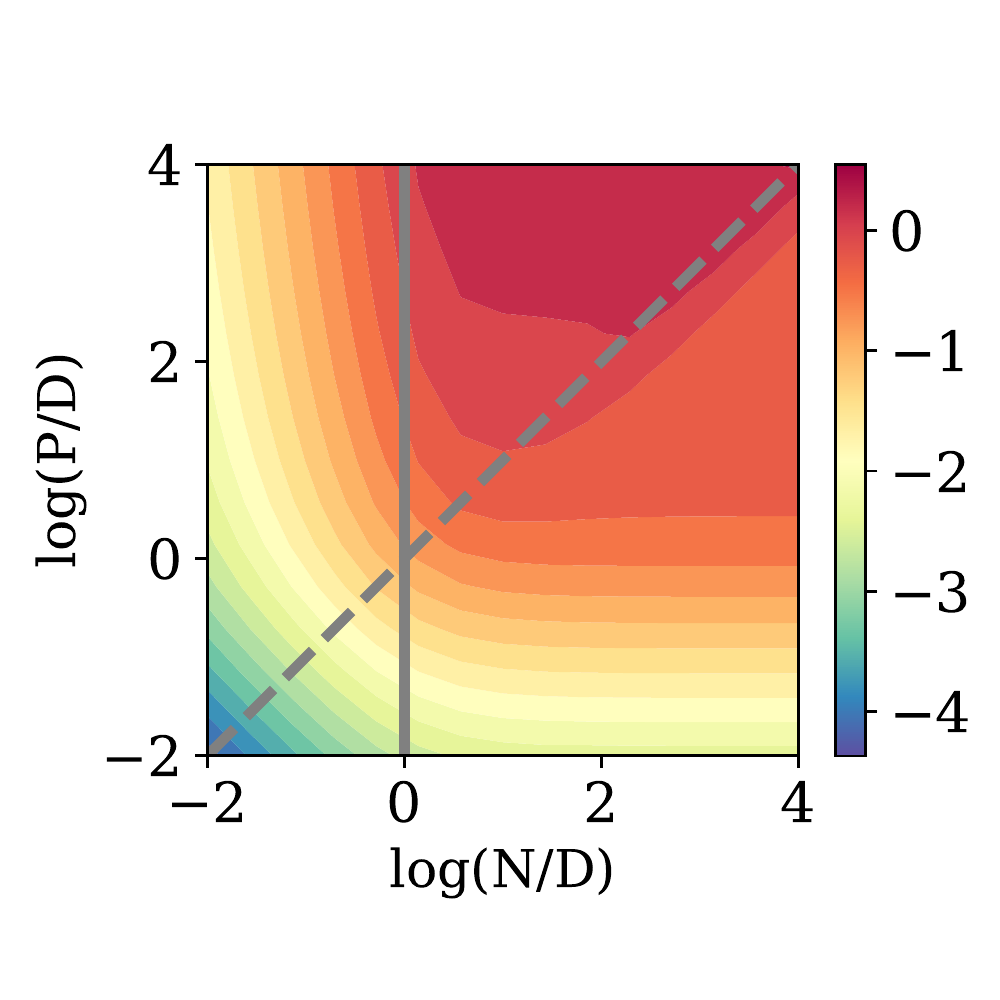}
    \caption{Isotropic, $\Delta=0.4$}
    \end{subfigure}
    \begin{subfigure}[b]{.24\textwidth}	   
    \includegraphics[width=\linewidth]{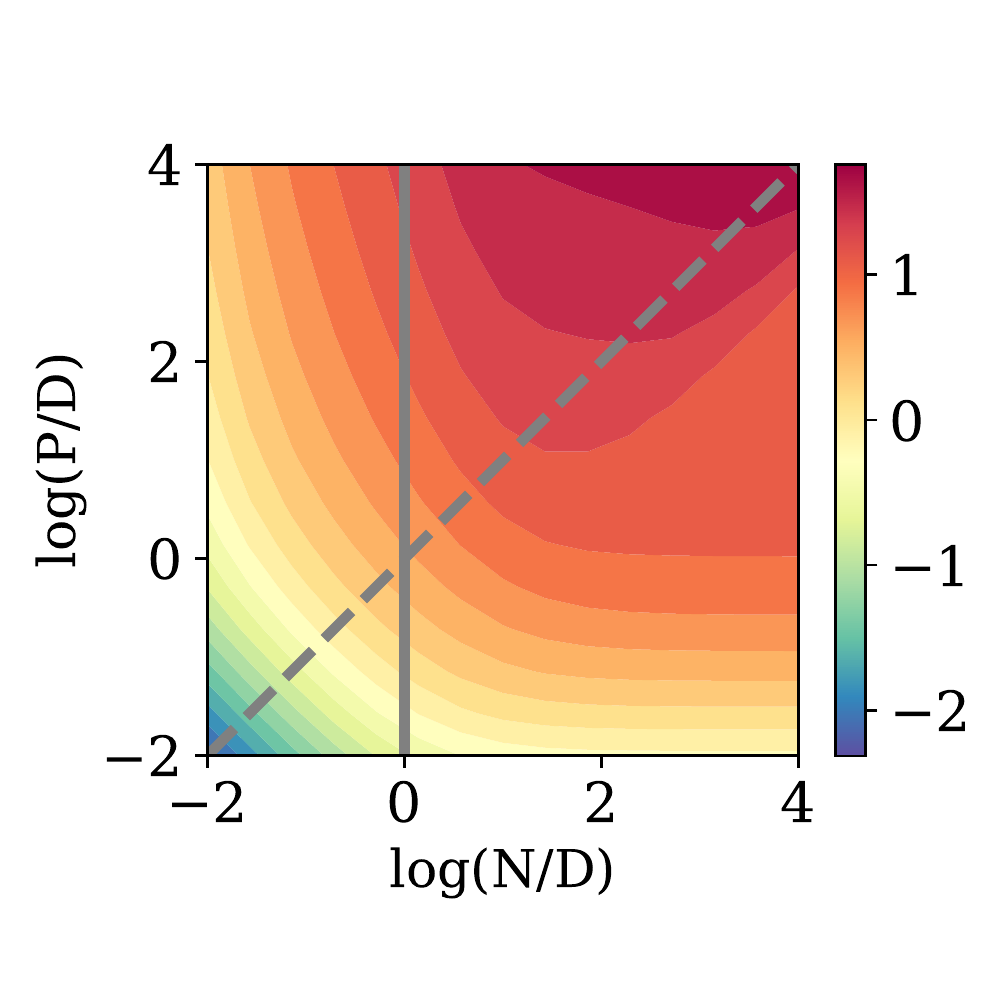}
    \caption{Aligned, $\Delta=0$}
    \end{subfigure}
    \begin{subfigure}[b]{.24\textwidth}	   
    \includegraphics[width=\linewidth]{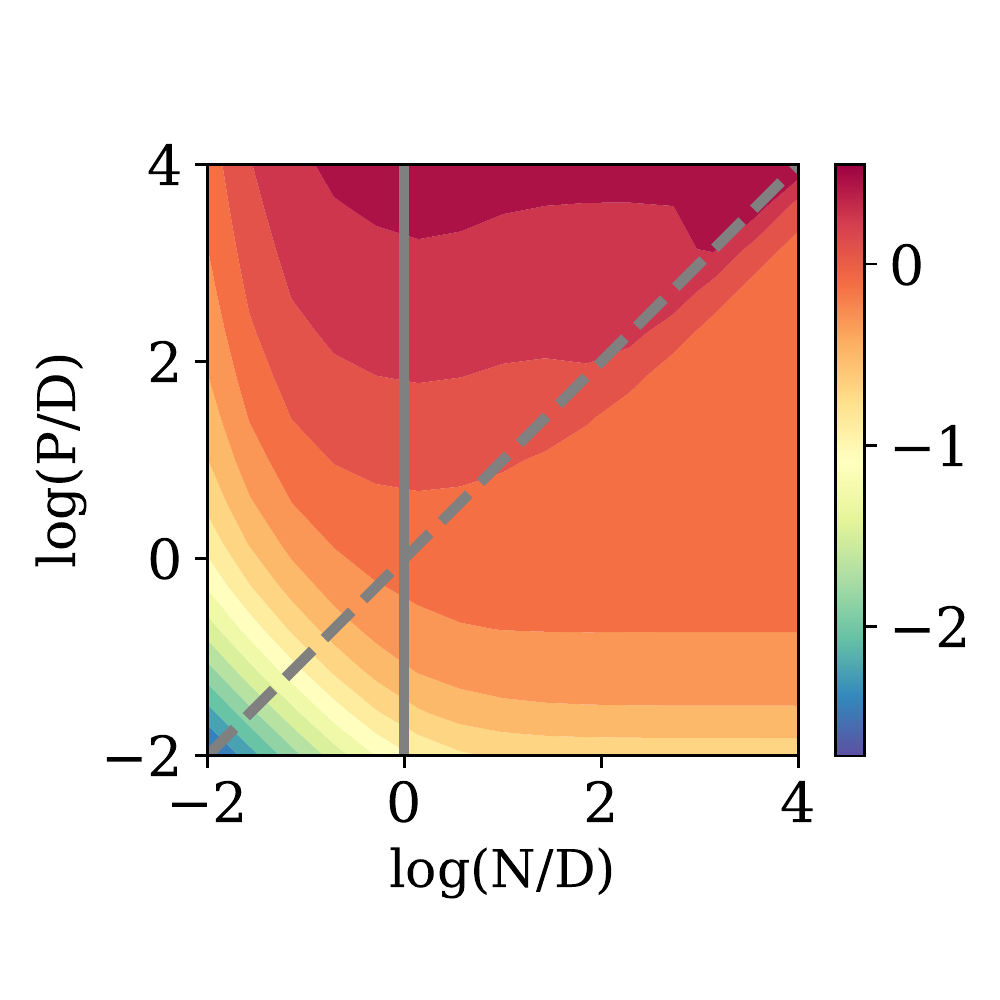}
    \caption{Aligned, $\Delta=0.4$}
    \end{subfigure}
    \caption{\textbf{Logistic loss phase spaces.} We studied the classification task for $\sigma = \mathrm{ReLU}$, $\lambda=0.1$. In the Aligned phase spaces we set $\phi_1=0.01$, $r_x=100$, $r_\beta=100$. \textit{First row:} test error. \textit{Second row:} train error. \textit{Third row:} Q. \textit{Fourth row:} M. The solid and dashed grey lines represent the $N=D$ and $N=P$ lines, where one can find overfitting peaks~\cite{d2020triple}. For $Q$ and $M$, the colormaps are logarithmic.}
    \label{fig:phase-spaces-logistic}
\end{figure}

\paragraph{Comparing square and logistic}
In Figs.~\ref{fig:phase-spaces-square} and \ref{fig:phase-spaces-logistic}, we show how the various observables of interest evolve in the $(N,P)$ phase space, respectively from square loss and logistic loss. To show that our results generalize to different activation functions and regularization levels, we choose $\sigma=\mathrm{ReLU}$ and $\lambda=0.1$. We make the following observations:
\begin{enumerate}
    \item \textbf{Test error}: the phase-space is almost symmetric in the isotropic setup, with the double descent peak clearly visible at $N=P$ (dashed grey line) for square loss but strongly attenuated for logistic loss. Interestingly, for logistic loss, the peak appears at $N>P$ in the noiseless setup as observed in~\cite{gerace2020generalisation}, but shifts to $N=P$ in presence of noise. In the anisotropic setup, the phase space dissymetrizes, with a wide overfitting region emerging in the overparametrized regime around $N=D$ (solid grey line), as observed in~\cite{d2020triple} for isotropic regression tasks. This overfitting is strongly regularized for logistic loss, explaining its superiority on structured datasets
    \item \textbf{Train loss}: Here a strong difference appears between square loss and logistic loss. For square loss, the overparametrized region $P>N>D$ reaches zero training loss, and the interpolation threshold clearly appears at $P=N$. For logistic loss, the interpolation threshold depends more strongly on the data structure: it shifts up when we increase the label noise and shifts down when we increase the anisotropy.
    \item \textbf{Order parameter $Q$}: recall that this observable represents the variance of the student's outputs. Here the phase space looks completely different for the two loss functions. For square loss, the phase space is almost symmetric in the noiseless isotropic setup and dissymetrizes in presence of noise or anisotropy, with a peak appearing at $N=D$, forming the ``linear peak'' in the test error~\cite{d2020triple}. For logistic loss, the phase space is not symmetric: $Q$ becomes very large in the overparametrized regime, leading to overconfidence, as observed in Fig.~\ref{fig:norm} of the main text.
    \item \textbf{Order parameter $M$}: recall that this observable represents the covariance between the student's outputs and the teacher's outputs, i.e. the dot product of their corresponding vectors. Again, behavior is very different for square loss and logistic loss. For square loss, the covariance increases symmetrically when increasing $P$ or $N$, reaching its maximal value respectively at $P=D$ and $N=D$ in the isotropic setup, or earlier in the aligned setup. For logistic loss, the phase space is more complex due to the fact that the norm of the student diverges in the overparametrized regime.
\end{enumerate}

\begin{figure}[htb]
    \centering
    \begin{subfigure}[b]{.24\textwidth}	   
    \includegraphics[width=\linewidth]{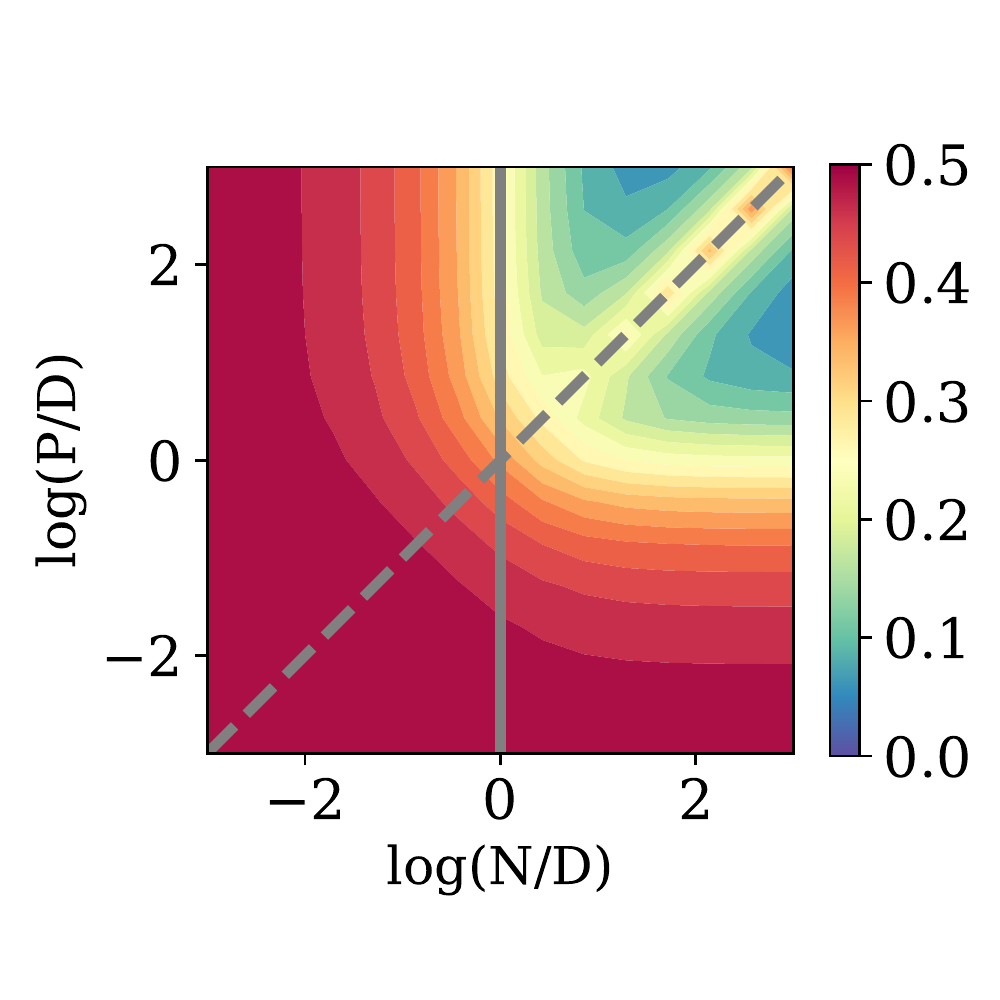}
    \caption{$r_x=1$}
    \end{subfigure}
    \begin{subfigure}[b]{.24\textwidth}	   
    \includegraphics[width=\linewidth]{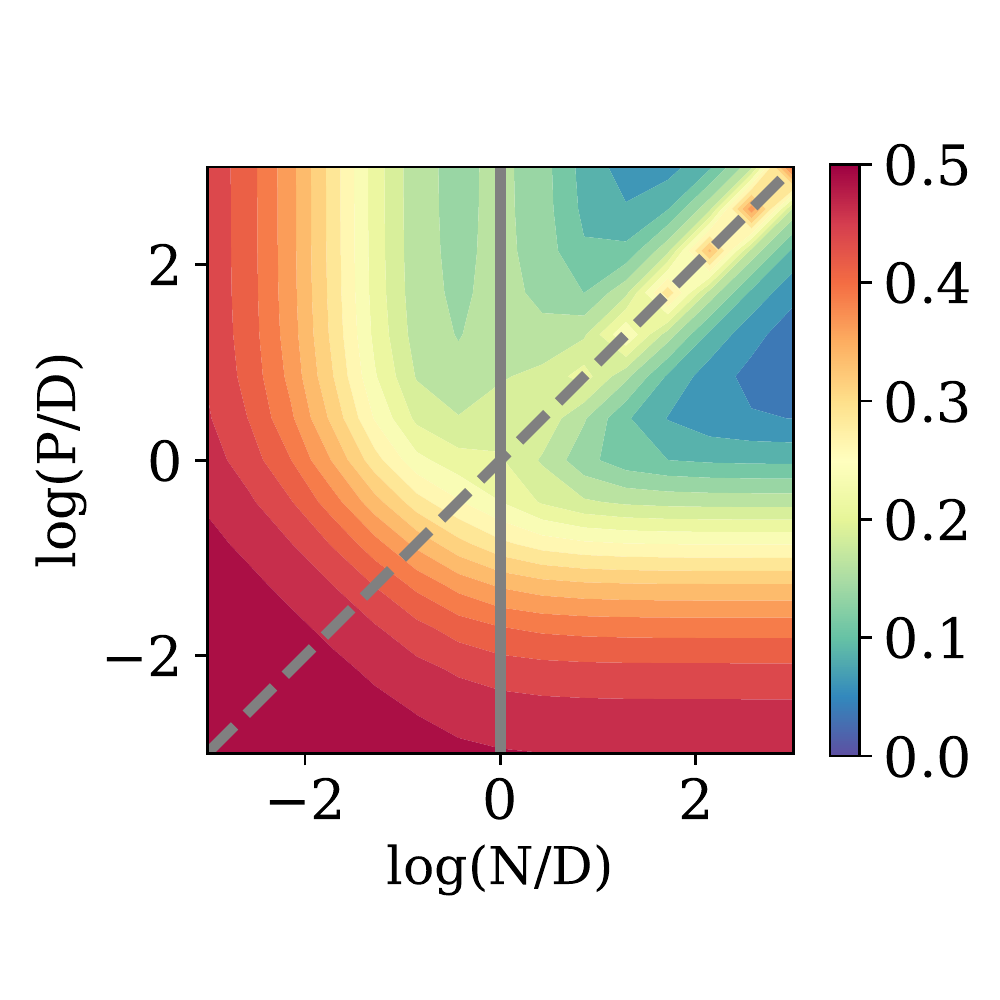}
    \caption{$r_x=10$}
    \end{subfigure}
    \begin{subfigure}[b]{.24\textwidth}	   
    \includegraphics[width=\linewidth]{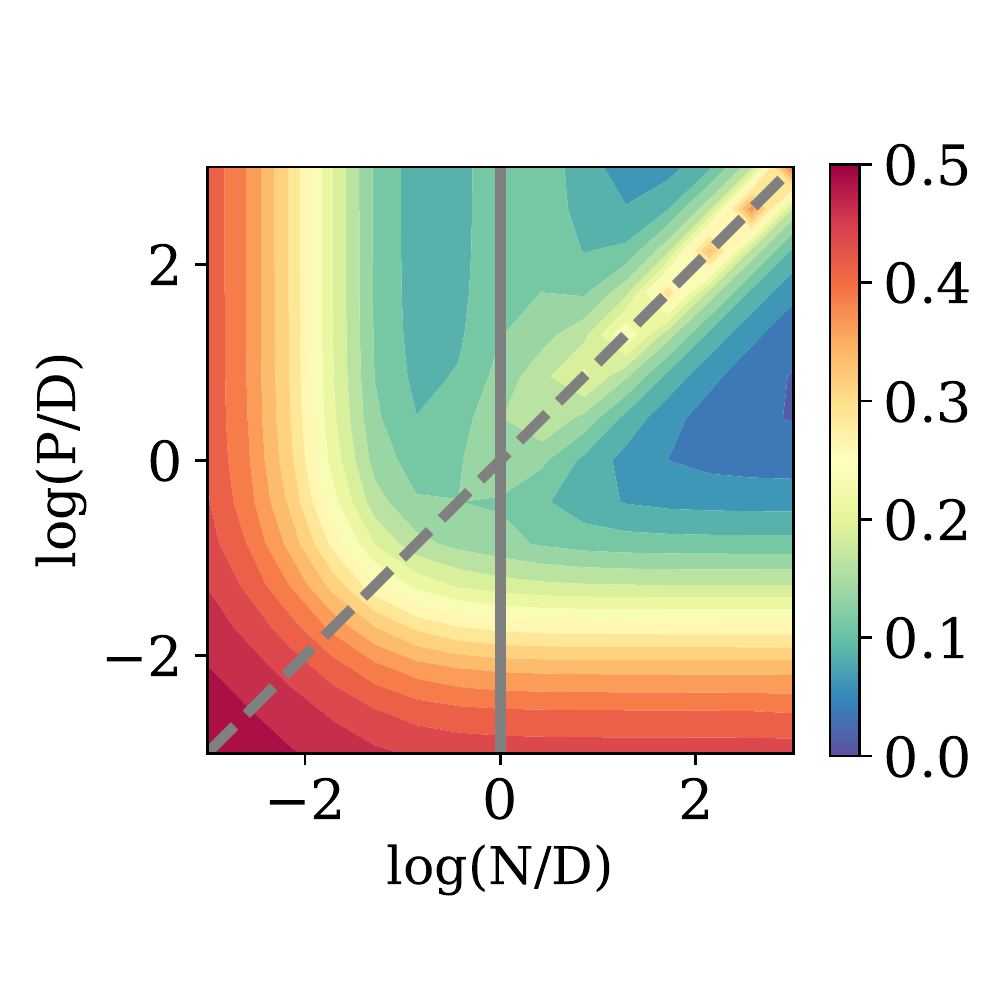}
    \caption{$r_x=100$}
    \end{subfigure}
    \begin{subfigure}[b]{.24\textwidth}	   
    \includegraphics[width=\linewidth]{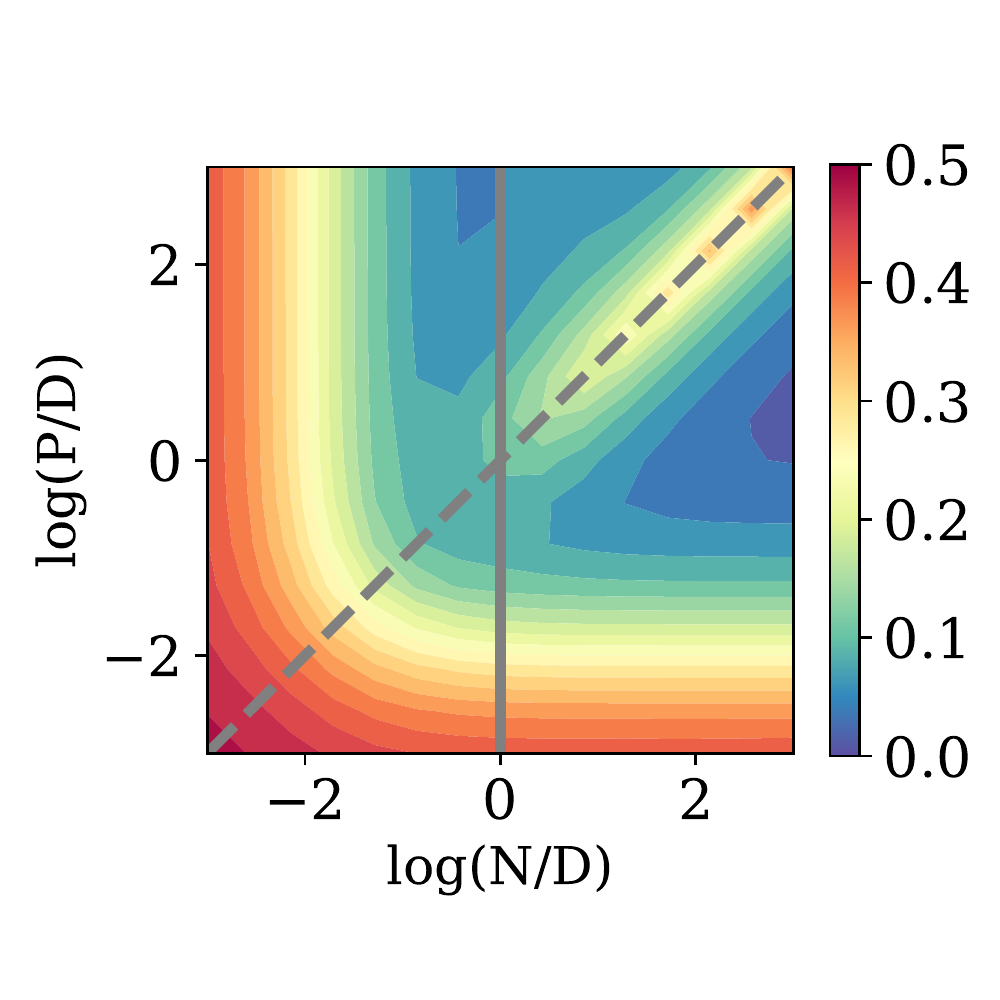}
    \caption{$r_x=1000$}
    \end{subfigure}
    \caption{Increasing the saliency of the first subspace from $1$ to $\infty$, the asymmetry forms then vanishes as the data becomes effectively isotropic in smaller dimension. We study the classification task for $\sigma = \mathrm{ReLU}$, $\lambda=0.1$, $\phi_1=0.01$, $r_\beta=1000$ and $\Delta=0$.}
    \label{fig:increasing-rx}
\end{figure}

\paragraph{Varying the anisotropy}
Fig. \ref{fig:increasing-rx} illustrates the modification of the phase space of the random features model trained on the strong and weak features model as the saliency $r_x$ of the $\phi_1 D = 0.01 D$ relevant features ($r_\beta=1000$) is gradually increased. At $r_x=1$ (panel (a)), the data is isotropic and the phase space is symmetric. When $r_x\to\infty$, all the variance goes in the salient subspace and we are left with an isotropic task of dimensionality $\phi_1 D$: the phase space is symmetric again (panel (d)). In between these two extreme scenarios, we see the asymmetry appear in the phase space under the form of an overfitting peak at $N=D$, as the irrelevant features come into play (panels (b) and (c)).

\clearpage
\section{Analytical derivations}
\label{app:replica}

The following sections present the analytical derivations of this work. First, we provide an outline of the computation of the generalization error. Second, we describe our extension of the Gaussian Equivalence Theorem to anisotropic data, a key ingredient to handle the strong and weak features model. The three following sections develop the steps of the replica computation, from the Gibbs formulation of the learning objective, to the set of self-consistent equations to obtain the order parameters. Lastly, we explain how to also obtain the asymptotic training loss from the output of the replica computation. 

\subsection{Outline}

In the main text, we study the strong and weak features scenario with two blocks, and only for classification tasks, but our derivation will be performed here in full generality. 

\paragraph{Setup and notations}
We recall notations:
$D$ is the input dimension, $P$ is the number of random features and $N$ is the number of training examples. We consider the high-dimensional limit where $D,P,N\to\infty$ and $\gamma = \frac{D}{P}$, $\alpha = \frac{N}{P}$ are of order one. The inputs $\bs x\in \mathbb R^D$ and the elements of the teacher vectors $\bs \beta \in \mathbb R^D$ are sampled from a block-structured covariance matrix :
\begin{align}
&\bs x\sim\mathcal{N}(0, \Sigma_x), \quad  \Sigma_x = \begin{bmatrix}
\sigx{1} \mathbb I_{\phi_1 D} & 0 & 0\\
0 & \sigx{2} \mathbb I_{\phi_2 D} & 0\\
0 & 0 & \ddots
\end{bmatrix}\\
&\bs \beta \sim\mathcal{N}(0, \Sigma_\beta), \quad  \Sigma_\beta = \begin{bmatrix}
    \sigb{1} \mathbb I_{\phi_1 D} & 0 & 0 \\
0 & \sigb{2} \mathbb I_{\phi_2 D} & 0\\
0 & 0 & \ddots
\end{bmatrix}\\
\end{align}

The output of the random feature model is given by:
\begin{equation}
    \hat y_\mu = \hat f\left(\sum_{i=1}^P\bs w_i \sigma\left(\frac{\bs F_i \cdot \bs x_\mu}{\sqrt{D}}\right)\right)\equiv \hat f\left(\bs w\cdot \bs z_\mu \right),
\end{equation} 
where $\sigma(\cdot)$ is a pointwise activation function, $\hat f$ is an output function and we introduced the notation $\bs z_\mu \in \sR^P$.
Elements of $\bs F$ are drawn i.i.d from $\mathcal{N}(0,1)$.  The labels are given by a linear ground truth, modulated by an output function $f^0$ and possibly corrupted by noise through a probabilistic channel $\gP$:
\begin{equation}
    \label{eq:teacher-def}
    y_\mu \sim \gP \left( \cdot \left\vert  f^0\left(\frac{\bs \beta\cdot\bs x_\mu}{\sqrt D}\right)\right.\right).
\end{equation}

The second layer weights, i.e. the elements of $\bs w \in \sR^P$, are trained by minimizing an $\ell_2$-regularized loss on $N$ training examples $\{\bs x_\mu\in\R^D\}_{\mu=1\ldots N}$ :
\begin{align}
    \hat{\boldsymbol{w}}=\underset{\boldsymbol{w}}{\operatorname{argmin}} \left[\epsilon_t(\bs w) \right],\quad
    \epsilon_t(\bs w) = \sum_{\mu=1}^{N} \ell\left(y^{\mu}, \bs w \cdot \bs z_\mu \right)+\frac{\lambda}{2}\|\boldsymbol{w}\|_{2}^{2}.
\end{align}
We consider both regression tasks, where $\hat f = f^0 = \text{id}$, and binary classification tasks, $\hat f = f^0 = \text{sign}$. In the first case, $\ell$ is the square loss and the noise is additive. In the second case, $\ell$ can be any type of loss (square, hinge, logistic) and the noise amounts to random label flipping.

The generalization error is given by
\begin{align}
    \epsilon_{g}=\frac{1}{2^{k}} \E_{\boldsymbol{x}, y}\left[\left(\hat{y}\left(\boldsymbol{x}\right)-y\right)^{2}\right]
\end{align}
with $k = 1$ for regression tasks (mean-square error) and $k = 2$ for binary classification tasks (zero-one error).

In the following, we denote as $\bs x|_i, \bs \beta|_i$ the orthogonal projection of $\bs x$ and $\bs \beta$ onto the subspace $i$ of $\mathbb R^D$. For example, $\bs x|_1$ amounts to the first $\phi_1 D$ components of $\bs x$.

\paragraph{Steps of the derivation of the generalization error}
The key observation is that the test error can be rewritten, in the high-dimensional limit, in terms of so-called \emph{order parameters}:
\begin{align}
\epsilon_{g} &=\frac{1}{2^{k}} \E_{\bs{x}^{\mathrm{new}}, y^{\mathrm{new}}}\left(y^{\mathrm{new}}-\hat{f}\left(\frac{1}{\sqrt P}\sigma\left(\frac{\mathrm{F}^{\top} \bs{x}^{\mathrm{new}}}{\sqrt D}\right) \cdot \hat{\bs{w}}\right)\right)^{2} \\
&=\frac{1}{2^{k}} \int \dd \nu \int \dd \lambda P(\nu, \lambda)(f^0(\nu)-\hat{f}(\lambda))^{2}
\end{align}
where we defined 
\begin{align}
    \lambda = \frac{1}{\sqrt P}\sigma\left(\frac{\mathrm{F}^{\top} \bs{x}^{\mathrm{new}}}{\sqrt D}\right) \cdot \hat{\bs{w}}\in \sR, \quad \nu=\frac{1}{\sqrt{D}} \bs{x}^{\mathrm{new}} \cdot \bs\beta \in \sR
\end{align}
and the joint probability distribution
\begin{align}
    P(\nu,\lambda) = \E_{\bs x^{\mathrm{new}}}\left[\delta\left(\nu-\frac{1}{\sqrt D}\bs{x}^{\mathrm{new}} \cdot \bs\beta \right) \delta\left(\lambda-\frac{1}{\sqrt P}\sigma\left(\frac{\mathrm{F}^{\top} \bs{x}^{\mathrm{new}}}{\sqrt D}\right) \cdot \hat{\bs{w}}\right)\right].
\end{align}

The key objective to calculate the test error is therefore to obtain the joint distribution $P(\nu,\lambda)$. To do so, our derivation is organized as follows.
\begin{enumerate}
    \item In Sec.~\ref{app:get}, we adapt the Gaussian equivalence theorem~\cite{el2010spectrum,mei2019generalization, goldt2019modelling, hu2020universality} to the case of anisotropic data. The latter shows that $P(\nu,\lambda)$ is a joint gaussian distribution whose covariance depends only on the following \emph{order parameters}: $$m_{s,i}=\frac{1}{D} \bs{s}|_i \cdot \bs\beta|_i, \quad q_{s,i}=\frac{1}{D} \bs{s}|_i \cdot \bs{s}|_i, \quad q_{w}=\frac{1}{P} \hat{\bs{w}} \cdot \hat{\bs{w}},$$ where we denoted $\bs s = \frac{1}{\sqrt P}\bs F \bs \hat{\bs w}\in\mathbb R^D$ and the index $i$ refers to the subspace of $\mathbb R^D$ the vectors are projected onto.
    \item In Sec.~\ref{app:gibbs}, we recast the optimization problem as a Gibbs measure over the weights, from which one can sample the average value of the order parameters $m_s, q_s, q_w$. Thanks to the convexity of the problem, this measure concentrates around the solution of the optimization problem in the limit of high temperature, see~\cite{gerace2020generalisation}.
    \item In Sec.~\ref{app:rmt}, we leverage tools from random matrix theory to derive the saddle-point equations allowing the obtain the values of the order parameters
    \item In Sec.~\ref{app:saddle-point}, we give the explicit expressions of the saddle-point equations for the two cases studied in the main text: square loss regression and classification
    \item In Sec.~\ref{app:train-derivation}, we also show how to obtain the train error from the order parameters.
\end{enumerate}

Once the order parameters are known, the joint law of $\lambda,\nu$ is determined as explained in the main text:
\begin{align}
    P(\lambda, \nu) = \mathcal{N}(0, \Sigma), \quad 
\Sigma=\left(\begin{array}{cc}
\rho & M \\
M & Q
\end{array}\right).
\end{align}
It is then easy to perform the Gaussian integral giving the generalization error.

For the regression problem where $\hat f = f^0 =  \mathrm{id}$, we have:
\begin{align*}
\epsilon_{g}=\frac{1}{2}\left(\rho+Q-2 M\right)
\end{align*}

For the classification problem where $\hat f = f^0 = \mathrm{sgn}$, we have:
\begin{align*}
\epsilon_{g}=\frac{1}{\pi} \cos ^{-1}\left(\frac{M}{\sqrt{\rho Q}}\right)
\end{align*}.

\subsection{The anisotropic Gaussian Equivalence Theorem}
\label{app:get}

We now present the derivation of the anisotropic Gaussian Equivalence Theorem. This is a key ingredient for the replica analysis presented in the next section. 

Define, for a Gaussian input vector $\bs x\in \mathbb R^D$, the family of vectors indexed by $a = 1\ldots r$:
\begin{align}
    \lambda^a=\frac{1}{\sqrt{P}} \bs{w}^{a} \cdot \sigma\left(\frac{\bs F \bs{x}}{\sqrt D}\right) \in \sR, \quad \nu=\frac{1}{\sqrt{D}} \bs{x} \cdot \bs\beta \in \sR
\end{align}

Using the Gaussian equivalence principle, we can obtain expectancies for this family in the high-dimensional limit:
\begin{align}
(\nu, \lambda^a) \sim \mathcal{N}(0,\Sigma), \quad 
\Sigma^{a b}=\left(\begin{array}{cc}
\rho & M^{a} \\
M^{a} & Q^{a b}
\end{array}\right) \in \sR^{(r+1) \times(r+1)}
\end{align}

\begin{align}
&\rho= \sum_i \phi_i \beta|_i \sigx{i}, \quad  M^{a}= \kappa_{1} \sum_i \sigx{i} m_{s,i}^{a}, \quad Q^{a b}=  \kappa_{1}^2 \sum_i\sigx{i} q_{s,i}^{ab} +  \kappa_{\star}^{2} q_w^{ab}  \\
&m_{s,i}^{a}=\frac{1}{D} \bs{s}^{a}|_i \cdot \bs\beta|_i, \quad q_{s,i}^{a b}=\frac{1}{D} \bs{s}^{a}|_i \cdot \bs{s}^{b}|_i, 
\quad q_{w}^{a b}=\frac{1}{P} \bs{w}^{a} \cdot \bs{w}^{b}
\end{align}

where
\begin{align}
& \kappa_{0}=\underset{z\sim\mathcal{N}(0,r)}{\E}[\sigma(z)], 
\quad \kappa_{1}=\frac{1}{r}\underset{z\sim\mathcal{N}(0,r)}{\E}[z \sigma(z)]], 
\quad \kappa_{\star}=\sqrt{\underset{z\sim\mathcal{N}(0,\sum_i \sigx{i} \phi_i)}{\E}\left[\sigma(z)^{2}\right]-\kappa_{0}^{2}- r \kappa_{1}^{2}}\\
&r=\sum_i \phi_i \sigx{i}, \quad \bs{s}^{a}=\frac{1}{\sqrt{P}} \bs F \bs{w}^{a} \in \sR^{D}, \quad \bs s^a_i = P_{E_i} \bs s^a, \quad \bs \beta|_i = P_{E_i} \bs \beta \\
\end{align}

\paragraph{Isotropic setup}

Let us start in the setup where we only have one block, of unit variance.

By rotational invariance, we can write :
\begin{align}
    \E_{\bs x}\left[\sigma\left(\frac{\bs x \cdot F_\mu}{\sqrt D}\right)\sigma\left(\frac{\bs x \cdot F_\nu}{\sqrt D}\right)\right]&= f\left(\frac{F_\mu \cdot F_\nu}{D}\right) \equiv f(q_{\mu\nu})\\
    &=\delta_{\mu\nu} f(1) + (1-\delta_{\mu\nu}) \left(f(0) + f'(0) q_{\mu\nu}\right) + o\left(q_{\mu\nu}\right)\\
    &=f(0) + f'(0) q_{\mu\nu} + \delta_{\mu\nu} (f(1)-f(0)-f'(0))
\end{align}

Therefore, the expectancy is the same as the "Gaussian equivalent model" where we replace
\begin{align}
    \sigma\left(\frac{\bs x \cdot F_\mu}{\sqrt D}\right)\rightarrow \sqrt{f(0)}  + \sqrt{f'(0)}\left( \frac{\bs x \cdot F_\mu}{\sqrt D}\right) + \sqrt{(f(1)-f(0)-f'(0)} \xi_\mu, \quad \xi_\mu \sim \mathcal{N}(0,1) 
\end{align}

What remains is to find the expression of $f(0), f(1)$ and $f'(0)$. If $q_{\mu\nu}=1$, then $F_\mu$ and $F_\nu$ are the same vector and clearly $f(1) = \E_{z} \left[ \sigma(z)^2 \right]$. 

Otherwise, we use the fact that $x_\mu = \frac{\bs x \cdot F_\mu}{\sqrt D}$ and $x_\nu = \frac{\bs x \cdot F_\nu}{\sqrt D}$ are correlated gaussian variables:
\begin{align}
    \mathbb E\left[ x_\mu^2\right] = 1, \quad\quad
    \mathbb E\left[ x_\mu x_\nu\right] = \frac{F_\mu \cdot F_\nu}{D} \equiv q_{\mu\nu} \sim O\left(\frac{1}{\sqrt D}\right)
\end{align}
Therefore we can parametrize as follows,
\begin{align}
    x_\mu = \eta_\mu \sqrt{r-q_{\mu\nu}} + z \sqrt{q_{\mu\nu}} \sim \eta_\mu (1-\frac{1}{2} q_{\mu\nu}) + z \sqrt{q_{\mu\nu}}\\
    x_\nu = \eta_\nu \sqrt{r-q_{\mu\nu}} + z \sqrt{q_{\mu\nu}} \sim \eta_\mu (1-\frac{1}{2} q_{\mu\nu}) + z \sqrt{q_{\mu\nu}}\\
\end{align}
where $\eta_\mu, \eta_\nu,z\sim\mathcal{N}(0,1)$. To calculate $f(0), f'(0)$, we can expand the nonlinearity,
\begin{align}
    f(q_{\mu\nu}) &\equiv f(0) + f'(0)q_{\mu\nu} + o(q_{\mu\nu}^2)\\
    &=\mathbb E\left[ \left(\sigma(\eta_\mu) + \sigma'(\eta_\mu) \left(-\frac{q}{2}\eta_\mu + z\sqrt q_{\mu\nu}\right) + \frac{\sigma''(\eta_\mu)}{2}\left( z^2q_{\mu\nu}\right)\right) \right.\\
    &\left. \left(\sigma(\eta_\nu) + \sigma'(\eta_\nu) \left(-\frac{q}{2}\eta_\nu + z\sqrt q_{\mu\nu}\right) + \frac{\sigma''(\eta_\nu)}{2}\left( z^2q_{\mu\nu}\right)\right)\right]\\
    &= \mathbb E_\eta \left[ \sigma(\eta)\right]^2 + q_{\mu\nu} \left( \mathbb E[\sigma'(\eta)]^2 + \mathbb E[\sigma''(\eta)]\mathbb E [\sigma(\eta)] - \mathbb E[\sigma'(\eta)\eta] \mathbb E[\sigma(\eta)] \right) + o(q_{\mu\nu}^2)\\
\end{align}

But since $\mathbb E[\sigma(\eta)\eta] = \mathbb E[\sigma'(\eta)]$, the two last terms cancel out and we are left with
\begin{align}
    f(0) = \mathbb E[\sigma(z)]^2, \quad\quad f'(0) = \mathbb E[\sigma(z)\eta]^2
\end{align}

\begin{align}
    \E_{\bs x}\left[\sigma\left(\frac{\bs x \cdot F_\mu}{\sqrt D}\right)\sigma\left(\frac{\bs x \cdot F_\nu}{\sqrt D}\right)\right] = \kappa_0^2 +  \kappa_1 ^2 \frac{F_\mu \cdot F_\nu}{D} + \kappa_\star^2 \delta_{\mu\nu}
\end{align}

\paragraph{Anisotropic setup}

Now in the block setup,
\begin{align}
    \mathbb E\left[ x_\mu^2\right] = \sum_i \sigx{i} \phi_i \equiv r, \quad\quad
    \mathbb E\left[ x_\mu x_\nu\right] = \sum_i \sigx{i} \frac{F^i_\mu \cdot F^i_\nu}{D} = \sum_i \sigx{i} q^i_{\mu\nu} \equiv q_{\mu\nu} \sim O\left(\frac{1}{\sqrt D}\right)
\end{align}

Therefore we can parametrize as follows,
\begin{align}
    x_\mu = \eta_\mu \sqrt{1-\frac{q_{\mu\nu}}{r}} + z \sqrt{\frac{q_{\mu\nu}}{r}} \sim \eta_\mu (1-\frac{1}{2} \frac{q_{\mu\nu}}{r}) + z \sqrt{\frac{q_{\mu\nu}}{r}}\\
    x_\nu = \eta_\nu \sqrt{1-\frac{q_{\mu\nu}}{r}} + z \sqrt{\frac{q_{\mu\nu}}{r}} \sim \eta_\mu (1-\frac{1}{2} \frac{q_{\mu\nu}}{r}) + z \sqrt{\frac{q_{\mu\nu}}{r}}\\
\end{align}
where $\eta_\mu, \eta_\nu,z\sim\mathcal{N}(0,r)$. As before,
\begin{align}
    f(q_{\mu\nu}) &= \mathbb E_\eta \left[ \sigma(\eta)\right]^2 + \frac{q_{\mu\nu}}{r} \left( \mathbb E[\sigma'(\eta)]^2 \E[z^2] + \mathbb E[\sigma''(\eta)]\mathbb E [\sigma(\eta)] \E[z^2] - \mathbb E[\sigma'(\eta)\eta] \mathbb E[\sigma(\eta)] \right) + o(q_{\mu\nu}^2)\\
\end{align}

Now since $\mathbb E[z^2]=r$ and $\mathbb E[\sigma(\eta)\eta] = r \mathbb E[\sigma'(\eta)]$, the two last terms do not cancel out like before and we have :
\begin{align}
    f(0) = \mathbb E[\sigma(z)]^2, \quad\quad f'(0) = \frac{1}{r}\mathbb E[\sigma(z)\eta]^2 \mathbb E[z^2]= \frac{1}{r^2} \mathbb E [\sigma(z)z]^2 
\end{align}

Finally we have 
\begin{align}
    &\E_{\bs x}\left[\sigma\left(\frac{\bs x \cdot F_\mu}{\sqrt D}\right)\sigma\left(\frac{\bs x \cdot F_\nu}{\sqrt D}\right)\right] = \kappa_0^2 +  \kappa_1 ^2 \sum_i \sigx{i} \frac{F^i_\mu \cdot F^i_\nu}{D} + \kappa_\star^2 \delta_{\mu\nu}\\
    & \kappa_{0}=\underset{z\sim\mathcal{N}(0,r)}{\E}[\sigma(z)], \quad \kappa_{1}=\frac{1}{r}\underset{z\sim\mathcal{N}(0,r)}{\E}[z \sigma(z)]], \quad \kappa_{\star}=\sqrt{\underset{z\sim\mathcal{N}(0,r)}{\E}\left[\sigma(z)^{2}\right]-\kappa_{0}^{2}- r\kappa_{1}^{2}}\\
\end{align}

\subsection{Gibbs formulation of the problem}
\label{app:gibbs}

To obtain the test error, we need to find the typical value of the order parameters $m_s, q_s, q_w$. To do so, we formulate the optimization problem for the second layer weights as a Gibbs measure over the weights as in~\cite{gerace2020generalisation}:
\begin{align}
\mu_{\beta}\left(\bs{w} \mid\left\{\bs{x}^{\mu}, y^{\mu}\right\}\right)=\frac{1}{\gZ_{\beta}} e^{-\beta\left[\sum_{\mu=1}^{N} \ell\left(y^{\mu}, \bs{x}^{\mu} \cdot \bs{w}\right)+\frac{\lambda}{2}\|\bs{w}\|_{2}^{2}\right]}=\frac{1}{\gZ_{\beta}} 
\underbrace{\prod_{\mu=1}^{N} e^{-\beta \ell\left(y^{\mu}, \bs{x}^{\mu} \cdot \bs{w}\right)}}_{\equiv P_{y}\left(\bs{y} \mid \bs{w} \cdot \bs{x}^{\mu}\right)} 
\underbrace{\prod_{i=1}^{P} e^{-\frac{\beta \lambda}{2} w_{i}^{2}}}_{\equiv P_{w}(\bs{w})}  
\end{align}

Of key interest is the behavior of the free energy density, which is self-averaging in the high-dimensional limit and whose minimization gives the optimal value of the overlaps:
\begin{align}
    f_{\beta}=-\lim _{P \rightarrow \infty} \frac{1}{P} \E_{\left\{\bs{x}^{\mu}, y^{\mu}\right\}} \log \gZ_{\beta}
\end{align}

To calculate the latter, we use the Replica Trick from statistical physics:
\begin{align}
    \mathbb E_{\left\{\bs{x}^{\mu}, y^{\mu}\right\}} \log \mathcal{Z_\beta} = \lim_{r\to 0}\frac{D}{Dr}\mathbb E_{\left\{\bs{x}^{\mu}, y^{\mu}\right\}} \mathcal Z_\beta^r
\end{align}

The right hand side can be written in terms of the \emph{order parameters} :
\begin{align}
\E_{\left\{\bs{x}^{\mu}, y^{\mu}\right\}} \gZ_{\beta}^{r}&=\int \frac{\dd \rho \dd \hat{\rho}}{2 \pi} \int \prod_{a=1}^{r} \frac{\dd m_{s}^{a} \dd \hat{m}_{s}^{a}}{2 \pi} \int \prod_{1 \leq a \leq b \leq r} \frac{\dd q_{s}^{a b} \dd \hat{q}_{s}^{a b}}{2 \pi} \frac{\dd q_{w}^{a b} \dd \hat{q}_{w}^{a b}}{2 \pi} e^{p \Phi^{(r)}}\\
\Phi^{(r)}&=- \gamma \rho \hat{\rho}-\gamma \sum_{a=1}^{r} \sum_i m_{s,i}^{a} \hat{m}_{s,i}^{a}-\sum_{1 \leq a \leq b \leq r}\left(\gamma \sum_i q_{s,i}^{a b} \hat{q}_{s,i}^{a b}+q_{w} \hat{q}_{w}\right)\\&+\alpha \Psi_{y}^{(r)}\left(\rho, m_{s}^{a}, q_{s}^{a b}, q_{w}^{a b}\right) +\Psi_{w}^{(r)}\left(\hat{\rho}, \hat{m}_{s}^{a}, \hat{q}_{s}^{a b}, \hat{q}_{w}^{a b}\right)
\end{align}

Where we introduced and \emph{energetic} part $\Psi_y$ which corresponds to the likelihood of the order parameters taking a certain value based on the data, and an \emph{entropic} part $\Psi_w$ which corresponds to the volume compatible with those order parameters :
\begin{align}
\begin{array}{l}
\Psi_{y}^{(r)}=\log \int \dd y \int \dd \nu P_{y}^{0}(y \mid \nu) \int \prod_{a=1}^{r}\left[\dd \lambda^{a} P_{y}\left(y \mid \lambda^{a}\right)\right] P\left(\nu,\left\{\lambda^{a}\right\}\right) \\
\Psi_{w}^{(r)}=\frac{1}{P} \log \int \dd \bs\beta P_{\beta}\left(\bs\beta\right) e^{-\hat{\rho}\left\|\bs\beta\right\|^{2}} \int \prod_{a=1}^{r} \dd \bs{w}^{a} P_{w}\left(\bs{w}^{a}\right) e^{\sum_{1 \leq a \leq b \leq r}\left[\hat{q}_{w}^{a b} \bs{w}^{a} \cdot \bs{w}^{b}+ \sum_{i} \hat{q}_{s,i}^{a b} \bs{s}^{a}|_i \cdot \bs{s}^{b}|_i\right]-\sum_{a=1}^{r} \sum_i \hat{m}_{s,i}^{a} \bs{s}^{a}|_i \cdot \bs\beta|_i}
\end{array}
\end{align}

The calculation of $\Psi_y$ is exactly the same as in~\cite{gerace2020generalisation} : we defer the reader to the latter for details. As for $\Psi_w$, considering that $P_w(\bs{w}) = e^{-\frac{\lambda}{2} \lVert \bs{w} \rVert^2} $ we have (denoting the block indices as $i$):

\begin{align}
\Psi_{w}=&\lim _{P \rightarrow \infty} \E_{\bs\beta, \xi, \eta}\left[
\frac{1}{2} \frac{\eta^{2} \hat{q}_{w}}{\beta \lambda+\hat{V}_{w}}-\frac{1}{2} \log \left(\beta \lambda+\hat{V}_{w}\right)-\frac{1}{2P} \operatorname{tr} \log \left(\mathrm{I}_{D}+\hat{V}_{s} \Sigma\right)
-\frac{1}{2P} \bs{\mu}^{\top} \Sigma^{-1} \bs{\mu}\right.\\
&\left.+\frac{1}{2P}\left(\bs{b}+\Sigma^{-1}\bs{\mu}\right)^\top\hat{V}^{-1}_{s}\left(\bs{b}+\Sigma^{-1}\bs{\mu}\right)
-\frac{1}{2P}\left(\bs{b}+\Sigma^{-1} \mu\right)^{\top}\hat{V}^{-1}_{s}\left(\mathrm{I}_{D}+\hat{V}_{s} \Sigma\right)^{-1}\left(\bs{b}+\Sigma^{-1} \bs{\mu}\right)
\right]
\end{align}

where $\hat V_s$ is a block diagonal matrix where the block diagonals read $(\hat V_{s,1}, \hat V_{s,2})$, and 

\begin{align*}
    \bs{b} &= \left(\sqrt{\hat{q}_{s,1}} \xi_1 \mathbf{1}_{\phi_1 D}+\hat{m}_{s,1} \bs\beta_1, \sqrt{\hat{q}_{s,2}} \xi_2 \mathbf{1}_{\phi_2 D}+\hat{m}_{s,2} \bs\beta_2 \right)\\
    \bs{\mu} &= \frac{\sqrt{\hat{q}_w} \eta}{\beta \lambda + \hat{W}_w} \frac{\mathrm{F}\mathbf{1}_{P}}{\sqrt{P}} \\
    \Sigma &= \frac{1}{\beta \lambda + \hat{V}_w}\frac{\mathrm{FF}^{\top}}{P}
\end{align*}
and $\lambda$ here is the coefficient in the $\ell_2$ regularization.
Then, we have

\begin{align}
\begin{array}{c}
\E_{\eta}\left[\bs{\mu}^{\top} \bs{\Sigma}^{-1} \bs{\mu}\right]=\frac{1}{P} \frac{\hat{q}_{w}}{\left(\beta \lambda+\hat{V}_{w}\right)^{2}}\left(\mathrm{F} \mathbf{1}_{P}\right)^{\top} \bs{\Sigma}^{-1}\left(\mathrm{F} \mathbf{1}_{P}\right)=D \frac{\hat{q}_{w}}{\beta \lambda+\hat{V}_{w}} \\
\E_{\eta, \xi, \bs\beta}\left\|\bs{b}+\Sigma^{-1} \bs{\mu}\right\|^{2}= \sum_i \phi_i D\left(\hat{m}_{s,i}^{2}+\hat{q}_{s,i}\right)+\frac{1}{P} \hat{q}_{w} \operatorname{tr}(\mathrm{FF}^\top)^{-1} \\
\E_{\eta, \xi, \bs\beta}\left(\bs{b}+\Sigma^{-1} \bs{\mu}\right)^{\top}\left(\mathrm{I}_{D}+\hat{V}_{s} \Sigma\right)^{-1}\left(\bs{b}+\Sigma^{-1} \bs{\mu}\right)=\frac{1}{P} \hat{q}_{w} \operatorname{tr}\left[\mathrm{FF}^{\top}\left(\mathrm{I}_{D}+\hat{V}_{s} \Sigma\right)^{-1}\right] \\
+\sum_i \left(\hat{m}_{s,i}^{2}+\hat{q}_{s,i}\right) \operatorname{tr}\left(\mathrm{I}_{D}+\hat{V}_{s_i} \Sigma_i \right)^{-1}
\end{array}
\end{align}


\begin{align}
\Psi_{w}= &- \frac{1}{2} \log \left(\beta \lambda+\hat{V}_{w}\right)-\frac{1}{2} \lim _{P \rightarrow \infty} \frac{1}{P} \operatorname{tr} \log \left(\mathrm{I}_{D}+\frac{\hat{V}_{s}}{\beta \lambda+\hat{V}_{w}} \frac{\mathrm{FF}^{\top}}{P}\right) \\
&+\sum_i \frac{\hat{m}_{s,i}^{2}+\hat{q}_{s,i}}{2 \hat{V}_{s,i}}\left[\phi_i \gamma \right]
-\lim _{P \rightarrow \infty} \frac{1}{P} \operatorname{tr}\left( (\hat{m}_s^2 + \hat{q_s} )\left(\mathrm{I}_{D}+\frac{\hat{V}_{s}}{\beta \lambda+\hat{V}_{w}} \frac{\mathrm{FF}^{\top}}{P}\right)^{-1}\right) \\
&+ \frac{1}{2} \frac{\hat{q}_{w}}{\beta \lambda+\hat{V}_{w}}\left[1 - \gamma \right] +  \frac{\hat{q}_{w}}{2} \lim _{P \rightarrow \infty}\frac{1}{P} \left[ \operatorname{tr}\left( \hat{V}_s^{-1} (\mathrm{FF}^{\top})^{ -1}\right)
-  \operatorname{tr}\left((\hat{V}_s \mathrm{FF}^\top) ^{-1}\left(\mathrm{I}_{D}+\frac{\hat{V}_{s}}{\beta \lambda+\hat{V}_{w}} \frac{\mathrm{FF}^{\top}}{P}\right)^{-1}\right)\right]
\end{align}
where $(\hat{m}_s^2 + \hat{q_s} )$ and $\hat V_s$ in the trace operator are to be understood as diagonal matrices here. Using the following simplification
\begin{align}
\operatorname{tr}\left( V_s^{-1} (\mathrm{FF}^{\top})^{ -1}\right)
- \operatorname{tr}\left((V_s \mathrm{FF}^\top) ^{-1}\left(\mathrm{I}_{D}+\frac{\hat{V}_{s}}{\beta \lambda+\hat{V}_{w}} \frac{\mathrm{FF}^{\top}}{P}\right)^{-1}\right) \\
= \frac{1}{P (\beta \lambda + \hat{V}_w)} \operatorname{tr} \left( \mathrm{I}_d +\frac{\hat{V}_{s}}{\beta \lambda+\hat{V}_{w}} \frac{\mathrm{FF}^{\top}}{P}\right),
\end{align}
we finally obtain
\begin{align}
\Psi_{w}= &- \frac{1}{2} \log \left(\beta \lambda+\hat{V}_{w}\right)-\frac{1}{2} \lim _{P \rightarrow \infty} \frac{1}{P} \operatorname{tr} \log \left(\mathrm{I}_{D}+\frac{\hat{V}_{s}}{\beta \lambda+\hat{V}_{w}} \frac{\mathrm{FF}^{\top}}{P}\right) \\
&+\sum_i \frac{\hat{m}_{s,i}^{2}+\hat{q}_{s,i}}{2 \hat{V}_{s,i}}\left[\phi_i \gamma \right]
-\lim _{P \rightarrow \infty} \frac{1}{P} \operatorname{tr}\left( (\hat{m}_s^2 + \hat{q_s} )\left(\mathrm{I}_{D}+\frac{\hat{V}_{s}}{\beta \lambda+\hat{V}_{w}} \frac{\mathrm{FF}^{\top}}{P}\right)^{-1}\right) \\
&+ \frac{1}{2} \frac{\hat{q}_{w}}{\beta \lambda+\hat{V}_{w}}\left[1 - \gamma 
+ \lim _{P \rightarrow \infty} \frac{1}{P} \operatorname{tr}\left(\left(\mathrm{I}_{D}+\frac{\hat{V}_{s}}{\beta \lambda+\hat{V}_{w}} \frac{\mathrm{FF}^{\top}}{P}\right)^{-1}\right)\right]
\end{align}.

Define $M = \frac{1}{P} \hat V_s F F^\top$. $\Psi_w$ involves the following term :
\begin{align}
    \frac{1}{P} \operatorname{tr}\left(\mathrm{I}_{D}+\frac{\hat{V}_{s}}{\beta \lambda+\hat{V}_{w}} M\right)^{-1} = \gamma (\beta \lambda + \hat V_w) g(-\beta \lambda + \hat V_w)
\end{align}
Where $g$ is the Stieljes transform of $M$ : $g(z) = \frac{1}{D} \Tr (z-M)^{-1}$. 

\subsection{Some random matix theory for block matrices}
\label{app:rmt}

To calculate the desired Stieljes transform, we specialize to the case of Gaussian random feature matrices and use again tools from Statistical Physics.

\paragraph{Isotropic setup}

We first consider the isotropic setup where $\hat V_s$ is a scalar. Then we have
\begin{align}
    g(z) &= \frac{1}{D}\Tr (z-M)^{-1}\\
    &= -\frac{1}{D} \frac{d}{dz} \Tr \log (z-M)\\
    &= - \frac{1}{D} \frac{d}{dz} \log \det (z-M)\\
    &= - \frac{2}{D} \frac{d}{dz} \left\langle \log \int dy e^{-\frac{1}{2} y (z-M) y^\top }\right\rangle
\end{align}
where $\langle .  \rangle$ stands for the average over disorder, here the matrix $F$.
Then we use the replica trick,
\begin{align}
    \langle \log Z\rangle \to_{n\to 0} \frac{1}{n} \log \langle Z^n\rangle
\end{align}

Therefore we need to calculate $Z^n$:
\begin{align}
    \langle Z^n \rangle &= \int \prod_{a=1}^n d\vec y_a e^{-\frac{1}{2} z \sum_a \vec y_a^2} \int dF e^{-\frac{1}{2} \mathrm{FLF}^\top}\\
    &= \int \prod_{a=1}^n d\vec y_a e^{-\frac{1}{2} z \sum_a \vec y_a^2} (\det L)^{-P/2}
\end{align}
where $L = \mathbb{I}_d - \frac{\gamma \hat V_s}{D} \sum_a  y_a \vec y_a^\top$. Here we decompose the vectors $\vec y$ into two parts. Then we use that $\det L = \det \tilde L$, where $\tilde L_{ab} = \delta_{ab} - \frac{\gamma \hat V_s}{D} \vec y_a \cdot \vec y_b$. Then,
\begin{align}
    \langle Z^n\rangle = \int \prod_a d\vec y_a e^{\frac{1}{2} \sum_a \vec y_a \cdot \vec y_a} \det \left[ \mathbb{I}_d - \frac{\gamma \hat V_s}{D} Y^\top Y \right]^{-\frac{P}{2}}
\end{align}
where $Y \in \sR^{d \times n }$ with $Y_{ia} = y^a_i$.
We introduce $1 = \int dQ_{ab} \delta (dQ_{ab} - \vec y_a \cdot \vec y_b)$, and use the Fourier representation of the delta function, yielding 
\begin{align}
    \langle Z^n \rangle &= \int dQ e^{-\frac{D}{2} z \Tr Q} \left( \det \left[ \mathbb{I} - \hat V_s Q\right]\right)^{-\frac{P}{2}} \int d\hat Q_{ab} e^{\sum_{ab} d Q_{ab} \hat Q_{ab} - \hat Q_{ab}\vec y_a \vec y_b} \\
    &= \int dQ d\hat Q e^{-d S[Q,\hat Q]}
\end{align}
where
\begin{align}
    S[Q,\hat Q] = \frac{1}{2} z \Tr Q + \frac{1}{2\gamma} \log \det \left(1-\hat V_s Q\right) + \frac{1}{2} \log \det\left(2\hat Q\right) - \Tr \left(Q\hat Q\right) 
\end{align}
A saddle-point on $\hat Q$ gives $Q = (2\hat Q)^{-1}$. Therefore we can replace in $S$,
\begin{align}
    S[Q] = \frac{1}{2} z \Tr Q + \frac{1}{2\gamma} \log \det \left(1-\hat V_s Q\right) - \frac{1}{2} \log \det\left(Q\right)
\end{align}

In the RS ansatz $Q_{ab} = q\delta_{ab}$, this yields
\begin{align}
    S[q]/n = \frac{1}{2} z q + \frac{1}{2\gamma} \log \left(1-\hat V_s q\right) - \frac{1}{2} \log q
\end{align}

Now we may apply a saddle point method to write
\begin{align}
    \langle Z^n \rangle = e^{-d S[q^\star]}
\end{align}
where $q^\star$ minimizes the action, i.e. $\frac{dS}{dq}|_{q^\star} = 0$:
\begin{align}
    z - \frac{\hat V_s}{1-\hat V_s q^\star} - \frac{1}{q^\star} = 0
\end{align}
Therefore, 
\begin{align}
    g(z) = -\frac{2}{D} \frac{d}{dz} (-D S[q^\star]) = q^\star(z)
\end{align}

\paragraph{Anisotropic setup}

Now we consider that $V$ is a black diagonal matrix, with blocks of size $\phi_i d$ with values $\hat V_{s,i}$. We need to adapt the calculation by decomposing the auxiliary fields $y$ along the different blocks, then define separately the overlaps of the blocks $q_i$. Then we obtain the following action:
\begin{align}
    S[\{q_i\}]/n = \frac{1}{2} z \sum_i \phi_i q_i + \frac{1}{2\gamma} \log \left(1- \sum_i \phi_i \hat V_{s,i} q_i \right) - \sum_i \frac{\phi_i }{2} \log (q_i)
\end{align}

To obtain the Stieljes transform, we need to solve a system of coupled equations :
\begin{align}
    &g(z) = \sum_i \phi_i q_i^\star \\
    &\phi_i z \Omega q_i^\star - \phi_i \hat V_{i} q_i^\star - \phi_i\Omega = 0\\
    &\Omega = 1-\sum_i \phi_i \hat V_{s,i} q_i^\star
\end{align}

We therefore conclude that
\begin{align}
    \lim _{P \rightarrow \infty} \frac{1}{P} \operatorname{tr}\left( \left(\mathrm{I}_{D}+\frac{\hat{V}_{s}}{\beta \lambda+\hat{V}_{w}} \frac{\mathrm{FF}^{\top}}{P}\right)^{-1}\right) &= \gamma (\beta \lambda + \hat V_w) \sum_i \phi_i q_i^\star \\
    \lim _{P \rightarrow \infty} \frac{1}{P} \operatorname{tr}\left( (\hat{m}_s^2 + \hat{q_s} )\left(\mathrm{I}_{D}+\frac{\hat{V}_{s}}{\beta \lambda+\hat{V}_{w}} \frac{\mathrm{FF}^{\top}}{P}\right)^{-1}\right) &= \gamma (\beta \lambda + \hat V_w) \sum_i (\hat{m}_{s,i}^2 + \hat{q}_{s,i} ) \phi_i q_i^\star 
\end{align}

\subsection{Obtaining the saddle-point equations}
\label{app:saddle-point}

The saddle-point equations of~\cite{gerace2020generalisation} become the following in the anisotropic setup :
\begin{align}
\left\{\begin{array}{ll}
\hat{r}_{s,i}=-2  \sigx{i} \frac{\alpha}{\gamma} \partial_{r_{s,i}} \Psi_{y}(R, Q, M) & \quad r_{s,i}=- \frac{2}{\gamma} \partial_{\hat{r}_{s,i}} \Psi_{w}\left(\hat{r}_{s,i}, \hat{q}_{s,i}, \hat{m}_{s,i}, \hat{r}_{w}, \hat{q}_{w}\right) \\
\hat{q}_{s,i}=-2  \sigx{i}\frac{\alpha}{\gamma} \partial_{q_{s,i}} \Psi_{y}(R, Q, M) & q_{s,i}=-\frac{2}{\gamma} \partial_{\hat{q}_{s,i}} \Psi_{w}\left(\hat{r}_{s,i}, \hat{q}_{s,i}, \hat{m}_{s,i}, \hat{r}_{w}, \hat{q}_{w}\right) \\
\hat{m}_{s,i}=  \sigx{i}\frac{\alpha}{\gamma} \partial_{m_{s,i}} \Psi_{y}(R, Q, M) & m_{s,i}=\frac{1}{\gamma} \partial_{\hat{m}_{s,i}} \Psi_{w}\left(\hat{r}_{s,i}, \hat{q}_{s,i}, \hat{m}_{s,i}, \hat{r}_{w}, \hat{q}_{w}\right) \\
\hat{r}_{w}=-2 \alpha \partial_{r_{w}} \Psi_{y}(R, Q, M) & r_{w}=-2 \partial_{\hat{r}_{w}} \Psi_{w}\left(\hat{r}_{s,i}, \hat{q}_{s,i}, \hat{m}_{s,i}, \hat{r}_{w}, \hat{q}_{w}\right) \\
\hat{q}_{w}=-2 \alpha \partial_{q_{w}} \Psi_{y}(R, Q, M) & q_{w}=-2 \partial_{\hat{q}_{w}} \Psi_{w}\left(\hat{r}_{s,i}, \hat{q}_{s,i}, \hat{m}_{s,i}, \hat{r}_{w}, \hat{q}_{w}\right)
\end{array}\right.
\end{align}

The saddle point equations corresponding to $\Psi_w$ do not depend on the learning task, and can be simplified in full generality to the following set of equations :
\begin{align}
\left\{\begin{array}{l}
V_{s,i}=\frac{1}{\hat{V}_{s,i}}\left(\phi_i-z_i g_{\mu}(-z_i)\right) \\
q_{s,i}=\frac{ \sigb{i} \hat m_{s,i}^2 +\hat{q}_{s,i}}{\hat{V}_{s,i}^{2}}\left[\phi_i -2 z_i g_{\mu}(-z_i)+z_i^{2} g_{\mu}^{\prime}(-z_i)\right]-\frac{\hat{q}_{w}}{\left(\beta \lambda+\hat{V}_{w}\right) \hat{V}_{s,i}}\left[-z_i g_{\mu}(-z_i)+z_i^{2} g_{\mu}^{\prime}(-z_i)\right] \\
m_{s,i}=\frac{ \sigb{i} \hat{m}_{s,i}}{\hat{V}_{s,i}}\left(\phi_i -z_i g_{\mu}(-z_i)\right) \\
V_{w}=\sum_i \frac{\gamma}{\beta \lambda+\hat{V}_{w}}\left[\frac{1}{\gamma}-1+z_i g_{\mu}(-z_i)\right] \\
q_{w}=\sum_i \ \frac{ \sigb{i} \hat m_s^2 + \hat{q}_{s,i}}{\left(\beta \lambda+\hat{V}_{w}\right) \hat{V}_{s,i}}\left[-z_i g_{\mu}(-z_i)+z_i^{2} g_{\mu}^{\prime}(-z_i)\right] + \gamma \frac{\hat{q}_{w}}{\left(\beta \lambda+\hat{V}_{w}\right)^{2}}\left[\frac{1}{\gamma}-1+z_i^{2} g_{\mu}^{\prime}(-z_i)\right]
\end{array}\right.
\end{align}

As for the saddle point equations corresponding to $\Psi_y$, they depend on the learning task. We can solve analytically for the two setups below.

The solution of the saddle-point equations allow to obtain the order parameter values that determine the covariance of  $\nu$ and $\lambda$ (for one given replica, say $a=1$) and hence to compute the test error as explained at the beginning of the appendix.   

\paragraph{Square loss regression}

Let us first specialize to a simple regression setup where the teacher is a Gaussian additive channel :
\begin{align}
    \ell(y, x) = \frac 1 2 (x - y)^2\\
    \gP(x|y) = \frac{1}{\sqrt{2 \pi \Delta}} e^{- \frac {(x-y)^2}{2 \Delta}}
\end{align}
In this case, the saddle-point equations simplify to :
\begin{align}
\left\{\begin{array}{l}
\hat{V}_{s,i}^{\infty}=\sigx{i}\frac{\alpha}{\gamma} \frac{\kappa_{1,i}^{2}}{1+V^{\infty}} \\
\hat{q}_{s,i}^{0}=\sigx{i}\alpha \kappa_{1,i}^{2} \frac{ \phi_i \sigb{i} +\Delta+Q^{\infty}-2 M^{\infty}}{\gamma\left(1+V^{\infty}\right)^{2}} \\ \hat{m}_{s,i}= \sigx{i}\frac{\alpha}{\gamma} \frac{\kappa_{1,i}}{1+V^{\infty}} \\
\hat{V}_{w}^{\infty}=\frac{\alpha\sum_i \phi_i \kappa_{\star,i}^{2}}{1+V^{\infty}} \\ \hat{q}_{w}^{\infty}=\alpha \sum_i \phi_i\kappa_{\star,i}^{2} \frac{1+\Delta+Q^{\infty}-2 M^{\infty}}{\left(1 +V^{\infty}\right)^{2}}
\end{array}\right.
\end{align}

\paragraph{Square loss classification}

Next we examine the classification setup where the teacher gives binary labels with a sign flip probability of $\Delta$, and the student learns them through the mean-square loss:
\begin{align}
    &\gP(x|y) = (1-\Delta) \delta(x-\sign(y)) + \Delta \delta(x+\sign(y)), \quad \Delta \in [0,1]\\
    &\ell(y, x) = \frac{1}{2} (x - y)^2
\end{align} 

In this case, the equations simplify to :
\begin{align}
\left\{\begin{array}{l}
\hat{V}_{s,i}^{\infty}=\sigx{i}\frac{\alpha}{\gamma} \frac{\kappa_{1,i}^{2}}{1+V^{\infty}} \\
\hat{q}_{s,i}^{0}=\sigx{i}\alpha \kappa_{1,i}^{2} \frac{ \phi_i \sigb{i} +Q^{\infty}-2 \frac{(1-2\Delta) \sqrt 2}{\sqrt \pi}M^{\infty}}{\gamma\left(1+V^{\infty}\right)^{2}} \\ \hat{m}_{s,i}= \sigx{i}\frac{\alpha}{\gamma} \frac{\kappa_{1,i}}{1+V^{\infty}} \\
\hat{V}_{w}^{\infty}=\frac{\alpha\sum_i \phi_i \kappa_{\star,i}^{2}}{1+V^{\infty}} \\ 
\hat{q}_{w}^{\infty}=\alpha \sum_i \phi_i\kappa_{\star,i}^{2} \frac{1+Q^{\infty}-2  \frac{(1-2\Delta)\sqrt 2}{\sqrt \pi} M^{\infty}}{\left(1 +V^{\infty}\right)^{2}}
\end{array}\right.
\end{align}

\paragraph{Logistic loss classification}

For general loss functions such as the cross-entropy loss, $\ell(x,y) = \log (1+e^{-xy})$, the saddle-point equations for $\Psi_y$ do not simplify and one needs to evaluate the integrals over $\xi$ numerically. 


\subsection{Training loss}
\label{app:train-derivation}

To calculate the training loss, we remove the regularization term:
\begin{align}
    \epsilon_{t}=\frac{1}{N} \E_{\left\{\boldsymbol{x}^{\mu}, y^{\mu}\right\}}\left[\sum_{\mu=1}^{N} \ell\left(y^{\mu}, \boldsymbol{x}^{\mu} \cdot \boldsymbol{\hat { w }}\right)\right]
\end{align}

As explained in~\cite{gerace2020generalisation}, the latter can be written as 
\begin{align}
    \epsilon_{t} = \E_{\xi}\left[\int_{\sR} \dd y \gZ_{y}^{0}\left(y, \omega_{0}, V_0\right) \ell\left(y, \eta\left(y, \omega_{1}\right)\right)\right], \quad \xi\sim\mathcal{N}(0,1)
\end{align}
where $\gP$ is the teacher channel defined in \eqref{eq:teacher-def}, and we have:
\begin{align}
    \omega_{0} &= M / \sqrt{Q} \xi\\
    V_0 &= \rho - M^2 / Q\\
    \omega_{1} &= \sqrt{Q}\xi\\
    \eta(y, \omega) &= \underset{x}{\argmin}\frac{(x-\omega)^2}{2 V} + \ell(y, x)\\
    \gZ_{y}^0 \left(y, \omega\right) &= \int \frac{\dd x}{\sqrt{2 \pi V_0}} \, e^{- \frac{1}{2 V_0}(x - \omega)^2}\gP(y | x)
\end{align} 

\paragraph{Square loss regression}

Consider again the regression setup where the teacher is a Gaussian additive channel $\gP(x|y) = \frac{1}{\sqrt{2 \pi \Delta}} e^{- \frac {(x-y)^2}{2 \Delta}}$ and the loss is $\ell(y, x) = \frac 1 2 (x - y)^2$. These assumptions imply
\begin{align}
    \gZ_{y}^0 \left(y, \omega\right) &= \frac{1}{\sqrt{2 \pi (V_0 + \Delta)}}e^{-\frac{(y - \omega)^2}{2(V_0 + \Delta)}}, \\
    \eta(y, \omega) &= \frac{\omega + Vy }{1 + V},
\end{align}
which yields the simple formula for the training error
\begin{align}
    \epsilon_t &= \frac{1}{2(1+V)^2}\E_\xi\left[ \int \dd y \gZ_{y}^0 \left(y, \omega\right) (y -\omega_1)^2 \right]\\
    &= \frac{1}{2(1+V)^2}\E_\xi\left[ V_0 + \Delta + (\omega_0- \omega_1)^2 \right]\\
    &= \frac{1}{2(1+V)^2}\left( \rho + Q - 2 M + \Delta \right)\\
    &= \frac{\epsilon_g+\Delta}{(1+V)^2}.
\end{align}

Notice that the training loss is closely related to the test loss; inverting the latter expression, we have $\epsilon_g = (1+V^2)\epsilon_t - \Delta$, showing that $V$ acts as a variance term opening up a generalization gap.

\paragraph{Square loss classification}

Next we specialize to our classification case study where the teacher is a Gaussian additive channel $\gP(x|y) = (1-\Delta) \delta(x-\sign(y)) + \Delta \delta(x+\sign(y))$ and the loss is $\ell(y, x) = \frac 1 2 (x - y)^2$. 
These assumptions imply
\begin{align}
    \gZ_{y}^0 \left(y, \omega\right) &= \frac{1}{\sqrt{2 \pi V_0 }}\left( (1-\Delta) e^{-\frac{(y - \omega)^2}{2V_0 }} + \Delta e^{-\frac{(-y - \omega)^2}{2V_0 }} \right), \\
    \eta(y, \omega) &= \frac{\omega + Vy }{1 + V},
\end{align}
which yields the simple formula for the training error
\begin{align}
    \epsilon_t &= \frac{1}{2(1+V)^2}\E_\xi\left[ V_0 + (1-\Delta) (\omega_0- \omega_1)^2 + \Delta (\omega_0 + \omega_1)^2 \right]\\
    &= \frac{1}{2(1+V)^2}\left( \rho + Q - 2 (1- 2 \Delta) M \right)\\
\end{align}

This expression is similar to the one obtained in the regression setup, except for the role of the noise, which reflects label flipping instead of additive noise. As a sanity check, note that flipping all the labels, i.e. $\Delta=1$, is equivalent to the transformation $M\to-M$, as one could expect.

\end{document}